\documentclass[5p,times,twocolumn,number]{elsarticle}

\usepackage[T1]{fontenc}
\usepackage[utf8]{inputenc}
\usepackage{amsmath,amssymb}
\usepackage{graphicx}
\usepackage{booktabs}
\usepackage{array}
\usepackage{xcolor}
\usepackage{microtype}
\usepackage[hidelinks]{hyperref}
\usepackage{xurl}%
\usepackage[capitalise,noabbrev]{cleveref}
\usepackage{orcidlink}

\newif\ifdraftmarks
\draftmarksfalse
\makeatletter
\newcommand{\num@fail}[1]{\PackageError{numbers}{#1}{}}

\newcommand{\nb}[1]{%
  \ifcsname num@#1\endcsname\csname num@#1\endcsname
  \else\ifdraftmarks{#1}\else\num@fail{number key '#1' is not defined in numbers.tex}\fi\fi}
\newcommand{\defnum}[2]{\expandafter\gdef\csname num@#1\endcsname{#2}}

\makeatother

\newcommand{\cond}[1]{\mbox{#1}}
\newcommand{\vitw}{\cond{ViT-W}}
\newcommand{\vith}{\cond{ViT-H}}
\newcommand{\llmw}{\cond{LLM-W}}
\newcommand{\llmh}{\cond{LLM-H}}
\newcommand{\orig}{\cond{Original}}
\newcommand{\hflip}{\cond{H-flip}}
\newcommand{\vflip}{\cond{V-flip}}
\newcommand{\qsmall}{Qwen2.5-VL-3B}
\newcommand{\qlarge}{Qwen2.5-VL-7B}
\newcommand{\ivl}{InternVL3-2B}

\defnum{cv-agree-vitw-3b}{87.5\%}
\defnum{cv-agree-orig-3b}{46.0\%}
\defnum{cv-agree-rot-3b}{82.2\%}
\defnum{cv-flip-vitw-lr-3b}{83.3\%}
\defnum{cv-flip-vitw-lr-3b-n}{186}
\defnum{cv-flip-vitw-ud-3b}{4.4\%}
\defnum{cv-flip-llmw-lr-3b}{0.5\%}
\defnum{cv-flip-vith-ud-3b}{45.5\%}
\defnum{cv-flip-llmh-ud-3b}{9.1\%}
\defnum{cv-agree-vitw-7b}{93.2\%}
\defnum{cv-agree-orig-7b}{8.6\%}
\defnum{cv-agree-rot-7b}{81.5\%}
\defnum{cv-flip-vitw-lr-7b}{95.4\%}
\defnum{cv-flip-vitw-lr-7b-n}{327}
\defnum{cv-flip-vitw-ud-7b}{5.1\%}
\defnum{cv-flip-llmw-lr-7b}{0.3\%}
\defnum{cv-flip-vith-ud-7b}{51.4\%}
\defnum{cv-flip-llmh-ud-7b}{12.1\%}
\defnum{cv-agree-vitw-ivl}{84.3\%}
\defnum{cv-agree-orig-ivl}{19.8\%}
\defnum{cv-agree-rot-ivl}{76.2\%}
\defnum{cv-flip-vitw-lr-ivl}{88.5\%}
\defnum{cv-flip-vitw-lr-ivl-n}{244}
\defnum{cv-flip-vitw-ud-ivl}{7.8\%}
\defnum{cv-flip-llmw-lr-ivl}{2.5\%}
\defnum{cv-flip-llmw-lr-ivl-hi}{5\%}
\defnum{cv-flip-vith-ud-ivl}{69.8\%}
\defnum{cv-flip-llmh-ud-ivl}{4.2\%}
\defnum{cv-left-pred-3b}{71.0\%}
\defnum{cv-left-gold}{47.8\%}
\defnum{order-only-w-lr-3b}{2.3\%}
\defnum{order-only-h-lr-3b}{3.7\%}
\defnum{order-only-h-ud-3b}{16.9\%}
\defnum{colmajor-lr-3b}{3.1\%}
\defnum{colmajor-ud-3b}{7.9\%}
\defnum{id-only-w-lr-3b}{0.8\%}
\defnum{order-only-w-lr-7b}{2.9\%}
\defnum{order-only-h-lr-7b}{4.4\%}
\defnum{order-only-h-ud-7b}{18.7\%}
\defnum{colmajor-lr-7b}{3.9\%}
\defnum{colmajor-ud-7b}{15.4\%}
\defnum{id-only-w-lr-7b}{0.3\%}
\defnum{rowperm-onlyh-7b}{20.8\%}
\defnum{rowperm-rowshift-7b}{11.6\%}
\defnum{rowperm-shuffle-7b}{11.2\%}
\defnum{rowperm-vflip-share-7b}{80.0\%}
\defnum{blk31-3b}{47.0\%}
\defnum{blkall-3b}{46.7\%}
\defnum{blkbutlast-3b}{18.3\%}
\defnum{blk23-3b}{2.1\%}
\defnum{blk31-7b}{68.1\%}
\defnum{blkall-7b}{89.3\%}
\defnum{blkbutlast-7b}{24.8\%}
\defnum{blkglobal-7b}{89.6\%}
\defnum{dose-suf4-lr}{47.8\%}
\defnum{dose-all-lr}{46.7\%}
\defnum{dose-pre28-lr}{17.8\%}
\defnum{dose-suf4-ud}{9.0\%}
\defnum{dose-all-ud}{27.7\%}
\defnum{dose-selfcheck}{100.0\%}
\defnum{wu-flip-vitw-lr-3b}{100.0\%}
\defnum{wu-mirror-vitw-lr-3b}{82.3\%}
\defnum{wu-flip-llmw-lr-3b}{0.0\%}
\defnum{wu-flip-vith-lr-3b}{50.5\%}
\defnum{wu-flip-vitw-ud-3b}{14.3\%}
\defnum{wu-flip-vith-ud-3b}{83.8\%}
\defnum{wu-mirror-vith-ud-3b}{81.0\%}
\defnum{wu-agree-vitw-3b}{81.1\%}
\defnum{wu-flip-vitw-lr-7b}{100.0\%}
\defnum{wu-mirror-vitw-lr-7b}{72.1\%}
\defnum{wu-flip-llmw-lr-7b}{0.0\%}
\defnum{wu-flip-vith-lr-7b}{60.4\%}
\defnum{wu-flip-vitw-ud-7b}{11.1\%}
\defnum{wu-flip-vith-ud-7b}{80.8\%}
\defnum{wu-mirror-vith-ud-7b}{79.8\%}
\defnum{wu-agree-vitw-7b}{71.4\%}
\defnum{wu-flip-vitw-lr-ivl}{99.0\%}
\defnum{wu-mirror-vitw-lr-ivl}{94.9\%}
\defnum{wu-flip-llmw-lr-ivl}{1.0\%}
\defnum{wu-flip-vith-lr-ivl}{37.8\%}
\defnum{wu-flip-vitw-ud-ivl}{19.1\%}
\defnum{wu-flip-vith-ud-ivl}{77.9\%}
\defnum{wu-mirror-vith-ud-ivl}{52.9\%}
\defnum{wu-agree-vitw-ivl}{93.2\%}
\defnum{wu-offaxis-swap-ivl}{33}
\defnum{wu-offaxis-swap-ivl-n}{196}
\defnum{wu-offaxis-swap-ivl8b}{1}
\defnum{wu-offaxis-swap-ivl8b-n}{155}
\defnum{wu-swap-8b-diff}{$-$16.2}
\defnum{wu-swap-8b-diff-ci}{[$-$21.7, $-$10.8]}
\defnum{wu-offaxis-onunder-7b}{111}
\defnum{wu-offaxis-changed-7b}{118}
\defnum{tie-agree-vitw-7b}{94.4\%}
\defnum{tie-flip-vitw-7b}{95.7\%}
\defnum{tie-flip-llmw-7b}{0.0\%}
\defnum{tie-rows-wu-3b}{1.9\%}
\defnum{tie-rows-wu-ivl}{0.9\%}
\defnum{blank-lr-pos-3b}{46.7\%}
\defnum{blank-lr-prior-3b}{53.0\%}
\defnum{blank-lr-pos-7b}{89.3\%}
\defnum{blank-ud-pos-7b}{42.7\%}
\defnum{blank-lr-pos-ivl}{77.0\%}
\defnum{blank-ud-pos-ivl}{38.2\%}
\defnum{depth-test-vith-3b}{37.3\%}
\defnum{depth-test-vith-3b-ci}{[32.1, 42.7]}
\defnum{depth-test-vitw-3b}{27.1\%}
\defnum{depth-test-vitw-3b-ci}{[22.5, 32.2]}
\defnum{depth-test-llmh-3b}{5.7\%}
\defnum{depth-test-n-3b}{314}
\defnum{depth-test-vith-7b}{43.9\%}
\defnum{depth-test-vith-7b-ci}{[38.6, 49.5]}
\defnum{depth-test-vitw-7b}{19.7\%}
\defnum{depth-test-vitw-7b-ci}{[15.7, 24.5]}
\defnum{depth-test-llmh-7b}{6.1\%}
\defnum{depth-test-n-7b}{314}
\defnum{depth-spec-vith-con-7bt}{51.0\%}
\defnum{depth-spec-vith-inc-7bt}{17.2\%}
\defnum{depth-spec-vith-con-n-7bt}{196}
\defnum{depth-spec-vith-inc-n-7bt}{29}
\defnum{depth-spec-vitw-con-7bt}{16.3\%}
\defnum{depth-spec-vitw-inc-7bt}{27.6\%}
\defnum{depth-spec-vitw-con-n-7bt}{196}
\defnum{depth-spec-vitw-inc-n-7bt}{29}
\defnum{depth-spec-vith-con-3bt}{46.8\%}
\defnum{depth-spec-vith-inc-3bt}{10.8\%}
\defnum{depth-spec-vith-con-n-3bt}{188}
\defnum{depth-spec-vith-inc-n-3bt}{37}
\defnum{depth-spec-vitw-con-3bt}{20.2\%}
\defnum{depth-spec-vitw-inc-3bt}{51.4\%}
\defnum{depth-spec-vitw-con-n-3bt}{188}
\defnum{depth-spec-vitw-inc-n-3bt}{37}
\defnum{depth-spec-vith-con-7bv}{48.1\%}
\defnum{depth-spec-vith-inc-7bv}{23.5\%}
\defnum{depth-spec-vith-con-n-7bv}{156}
\defnum{depth-spec-vith-inc-n-7bv}{34}
\defnum{depth-spec-vitw-con-7bv}{10.5\%}
\defnum{depth-spec-vitw-inc-7bv}{35.5\%}
\defnum{depth-spec-vitw-con-n-7bv}{162}
\defnum{depth-spec-vitw-inc-n-7bv}{31}
\defnum{depth-spec-vith-con-3bv}{37.4\%}
\defnum{depth-spec-vith-inc-3bv}{9.7\%}
\defnum{depth-spec-vith-con-n-3bv}{147}
\defnum{depth-spec-vith-inc-n-3bv}{31}
\defnum{depth-spec-vitw-con-3bv}{15.0\%}
\defnum{depth-spec-vitw-inc-3bv}{50.0\%}
\defnum{depth-spec-vitw-con-n-3bv}{153}
\defnum{depth-spec-vitw-inc-n-3bv}{30}
\defnum{video-vid-vitw-7b}{96.0\%}
\defnum{video-img-vitw-7b}{97.0\%}
\defnum{video-vid-llmw-7b}{0.0\%}
\defnum{words-rev2-3b}{32.0\%}
\defnum{words-ctl2-3b}{3.0\%}
\defnum{words-rev2-7b}{34.0\%}
\defnum{words-ctl2-7b}{4.0\%}
\defnum{refbox-n-7b}{131}
\defnum{pubft-agree-base}{79.6\%}
\defnum{pubft-agree-base-n}{225}
\defnum{pubft-llmw-base}{0.0\%}
\defnum{pubft-d3conf-base}{54.7\%}
\defnum{pubft-agree-sl}{87.5\%}
\defnum{pubft-agree-sl-n}{279}
\defnum{pubft-llmw-sl}{0.0\%}
\defnum{pubft-d3conf-sl}{45.3\%}
\defnum{pubft-gain-sl}{+8.0}
\defnum{pubft-gain-sl-ci}{[+5.9, +10.0]}
\defnum{pubft-agree-sq}{81.6\%}
\defnum{pubft-agree-sq-n}{136}
\defnum{pubft-llmw-sq}{0.8\%}
\defnum{pubft-d3conf-sq}{50.0\%}
\defnum{pubft-gain-sq}{$-$3.2}
\defnum{pubft-gain-sq-ci}{[$-$4.8, $-$1.5]}
\defnum{pubft-agree-sr}{77.3\%}
\defnum{pubft-agree-sr-n}{203}
\defnum{pubft-llmw-sr}{0.3\%}
\defnum{pubft-d3conf-sr}{50.0\%}
\defnum{pubft-gain-sr}{$-$0.8}
\defnum{pubft-gain-sr-ci}{[$-$1.9, +0.2]}
\defnum{pubft-agree-r1}{83.9\%}
\defnum{pubft-agree-r1-n}{254}
\defnum{pubft-llmw-r1}{1.6\%}
\defnum{pubft-d3conf-r1}{56.2\%}
\defnum{pubft-gain-r1}{+1.8}
\defnum{pubft-gain-r1-ci}{[+0.8, +2.8]}
\defnum{pubft-gen-r1}{80.0\%}
\defnum{swap-acc-base}{76.1\%}
\defnum{swap-agree-base}{79.6\%}
\defnum{swap-d3conf-base}{54.7\%}
\defnum{swap-acc-ladder}{84.1\%}
\defnum{swap-agree-ladder}{87.5\%}
\defnum{swap-d3conf-ladder}{45.3\%}
\defnum{swap-diff-ladder}{+8.0}
\defnum{swap-diff-ladder-ci}{[+5.9, +10.0]}
\defnum{swap-acc-vits}{72.0\%}
\defnum{swap-agree-vits}{75.4\%}
\defnum{swap-d3conf-vits}{56.2\%}
\defnum{swap-diff-vits}{$-$4.1}
\defnum{swap-diff-vits-ci}{[$-$5.7, $-$2.4]}
\defnum{swap-share-vits}{$-$0.51}
\defnum{swap-share-vits-ci}{[$-$0.86, $-$0.26]}
\defnum{swap-acc-llms}{85.3\%}
\defnum{swap-agree-llms}{88.4\%}
\defnum{swap-d3conf-llms}{39.1\%}
\defnum{swap-diff-llms}{+9.2}
\defnum{swap-diff-llms-ci}{[+7.2, +11.2]}
\defnum{swap-share-llms}{1.15}
\defnum{swap-share-llms-ci}{[0.98, 1.40]}
\defnum{swap-diff-vitsmb}{$-$4.2}
\defnum{swap-diff-llmsms}{+9.3}
\defnum{swap-acc-last4}{75.9\%}
\defnum{swap-agree-last4}{80.0\%}
\defnum{swap-diff-last4}{$-$0.2}
\defnum{swap-diff-last4-ci}{[$-$0.9, +0.6]}
\defnum{swap-share-last4}{$-$0.02}
\defnum{swap-share-last4-ci}{[$-$0.13, 0.08]}
\defnum{swap-repro-n}{9100}
\defnum{swap-repro-same}{100.00\%}
\defnum{optorder-agree-vitw-lr-orig-3b}{87.5\%}
\defnum{optorder-agree-vitw-lr-sw-3b}{86.4\%}
\defnum{optorder-agree-vitw-lr-diff-3b}{$-$1.0}
\defnum{optorder-agree-vitw-lr-diff-3b-ci}{[$-$4.4, +2.2]}
\defnum{optorder-flip-vitw-lr-orig-3b}{83.3\%}
\defnum{optorder-flip-vitw-lr-sw-3b}{85.6\%}
\defnum{optorder-flip-vitw-lr-diff-3b}{+2.3}
\defnum{optorder-flip-vitw-lr-diff-3b-ci}{[$-$2.5, +7.1]}
\defnum{optorder-flip-llmw-lr-orig-3b}{0.5\%}
\defnum{optorder-flip-llmw-lr-sw-3b}{1.4\%}
\defnum{optorder-flip-llmw-lr-diff-3b}{+0.9}
\defnum{optorder-flip-llmw-lr-diff-3b-ci}{[$-$1.0, +2.8]}
\defnum{optorder-flip-vith-ud-orig-3b}{45.5\%}
\defnum{optorder-flip-vith-ud-sw-3b}{58.9\%}
\defnum{optorder-flip-vith-ud-diff-3b}{+13.4}
\defnum{optorder-flip-vith-ud-diff-3b-ci}{[+1.2, +25.8]}
\defnum{optorder-flip-vith-ud-avg-3b}{52.2\%}
\defnum{optorder-flip-llmh-ud-orig-3b}{9.1\%}
\defnum{optorder-flip-llmh-ud-sw-3b}{7.8\%}
\defnum{optorder-flip-llmh-ud-diff-3b}{$-$1.3}
\defnum{optorder-flip-llmh-ud-diff-3b-ci}{[$-$8.6, +6.1]}
\defnum{optorder-flip-vitw-ud-orig-3b}{4.4\%}
\defnum{optorder-flip-vitw-ud-sw-3b}{9.0\%}
\defnum{optorder-flip-vitw-ud-diff-3b}{+4.6}
\defnum{optorder-flip-vitw-ud-diff-3b-ci}{[$-$0.3, +9.8]}
\defnum{optorder-ties-orig-3b}{754}
\defnum{optorder-ties-sw-3b}{721}
\defnum{optorder-flipby-left-sw-3b}{68.7\%}
\defnum{optorder-flipby-right-sw-3b}{100.0\%}
\defnum{optorder-cnt-drop-hflip-3b}{+0.0}
\defnum{optorder-cnt-drop-hflip-3b-ci}{[$-$2.2, +2.2]}
\defnum{optorder-cnt-drop-vitw-3b}{+15.6}
\defnum{optorder-cnt-drop-vitw-3b-ci}{[+12.2, +19.0]}
\defnum{optorder-cnt-dropabs-vitw-3b}{15.6}
\defnum{optorder-cnt-drop-vith-3b}{+22.8}
\defnum{optorder-cnt-drop-vith-3b-ci}{[+19.2, +26.5]}
\defnum{optorder-cnt-dropabs-vith-3b}{22.8}
\defnum{optorder-cnt-drop-llmw-3b}{$-$0.1}
\defnum{optorder-cnt-drop-llmw-3b-ci}{[$-$0.8, +0.4]}
\defnum{optorder-agree-vitw-lr-orig-7b}{93.2\%}
\defnum{optorder-agree-vitw-lr-sw-7b}{89.3\%}
\defnum{optorder-agree-vitw-lr-diff-7b}{$-$3.9}
\defnum{optorder-agree-vitw-lr-diff-7b-ci}{[$-$6.6, $-$1.3]}
\defnum{optorder-flip-vitw-lr-orig-7b}{95.4\%}
\defnum{optorder-flip-vitw-lr-sw-7b}{91.7\%}
\defnum{optorder-flip-vitw-lr-diff-7b}{$-$3.7}
\defnum{optorder-flip-vitw-lr-diff-7b-ci}{[$-$6.3, $-$1.2]}
\defnum{optorder-flip-llmw-lr-orig-7b}{0.3\%}
\defnum{optorder-flip-llmw-lr-sw-7b}{0.6\%}
\defnum{optorder-flip-llmw-lr-diff-7b}{+0.3}
\defnum{optorder-flip-llmw-lr-diff-7b-ci}{[$-$0.6, +1.5]}
\defnum{optorder-flip-vith-ud-orig-7b}{51.4\%}
\defnum{optorder-flip-vith-ud-sw-7b}{54.5\%}
\defnum{optorder-flip-vith-ud-diff-7b}{+3.0}
\defnum{optorder-flip-vith-ud-diff-7b-ci}{[$-$4.4, +10.4]}
\defnum{optorder-flip-llmh-ud-orig-7b}{12.1\%}
\defnum{optorder-flip-llmh-ud-sw-7b}{10.7\%}
\defnum{optorder-flip-llmh-ud-diff-7b}{$-$1.5}
\defnum{optorder-flip-llmh-ud-diff-7b-ci}{[$-$6.5, +3.5]}
\defnum{optorder-flip-vitw-ud-orig-7b}{5.1\%}
\defnum{optorder-flip-vitw-ud-sw-7b}{6.2\%}
\defnum{optorder-flip-vitw-ud-diff-7b}{+1.1}
\defnum{optorder-flip-vitw-ud-diff-7b-ci}{[$-$2.0, +4.4]}
\defnum{optorder-ties-orig-7b}{239}
\defnum{optorder-ties-sw-7b}{258}
\defnum{optorder-flipby-left-sw-7b}{84.5\%}
\defnum{optorder-flipby-right-sw-7b}{98.2\%}
\defnum{optorder-cnt-drop-hflip-7b}{+0.0}
\defnum{optorder-cnt-drop-hflip-7b-ci}{[$-$2.3, +2.2]}
\defnum{optorder-cnt-drop-vitw-7b}{+18.3}
\defnum{optorder-cnt-drop-vitw-7b-ci}{[+14.7, +21.8]}
\defnum{optorder-cnt-dropabs-vitw-7b}{18.3}
\defnum{optorder-cnt-drop-vith-7b}{+25.9}
\defnum{optorder-cnt-drop-vith-7b-ci}{[+22.1, +29.7]}
\defnum{optorder-cnt-dropabs-vith-7b}{25.9}
\defnum{optorder-cnt-drop-llmw-7b}{$-$0.1}
\defnum{optorder-cnt-drop-llmw-7b-ci}{[$-$0.5, +0.3]}
\defnum{optorder-cnt-n}{788}
\defnum{swaperr-prior-ud-7b}{64}
\defnum{swaperr-prior-lr-ivl}{93}
\defnum{emb-anti-3b}{+8.4}
\defnum{emb-anti-3b-ci}{[+4.4, +12.4]}
\defnum{emb-anti-7b}{+8.3}
\defnum{emb-anti-7b-ci}{[+4.6, +12.1]}
\defnum{tiles-s-lr}{73.1\%}
\defnum{tiles-ht-lr}{66.7\%}
\defnum{tiles-swap-lr}{63.0\%}
\defnum{tiles-swap-dm-ci}{[$-$0.901, $-$0.395]}
\defnum{acc-lr-3b}{73.1\%}
\defnum{acc-ud-3b}{74.2\%}
\defnum{acc-lr-7b}{90.1\%}
\defnum{acc-ud-7b}{89.5\%}
\defnum{acc-lr-ivl}{72.1\%}
\defnum{acc-ud-ivl}{78.3\%}
\defnum{wu-acc-lr-3b}{95.1\%}
\defnum{wu-acc-ud-3b}{94.7\%}
\defnum{wu-flip-vitw-lr-3b-n}{192}
\defnum{wu-flip-vith-ud-3b-n}{105}
\defnum{wu-acc-lr-7b}{85.0\%}
\defnum{wu-acc-ud-7b}{98.1\%}
\defnum{wu-flip-vitw-lr-7b-n}{154}
\defnum{wu-flip-vith-ud-7b-n}{99}
\defnum{wu-acc-lr-ivl}{98.1\%}
\defnum{wu-acc-ud-ivl}{92.7\%}
\defnum{wu-flip-vitw-lr-ivl-n}{196}
\defnum{wu-flip-vith-ud-ivl-n}{68}
\defnum{cv-flip-vith-ud-3b-n}{88}
\defnum{cv-flip-vith-lr-3b-n}{241}
\defnum{cv-flip-vith-ud-7b-n}{173}
\defnum{cv-flip-vith-lr-7b-n}{318}
\defnum{cv-flip-vith-ud-ivl-n}{96}
\defnum{cv-flip-vith-lr-ivl-n}{245}
\defnum{order-and-id-w-lr-3b}{2.3\%}
\defnum{id-only-h-ud-3b}{5.2\%}
\defnum{order-and-id-w-lr-7b}{2.3\%}
\defnum{id-only-h-ud-7b}{11.6\%}
\defnum{rowperm-onlyh-3b}{23.9\%}
\defnum{rowperm-rowshift-3b}{15.9\%}
\defnum{rowperm-shuffle-3b}{12.5\%}
\defnum{rowperm-vflip-share-3b}{53.3\%}
\defnum{tie-agree-vitw-3b}{89.9\%}
\defnum{tie-flip-vitw-3b}{88.1\%}
\defnum{tie-flip-llmw-3b}{0.0\%}
\defnum{tie-agree-vitw-ivl}{85.9\%}
\defnum{tie-flip-vitw-ivl}{89.7\%}
\defnum{tie-flip-llmw-ivl}{1.7\%}
\defnum{emb-harm-vith-3b}{49.6\%}
\defnum{emb-harm-vith-7b}{48.4\%}
\defnum{emb-fix-vith-3b}{20.4\%}
\defnum{emb-fix-vith-3b-ci}{[15.5, 26.2]}
\defnum{emb-fix-vitw-3b}{19.4\%}
\defnum{emb-fix-vitw-3b-ci}{[14.7, 25.2]}
\defnum{emb-fix-vith-7b}{22.3\%}
\defnum{emb-fix-vith-7b-ci}{[17.1, 28.5]}
\defnum{emb-fix-vitw-7b}{16.8\%}
\defnum{emb-fix-vitw-7b-ci}{[12.3, 22.6]}
\defnum{cv-agree-vitw-ivl8b}{93.2\%}
\defnum{cv-agree-orig-ivl8b}{8.6\%}
\defnum{cv-agree-rot-ivl8b}{84.1\%}
\defnum{cv-flip-vitw-lr-ivl8b}{95.5\%}
\defnum{cv-flip-vitw-ud-ivl8b}{3.9\%}
\defnum{cv-flip-llmw-lr-ivl8b}{1.2\%}
\defnum{cv-flip-vith-ud-ivl8b}{48.1\%}
\defnum{cv-flip-llmh-ud-ivl8b}{3.1\%}
\defnum{cv-flip-vitw-lr-ivl8b-n}{335}
\defnum{cv-flip-vith-ud-ivl8b-n}{162}
\defnum{leftsplit-flip-left-3b}{62.2\%}
\defnum{leftsplit-flip-right-3b}{100.0\%}
\defnum{leftsplit-vitw-left-3b}{82.2\%}
\defnum{cv-left-orig-3b}{71.0\%}
\defnum{blankclass-ud-real-3b}{27.1\%}
\defnum{leftsplit-flip-left-7b}{92.8\%}
\defnum{leftsplit-flip-right-7b}{97.7\%}
\defnum{leftsplit-vitw-left-7b}{55.6\%}
\defnum{cv-left-orig-7b}{50.4\%}
\defnum{blankclass-ud-real-7b}{53.2\%}
\defnum{leftsplit-flip-left-ivl}{85.5\%}
\defnum{leftsplit-flip-right-ivl}{91.0\%}
\defnum{leftsplit-vitw-left-ivl}{54.6\%}
\defnum{cv-left-orig-ivl}{50.7\%}
\defnum{blankclass-ud-real-ivl}{20.2\%}
\defnum{depth-hw-diff-3b}{+10.2}
\defnum{depth-hw-diff-3b-ci}{[+3.2, +17.2]}
\defnum{depth-hw-diff-7b}{+24.2}
\defnum{depth-hw-diff-7b-ci}{[+17.8, +30.3]}
\defnum{mrope-w-freq-hi}{$1.78\times 10^{-4}$}
\defnum{mrope-w-angle-36}{0.0062}
\defnum{flipback-3b}{42.8\%}
\defnum{flipback-7b}{84.1\%}
\defnum{flipback-ivl}{76.0\%}
\defnum{video-n-vid-7b}{100}
\defnum{cv-agree-vitw-3b-ci}{[84, 90]}
\defnum{cv-agree-orig-3b-ci}{[41, 51]}
\defnum{cv-flip-vitw-lr-3b-ci}{[77, 88]}
\defnum{cv-flip-vitw-ud-3b-ci}{[2, 8]}
\defnum{cv-flip-vitw-ud-3b-n}{182}
\defnum{cv-flip-llmw-lr-3b-ci}{[0, 3]}
\defnum{cv-flip-vith-ud-3b-ci}{[35, 56]}
\defnum{cv-flip-llmh-ud-3b-ci}{[5, 17]}
\defnum{cv-agree-vitw-7b-ci}{[90, 95]}
\defnum{cv-agree-orig-7b-ci}{[6, 12]}
\defnum{cv-flip-vitw-lr-7b-ci}{[93, 97]}
\defnum{cv-flip-vitw-ud-7b-ci}{[3, 9]}
\defnum{cv-flip-vitw-ud-7b-n}{234}
\defnum{cv-flip-llmw-lr-7b-ci}{[0, 2]}
\defnum{cv-flip-vith-ud-7b-ci}{[44, 59]}
\defnum{cv-flip-llmh-ud-7b-ci}{[8, 18]}
\defnum{cv-agree-vitw-ivl-ci}{[80, 88]}
\defnum{cv-agree-orig-ivl-ci}{[16, 24]}
\defnum{cv-flip-vitw-lr-ivl-ci}{[84, 92]}
\defnum{cv-flip-vitw-ud-ivl-ci}{[5, 12]}
\defnum{cv-flip-vitw-ud-ivl-n}{193}
\defnum{cv-flip-llmw-lr-ivl-ci}{[1, 5]}
\defnum{cv-flip-vith-ud-ivl-ci}{[60, 78]}
\defnum{cv-flip-llmh-ud-ivl-ci}{[2, 10]}
\defnum{wu-mirror-vitw-lr-3b-ci}{[76, 87]}
\defnum{wu-mirror-vitw-lr-7b-ci}{[65, 79]}
\defnum{wu-mirror-vitw-lr-ivl-ci}{[91, 97]}
\defnum{depth-test-dir-vith-7b}{95.2\%}
\defnum{depth-test-dir-vitw-7b}{80.0\%}
\defnum{tiles-swap-dm}{$-$0.643}
\defnum{cv-flip-vith-lr-3b}{10.8\%}
\defnum{cv-flip-vith-lr-7b}{8.2\%}
\defnum{cv-flip-vith-lr-ivl}{15.5\%}
\defnum{acc-lr-ivl8b}{93.2\%}
\defnum{acc-ud-ivl8b}{89.5\%}
\defnum{cv-agree-vitw-ivl8b-ci}{[90, 95]}
\defnum{cv-agree-orig-ivl8b-ci}{[6, 12]}
\defnum{cv-flip-vitw-lr-ivl8b-ci}{[93, 97]}
\defnum{cv-flip-vitw-ud-ivl8b-ci}{[2, 7]}
\defnum{cv-flip-llmw-lr-ivl8b-ci}{[0, 3]}
\defnum{cv-flip-vith-ud-ivl8b-ci}{[41, 56]}
\defnum{cv-flip-llmh-ud-ivl8b-ci}{[1, 7]}
\defnum{cv-flip-vitw-ud-ivl8b-n}{231}
\defnum{cv-flip-vith-lr-ivl8b}{12.0\%}
\defnum{cv-flip-vith-lr-ivl8b-n}{334}
\defnum{leak-all-2b}{14.6\%}
\defnum{leak-all-8b}{8.6\%}
\defnum{leak-all-3b}{11.2\%}
\defnum{leak-all-7b}{8.6\%}
\defnum{leak-all-diff}{$-$6.0}
\defnum{leak-all-diff-ci}{[$-$11.2, $-$0.8]}
\defnum{leak-gc-diff}{$-$3.9}
\defnum{leak-gc-diff-ci}{[$-$8.0, +0.1]}
\defnum{leak-all-q3diff}{$-$3.4}
\defnum{leak-all-q3diff-ci}{[$-$9.0, +2.3]}
\defnum{leak-all-q7diff}{$-$6.0}
\defnum{leak-all-q7diff-ci}{[$-$10.8, $-$1.2]}
\defnum{depth-did-did-3bt}{+67.1}
\defnum{depth-did-did-3bt-ci}{[+47.2, +86.6]}
\defnum{depth-did-did-7bt}{+45.0}
\defnum{depth-did-did-7bt-ci}{[+23.8, +65.9]}
\defnum{ivl-angle-ivl-llmw-maxshift}{15}
\defnum{ivl-angle-ivl-llmw-meanang-wrap}{0.32}
\defnum{ties-ivl}{4.5\%}
\defnum{cmp7b8b-orig}{42}
\defnum{cmp7b8b-vitw}{74}
\defnum{cmp7b8b-n}{650}
\defnum{depth-val-vith-3b}{32.5\%}
\defnum{depth-val-vith-3b-ci}{[27.4, 38.1]}
\defnum{depth-val-vitw-3b}{27.6\%}
\defnum{depth-val-vitw-3b-ci}{[22.8, 33.1]}
\defnum{cv-agree-rot-3b-ci}{[78, 86]}
\defnum{cv-agree-rotorig-3b}{48.3\%}
\defnum{cv-agree-rotorig-3b-ci}{[43, 53]}
\defnum{cv-flip-vith-lr-3b-ci}{[7, 15]}
\defnum{cv-agree-rot-7b-ci}{[77, 85]}
\defnum{cv-agree-rotorig-7b}{12.0\%}
\defnum{cv-agree-rotorig-7b-ci}{[9, 16]}
\defnum{cv-flip-vith-lr-7b-ci}{[6, 12]}
\defnum{cv-agree-rot-ivl-ci}{[72, 80]}
\defnum{cv-agree-rotorig-ivl}{26.9\%}
\defnum{cv-agree-rotorig-ivl-ci}{[23, 32]}
\defnum{cv-flip-vith-lr-ivl-ci}{[12, 21]}
\defnum{cv-agree-rot-ivl8b-ci}{[80, 87]}
\defnum{cv-agree-rotorig-ivl8b}{9.4\%}
\defnum{cv-agree-rotorig-ivl8b-ci}{[7, 13]}
\defnum{cv-flip-vith-lr-ivl8b-ci}{[9, 16]}
\defnum{nowidth-lr-3b}{48.0\%}
\defnum{nowidth-lr-7b}{50.1\%}
\defnum{nowidth-lr-ivl}{49.9\%}
\defnum{nowidth-oid-7b}{29.8\%}
\defnum{nowidth-gc-n-7b}{47}
\defnum{nowidth-oid-ivl}{29.5\%}
\defnum{nowidth-gc-n-ivl}{44}
\defnum{ivl-ordonly-w}{2.0\%}
\defnum{ivl-ordonly-w-ci}{[0.9, 4.7]}
\defnum{depth-perm-dir-vith-7b}{95.2\%}
\defnum{depth-perm-dir-perm-7b}{94.0\%}
\defnum{depth-vflipagree-vith-7b}{64.5\%}
\defnum{depth-vflipagree-perm-7b}{71.7\%}
\defnum{perm-ratio-cv-7b}{0.50}
\defnum{perm-ratio-cv-7b-ci}{[0.46, 0.54]}
\defnum{depth-perm-shift-7b}{22.6\%}
\defnum{depth-perm-shuf-7b}{22.1\%}
\defnum{nowidth-oid-7b-ci}{[18.7, 44.0]}
\defnum{nowidth-oid-ivl-ci}{[18.2, 44.2]}
\defnum{emb-split-vith-con-3b}{52.5\%}
\defnum{emb-split-vith-con-3b-ci}{[48.0, 57.0]}
\defnum{emb-split-vith-inc-3b}{47.8\%}
\defnum{emb-split-vith-inc-3b-ci}{[43.8, 51.9]}
\defnum{emb-split-vith-con-7b}{55.6\%}
\defnum{emb-split-vith-con-7b-ci}{[51.3, 59.8]}
\defnum{emb-split-vith-inc-7b}{53.6\%}
\defnum{emb-split-vith-inc-7b-ci}{[49.4, 57.8]}
\defnum{blkdir-butlast-rl-3b}{69}
\defnum{blkdir-butlast-lr-3b}{1}
\defnum{blkdir-blk31-rl-3b}{111}
\defnum{blkdir-blk31-lr-3b}{69}
\defnum{blkdir-butlast-rl-7b}{43}
\defnum{blkdir-butlast-lr-7b}{52}
\defnum{perm-shift-cv-lr-3b}{80.1\%}
\defnum{perm-shift-cv-lr-7b}{80.4\%}
\defnum{perm-shift-cv-lr-ivl}{84.0\%}
\defnum{nowidth-llmw-7b}{6.4\%}
\defnum{nowidth-llmw-7b-ci}{[2.2, 17.2]}
\defnum{nowidth-ordonly-7b}{25.5\%}
\defnum{nowidth-ordonly-7b-ci}{[15.3, 39.5]}
\defnum{nowidth-llmw-ivl}{20.5\%}
\defnum{nowidth-llmw-ivl-ci}{[11.2, 34.5]}
\defnum{nowidth-ordonly-ivl}{29.5\%}
\defnum{nowidth-ordonly-ivl-ci}{[18.2, 44.2]}
\defnum{depth-perm-w-dir-perm-7b}{50.0\%}
\defnum{depth-permsplit-h-follow-3b}{23.1\%}
\defnum{depth-permsplit-h-against-3b}{3.4\%}
\defnum{depth-permsplit-h-follow-7b}{24.1\%}
\defnum{depth-permsplit-h-against-7b}{10.3\%}
\defnum{depth-permsplit-w-follow-7b}{6.5\%}
\defnum{depth-permsplit-w-against-7b}{44.0\%}
\defnum{nowidth-left-3b}{99.2\%}
\defnum{bx-n-3b}{140}
\defnum{bx-hflip-vitw-3b}{56.4\%}
\defnum{bx-hflip-vitw-3b-ci}{[48.2, 64.4]}
\defnum{bx-hflip-orig-3b}{5.7\%}
\defnum{bx-hflip-llmw-3b}{5.7\%}
\defnum{bx-n-7b}{131}
\defnum{bx-hflip-vitw-7b}{57.3\%}
\defnum{bx-hflip-vitw-7b-ci}{[48.7, 65.4]}
\defnum{bx-hflip-orig-7b}{3.8\%}
\defnum{bx-hflip-llmw-7b}{3.8\%}
\defnum{bx-other-7b}{32.8\%}
\defnum{objorder-rev-3b}{71.6\%}
\defnum{objorder-keep-3b}{44.0\%}
\defnum{objorder-diff-3b}{+27.6}
\defnum{objorder-diff-3b-ci}{[18.4, 36.4]}
\defnum{objorder-c33-3b}{5}
\defnum{objorder-c33n-3b}{40}
\defnum{objorder-rev-7b}{71.6\%}
\defnum{objorder-keep-7b}{32.5\%}
\defnum{objorder-diff-7b}{+39.0}
\defnum{objorder-diff-7b-ci}{[33.0, 45.0]}
\defnum{objorder-c33-7b}{7}
\defnum{objorder-c33n-7b}{24}
\defnum{objorder-rev-ivl}{74.1\%}
\defnum{objorder-keep-ivl}{50.2\%}
\defnum{objorder-diff-ivl}{+23.9}
\defnum{objorder-diff-ivl-ci}{[15.8, 31.7]}
\defnum{objorder-c33-ivl}{3}
\defnum{objorder-c33n-ivl}{27}
\defnum{depth-conflict-n-3b}{29}
\defnum{depth-conflict-vith-3b}{12}
\defnum{depth-conflict-vitw-3b}{15}
\defnum{depth-matched-match-3b}{+42.6}
\defnum{depth-matched-match-3b-ci}{[24.7, 60.5]}
\defnum{depth-follow-follow-all-3b}{81.6\%}
\defnum{depth-follow-diff-all-3b}{$-$3.3}
\defnum{depth-follow-diff-all-3b-ci}{[$-$7.3, 0.5]}
\defnum{depth-conflict-n-7b}{20}
\defnum{depth-conflict-vith-7b}{11}
\defnum{depth-conflict-vitw-7b}{6}
\defnum{depth-conflict-diff-7b}{+25.0}
\defnum{depth-conflict-diff-7b-ci}{[0.0, 50.0]}
\defnum{depth-matched-match-7b}{+52.6}
\defnum{depth-matched-match-7b-ci}{[28.4, 75.5]}
\defnum{depth-matched-match-h-7b}{18.2\%}
\defnum{depth-matched-match-w-7b}{13.8\%}
\defnum{depth-follow-follow-all-7b}{84.9\%}
\defnum{depth-follow-diff-all-7b}{+0.0}
\defnum{depth-follow-diff-all-7b-ci}{[$-$2.8, 2.8]}
\defnum{depth-follow-base}{84.9\%}
\defnum{depth-halves-test-n}{314}
\defnum{depth-halves-shared-img}{35}
\defnum{depth-halves-shared-test}{45}
\defnum{depth-halves-shared-test-counter}{1}
\defnum{conly-wu-gcv-3b}{71.4\%}
\defnum{conly-wu-gcv-7b}{60.6\%}
\defnum{conly-wu-gcv-ivl}{66.2\%}
\defnum{conly-cv-fbv-3b}{38.2\%}
\defnum{conly-cv-fbh-3b}{42.8\%}
\defnum{conly-cv-gcv-3b}{80.7\%}
\defnum{conly-cv-gch-3b}{77.4\%}
\defnum{conly-cv-fbv-7b}{58.8\%}
\defnum{conly-cv-fbh-7b}{84.1\%}
\defnum{conly-cv-gcv-7b}{80.9\%}
\defnum{conly-cv-gch-7b}{90.8\%}
\defnum{conly-cv-fbv-ivl}{38.2\%}
\defnum{conly-cv-fbh-ivl}{76.0\%}
\defnum{conly-cv-gcv-ivl}{80.2\%}
\defnum{conly-cv-gch-ivl}{91.0\%}
\defnum{blkset-b31-3b}{86.6\%}
\defnum{blkset-all-3b}{83.3\%}
\defnum{blkset-nonglobal-3b}{18.3\%}
\defnum{blkset-n-3b}{186}
\defnum{blkset-b31-7b}{72.5\%}
\defnum{blkset-g4-7b}{95.4\%}
\defnum{blkset-all-7b}{95.4\%}
\defnum{blkset-g3early-7b}{18.7\%}
\defnum{blkset-win2430-7b}{0.9\%}
\defnum{blkset-last8-7b}{81.0\%}
\defnum{blkset-nonglobal-7b}{6.7\%}
\defnum{blkset-gratio-7b}{1.00}
\defnum{blkset-gratio-7b-ci}{[0.97, 1.03]}
\defnum{blkset-l8b31-7b}{8.6}
\defnum{blkset-l8b31-7b-ci}{[4.9, 12.2]}
\defnum{blkset-n-7b}{327}
\defnum{tiling-gc-n}{246}
\defnum{tiling-vitwtile}{86.2\%}
\defnum{tiling-vitwtile-ci}{[81.3, 89.9]}
\defnum{tiling-llmwimg}{2.8\%}
\defnum{tiling-llmwimg-ci}{[1.4, 5.8]}
\defnum{tiling-orderonlywimg}{2.8\%}
\defnum{tiling-orderonlywimg-ci}{[1.4, 5.8]}
\defnum{ivl-optorder-agreevitwhfliplr-orig}{84.3\%}
\defnum{ivl-optorder-agreevitwhfliplr-sw}{88.5\%}
\defnum{ivl-optorder-agreevitwhfliplr-diff}{+4.2}
\defnum{ivl-optorder-flipvitwlr-orig}{88.5\%}
\defnum{ivl-optorder-flipvitwlr-sw}{92.8\%}
\defnum{ivl-optorder-flipvitwlr-diff}{+4.2}
\defnum{ivl-optorder-flipllmwlr-orig}{2.5\%}
\defnum{ivl-optorder-flipllmwlr-sw}{2.6\%}
\defnum{ivl-optorder-flipllmwlr-diff}{+0.1}
\defnum{ivl-optorder-flipvithud-orig}{69.8\%}
\defnum{ivl-optorder-flipvithud-sw}{66.4\%}
\defnum{ivl-optorder-flipvithud-diff}{$-$3.4}
\defnum{ivl-optorder-flipllmhud-orig}{4.2\%}
\defnum{ivl-optorder-flipllmhud-sw}{12.6\%}
\defnum{ivl-optorder-flipllmhud-diff}{+8.4}
\defnum{ivl-optorder-flipvitwud-orig}{13.5\%}
\defnum{ivl-optorder-flipvitwud-sw}{16.0\%}
\defnum{ivl-optorder-flipvitwud-diff}{+2.4}
\defnum{qwen3-cv-gc-n}{344}
\defnum{qwen3-wu-gc-n}{203}
\defnum{qwen3-cv-vitw}{93.9\%}
\defnum{qwen3-cv-vitw-ci}{[90.8, 96.0]}
\defnum{qwen3-cv-llmw}{2.0\%}
\defnum{qwen3-cv-llmw-ci}{[1.0, 4.1]}
\defnum{qwen3-cv-ordonly}{0.3\%}
\defnum{qwen3-cv-ordonly-ci}{[0.1, 1.6]}
\defnum{qwen3-wu-vitw}{99.5\%}
\defnum{qwen3-wu-vitw-ci}{[97.3, 99.9]}
\defnum{qwen3-wu-llmw}{1.5\%}
\defnum{qwen3-wu-llmw-ci}{[0.5, 4.3]}
\defnum{qwen3-wu-ordonly}{0.0\%}
\defnum{qwen3-wu-ordonly-ci}{[0.0, 1.9]}
\defnum{qwen3-cv-agree}{89.6\%}
\defnum{qwen3-cv-agree-orig}{6.3\%}
\defnum{qwen3-cv-vith-ud}{51.5\%}
\defnum{qwen3-cv-llmh-ud}{1.0\%}
\defnum{qwen3-wu-agree}{90.3\%}
\defnum{objorder-mirror-3b}{83.8\%}
\defnum{objorder-mirror-7b}{96.3\%}
\defnum{objorder-mirror-ivl}{87.1\%}
\defnum{objorder-coco-n}{359}
\defnum{objorder-coco-ok}{347}
\defnum{objorder-n-3b}{117}
\defnum{objorder-n-7b}{191}
\defnum{objorder-n-ivl}{140}
\defnum{perm-shift-cv-lr-qwen3}{77.6\%}
\defnum{perm-mmax-cv-qwen3}{16.3}
\defnum{perm-mmax-cv-qwen3-ci}{[11.9, 20.9]}
\defnum{perm-ratio-cv-qwen3}{0.30}
\defnum{perm-ratio-cv-qwen3-ci}{[0.27, 0.34]}
\defnum{objorder-n-qwen3}{202}
\defnum{objorder-rev-qwen3}{60.3\%}
\defnum{objorder-keep-qwen3}{19.3\%}
\defnum{objorder-diff-qwen3}{+41.0}
\defnum{objorder-diff-qwen3-ci}{[35.2, 46.9]}
\defnum{objorder-ambig-qwen3}{4}
\defnum{objorder-mirror-qwen3}{93.6\%}
\defnum{objorder-offaxis-qwen3}{2.8\%}
\defnum{objorder-c33-qwen3}{1}
\defnum{objorder-c33n-qwen3}{5}
\defnum{perm-mrel-wu-mirror-qwen3}{91.1\%}
\defnum{perm-mrel-wu-shift-qwen3}{83.1\%}
\defnum{qwen3-cv-acc-lr}{93.2\%}
\defnum{qwen3-cv-ud-n}{196}
\defnum{qwen3-cv-vitw-ud}{3.1\%}
\defnum{qwen3-wu-acc-lr}{99.0\%}
\defnum{perm-mmax-wu-qwen3}{3.4}
\defnum{perm-mmax-wu-qwen3-ci}{[1.0, 6.4]}
\defnum{objorder-shift-diff-3b}{+26.6}
\defnum{objorder-shift-diff-3b-ci}{[5.5, 49.4]}
\defnum{objorder-shuf-diff-3b}{+18.9}
\defnum{objorder-shuf-diff-3b-ci}{[8.7, 28.8]}
\defnum{objorder-shift-diff-7b}{+46.2}
\defnum{objorder-shift-diff-7b-ci}{[29.7, 62.0]}
\defnum{objorder-shuf-diff-7b}{+31.0}
\defnum{objorder-shuf-diff-7b-ci}{[23.4, 38.8]}
\defnum{objorder-shift-diff-ivl}{$-$14.2}
\defnum{objorder-shift-diff-ivl-ci}{[$-$23.3, $-$4.1]}
\defnum{objorder-shuf-diff-ivl}{+26.1}
\defnum{objorder-shuf-diff-ivl-ci}{[16.9, 34.9]}
\defnum{objorder-shuf-keep-3b}{42.1\%}
\defnum{objorder-shuf-keep-7b}{31.2\%}
\defnum{objorder-shuf-keep-ivl}{43.5\%}
\defnum{objorder-shuf-keep-qwen3}{16.6\%}
\defnum{objorder-shift-rev-ivl}{82.6\%}
\defnum{objorder-shift-keep-ivl}{96.8\%}
\defnum{objorder-shift-diff-qwen3}{+50.1}
\defnum{objorder-shift-diff-qwen3-ci}{[35.2, 64.2]}
\defnum{objorder-shuf-diff-qwen3}{+28.0}
\defnum{objorder-shuf-diff-qwen3-ci}{[20.7, 35.2]}
\defnum{perm-mmax-cv-3b}{3.2}
\defnum{perm-mmax-cv-3b-ci}{[$-$3.2, 10.2]}
\defnum{hperm-mmax-cv-3b}{10.2}
\defnum{hperm-mmax-cv-3b-ci}{[$-$1.1, 21.2]}
\defnum{perm-mmax-wu-3b}{5.7}
\defnum{perm-mmax-wu-3b-ci}{[2.6, 9.4]}
\defnum{perm-mmax-cv-7b}{15.0}
\defnum{perm-mmax-cv-7b-ci}{[10.4, 19.6]}
\defnum{hperm-mmax-cv-7b}{20.2}
\defnum{hperm-mmax-cv-7b-ci}{[12.7, 27.7]}
\defnum{perm-mmax-wu-7b}{0.6}
\defnum{perm-mmax-wu-7b-ci}{[0.0, 1.9]}
\defnum{perm-mmax-cv-ivl}{4.5}
\defnum{perm-mmax-cv-ivl-ci}{[$-$0.8, 9.8]}
\defnum{hperm-mmax-cv-ivl}{26.7}
\defnum{hperm-mmax-cv-ivl-ci}{[16.0, 36.1]}
\defnum{perm-mmax-wu-ivl}{$-$0.5}
\defnum{perm-mmax-wu-ivl-ci}{[$-$2.6, 1.0]}
\defnum{udcontent-dis-d-3b}{23.5}
\defnum{udcontent-dis-d-3b-ci}{[15.4, 31.9]}
\defnum{udcontent-dis-d-7b}{32.8}
\defnum{udcontent-dis-d-7b-ci}{[23.0, 43.0]}
\defnum{udcontent-dis-d-ivl}{12.3}
\defnum{udcontent-dis-d-ivl-ci}{[5.5, 19.4]}
\defnum{udcontent-gc-lr-3b}{81.2\%}
\defnum{udcontent-gc-d-3b}{$-$9.8}
\defnum{udcontent-gc-d-3b-ci}{[$-$20.4, 0.4]}
\defnum{blankerr-lr-3b}{93.2\%}
\defnum{blankcorr-lr-3b}{62.5\%}
\defnum{udcontent-gc-lr-7b}{66.2\%}
\defnum{udcontent-gc-d-7b}{$-$5.6}
\defnum{udcontent-gc-d-7b-ci}{[$-$18.0, 6.4]}
\defnum{udcontent-gc-lroff-7b}{33.8\%}
\defnum{blankerr-lr-7b}{65.8\%}
\defnum{blankcorr-lr-7b}{49.9\%}
\defnum{udcontent-gc-lr-ivl}{91.3\%}
\defnum{udcontent-gc-d-ivl}{$-$25.2}
\defnum{udcontent-gc-d-ivl-ci}{[$-$37.4, $-$13.4]}
\defnum{blankerr-all}{64.3\%}
\defnum{blankerr-all-ci}{[59.5, 68.8]}
\defnum{qwen3-flipback}{91.4\%}
\defnum{qwen3-cv-vitw-left}{95.7\%}
\defnum{qwen3-cv-vitw-left-n}{163}
\defnum{qwen3-cv-vitw-right}{92.3\%}
\defnum{qwen3-cv-vitw-right-n}{181}
\defnum{conly-wu-fbv-range}{36.4--64.6\%}
\defnum{conly-wu-fbh-range}{91.7--97.6\%}
\defnum{conly-wu-d-range}{29.6--55.3}
\defnum{conly-wu-gcv-range}{61--71\%}
\defnum{objorder-keep-range}{19--50\%}
\defnum{objorder-offaxis-core-range}{2--9\%}
\defnum{objorder-cover-range}{57--63\%}
\defnum{udcontent-same-ud-range}{28--50\%}
\defnum{udcontent-same-lr-range}{0--4\%}
\defnum{udcontent-dis-d-range}{12--33}
\defnum{udcontent-gc-udreal-range}{18--37\%}
\defnum{udcontent-gc-lrreal-range}{0--5\%}
\defnum{objorder-mirror-range}{84--96\%}
\defnum{qwen3-cv-agree-ci}{[86, 92]}
\defnum{qwen3-cv-agree-orig-ci}{[4, 9]}
\defnum{qwen3-cv-vitw-cii}{[91, 96]}
\defnum{qwen3-cv-llmw-cii}{[1, 4]}
\defnum{qwen3-cv-vitw-ud-ci}{[1, 7]}
\defnum{qwen3-cv-vith-ud-ci}{[45, 58]}
\defnum{qwen3-cv-llmh-ud-ci}{[0, 4]}
\defnum{qwen3-cv-tie-k}{14}
\defnum{qwen3-wu-mirror}{90.6\%}
\defnum{qwen3-wu-mirror-ci}{[86, 94]}
\defnum{objorder-shuf-keep-range}{16.6--43.5\%}
\defnum{objorder-shuf-diff-range}{18.9--31.0}
\defnum{hperm-mmax-wu-range}{33.3--57.5}
\defnum{llmw-lr-max}{2.8\%}
\defnum{tie-wu-vitw-range}{99.0--100.0\%}
\defnum{tie-wu-llmw-range}{0.0--1.0\%}
\defnum{perm-shift-cv-lr-range}{77.6--84.0\%}
\defnum{cv-flip-llmw-lr-range4}{0.3--2.5\%}
\defnum{cv-flip-vitw-lr-range4}{83.3--95.4\%}
\defnum{objorder-rev-range}{60--74\%}
\defnum{perm-mrel-wu-shift-range}{67--97\%}
\defnum{perm-mrel-wu-mirror-range}{72--96\%}
\defnum{perm-mrel-wu-shuf-q-range}{3--39\%}
\defnum{perm-mrel-wu-shuf-qwen3-range}{3.0--12.0\%}
\defnum{perm-mrel-wu-mirror-q-range}{72--91\%}
\defnum{hperm-mrel-wu-shuf-q-range}{86--100\%}
\defnum{hperm-mrel-wu-shuf-ivl-range}{70--86\%}
\defnum{perm-mrel-wu-shuf-ivl-range}{40--84\%}
\defnum{perm-mrel-wu-mirror-ivl-r}{96\%}
\defnum{depth-heur-base-7bt}{87.1\%}
\defnum{nowidth-lr-range}{48.0--50.1\%}
\defnum{nowidth-ud-maxdiff}{5.6}
\defnum{perm-ratio-other-range}{0.30--0.82}
\defnum{video-vid-llmw-7b-hi}{3.7\%}
\defnum{wu-flip-llmw-lr-hi-range}{2--4\%}
\defnum{depth-spec-did-range}{45--67}
\defnum{cv-agree-vitw-range}{84.3--93.2\%}
\defnum{cv-agree-vitw-head-range}{84.3--93.2\%}
\defnum{cv-agree-orig-range}{8.6--46.0\%}
\defnum{cv-flip-vitw-lr-range}{83.3--95.4\%}
\defnum{cv-flip-vitw-ud-range}{4.4--7.8\%}
\defnum{cv-flip-llmw-lr-range}{0.3--2.5\%}
\defnum{cv-flip-vith-ud-range-avg}{51.4--69.8\%}
\defnum{cv-flip-llmh-ud-range}{4.2--12.1\%}
\defnum{cv-agree-rot-range}{76.2--82.2\%}
\defnum{order-only-w-lr-range}{2.3--2.9\%}
\defnum{order-only-h-lr-range}{3.7--4.4\%}
\defnum{colmajor-lr-range}{3.1--3.9\%}
\defnum{order-only-h-ud-range}{16.9--18.7\%}
\defnum{wu-flip-vitw-lr-range}{99.0--100.0\%}
\defnum{wu-mirror-vitw-lr-range}{72.1--94.9\%}
\defnum{wu-flip-llmw-lr-range}{0.0--1.0\%}
\defnum{wu-flip-vith-ud-range}{77.9--83.8\%}
\defnum{wu-flip-vitw-ud-range}{11.1--19.1\%}
\defnum{wu-nonmirror-vitw-3b}{17.7\%}
\defnum{wu-nonmirror-vitw-7b}{27.9\%}
\defnum{blk-glob-other-3b-range}{2.1--6.3\%}
\defnum{blk-glob-other-7b-range}{1.8--4.2\%}
\defnum{pubft-agree-ft-range}{77.3--87.5\%}
\defnum{pubft-llmw-max}{1.6\%}
\defnum{blk-late-3b-range}{1.0--3.4\%}
\defnum{blk-late-7b-range}{0.0--1.8\%}
\defnum{wu-flip-vith-lr-range}{37.8--60.4\%}
\defnum{wu-offaxis-swap-ivl-pct}{16.8\%}
\defnum{blk-ratio-last-7b}{76\%}
\defnum{blk-ratio-last-3b}{101\%}
\defnum{blk-ratio-butlast-range}{28--39\%}
\defnum{blank-lr-pos-range}{77.0--89.3\%}
\defnum{blank-ud-pos-range}{38.2--42.7\%}
\defnum{words-rev2-range}{32--34\%}
\defnum{words-ctl2-range}{3--4\%}
\defnum{swaperr-prior-range}{64--93\%}
\defnum{tie-cv-3b}{6.1\%}
\defnum{tie-cv-7b}{1.9\%}
\defnum{cv-val-n}{187}
\defnum{cv-split-n}{650}
\defnum{depth-spec-gap-range}{25--36}

\journal{Neurocomputing}

\begin{document}

\begin{frontmatter}

\title{Where Do Vision-Language Models Get Left and Right? Position-Code Counterfactuals in Qwen2.5-VL, Qwen3-VL and InternVL3}

\author[aff1]{Enhao Cui\,\orcidlink{0009-0004-1776-4047}\corref{cor1}}
\ead{2500160612@smail.cczu.edu.cn}
\author[aff1]{Junzhi Yang}
\author[aff1]{Yaoshun Lin}
\author[aff1]{Yuhan Feng}
\author[aff1]{Zijian Chen}
\cortext[cor1]{Corresponding author.}
\affiliation[aff1]{organization={Changzhou University},
            city={Changzhou},
            postcode={213164},
            state={Jiangsu},
            country={China}}

\begin{abstract}
A vision-language model may take left--right direction from the orientation of image content within patches, its vision transformer's (ViT) position code, or token positions and order in its language model (LLM). Ablations show the ViT code is necessary; we test sufficiency by mirroring only a position code. In Qwen2.5-VL-7B, Qwen3-VL-4B and InternVL3-2B, mirroring ViT width positions reproduces the flipped-image answer on \nb{cv-agree-vitw-head-range} of CV-Bench left--right questions (Qwen2.5-VL-3B: \nb{cv-agree-vitw-3b}, partly a \emph{left} fallback). With two options, this also counts disruption; a half-grid shift flips most answers too. With the ViT code intact, LLM-side id and token-order mirrors change few left--right answers in all three series (a weak id test in Qwen2.5-VL). In Qwen2.5-VL-7B the effect sits in the global-attention ViT blocks, mostly the last. On What'sUp, flipping content and token order, not the position code, keeps more \emph{on}/\emph{under} than left--right answers at the flipped-image answer. Post hoc analyses qualify these results. On CV-Bench items with located objects, random permutations flip \nb{objorder-shuf-diff-range} points more answers when reversing the objects' order than when keeping it. On What'sUp, in Qwen2.5-VL and Qwen3-VL, the share of the left--right changes made by random permutations that land on the mirror relation is mostly below chance. Counted as returns to the original answer, the What'sUp contrast of support with left--right relations is clear only in InternVL3-2B. Most remaining errors give the same relation whichever object is asked about, suggesting that repair may target relation binding, object localization or use of the reference object.

\end{abstract}


\end{frontmatter}

\section{Introduction}
\label{sec:intro}

Asked whether a cup is to the left of a plate, a vision-language model (VLM) has to turn the layout of an image into a word. Current VLMs often get such questions wrong~\cite{kamath2023whatsup,tong2024cambrian}, their answers about up and down can follow priors more than the image~\cite{ax2606.31257}, and near--far answers can follow statistical shortcuts such as height in the image~\cite{ax2605.30161}. Diagnosing and repairing these failures requires knowing where the directional information that the model uses comes from. A VLM offers three routes: the content of the image patches, the position code that the vision transformer (ViT) attaches to each patch, and the position ids and order of the visual tokens in the language model (LLM).

Prior work shows that the ViT position code is necessary for spatial answers, or correlated with them. Ablating or shuffling position encodings shows that spatial understanding relies more on the vision encoder's position encoding than on the LLM's~\cite{ax2509.02359}, and the horizontal component of ViT attention is associated with left--right judgments~\cite{ax2509.19191}. Patching visual-token content or activations from a counterfactual or mirrored image flips left--right answers by editing what the tokens contain~\cite{ax2603.22278,ax2601.12626}. None of these studies compares the answer after an edit of the position code alone, item by item, with the answer to a real flip (\cref{sec:related}). In LLaVA-1.5, shuffling the visual tokens barely changes CV-Bench accuracy but sharply lowers it on a synthetic benchmark without semantic shortcuts~\cite{ax2503.17349}, so the LLM can depend on token positions. That it reads left and right from its own position ids or from the token order, which such shuffles do not separate, has not been ruled out with structured edits. A real image flip cannot settle this, because it changes all three routes at once. %

We answer the question with position-only counterfactuals. Leaving every pixel unchanged, we mirror the position information along one axis, in the ViT or in the LLM, and compare the answer with the one the model gives for the flipped image (\cref{fig:schematic}). If mirroring only the ViT width positions makes the model answer as if the image were flipped, the ViT position code is \emph{sufficient} to change the answer. With four options (What'sUp), the share of changes that go to the mirror relation, rather than to an off-axis option, shows that the change goes to the opposite side; a half-grid shift of the code reaches a similar share. We treat sufficiency as a graded quantity: the share of questions on which the edit alone reproduces the answer to the real flip. With two options this share also contains changes that any disturbance of the code would cause, so we compare the mirror with non-mirror permutations of the same code. Part of the outcome can be expected from the architecture. Apart from its position code, a ViT processes each patch in nearly the same way wherever the patch lies, so mirroring the width positions comes close to moving every patch to its mirrored cell without flipping its content. Only the patch content, the window partition and the merge order stay unmirrored, and a content-only control tests the first. That the edit changes left--right answers is thus expected; only the intervention shows how much of the answer changes with the code. The design tests the LLM-side ids and the token order separately and applies to rotary and learned absolute position embeddings.

Our headline results use one model from each of the three series, Qwen2.5-VL-7B, Qwen3-VL-4B and InternVL3-2B. Qwen2.5-VL-3B, Qwen2.5-VL-7B and InternVL3-2B form the core models of the main runs, and InternVL3-8B serves as a scale check. The left--right permutation controls were run on the three headline models and on Qwen2.5-VL-3B. Two of the results below could not be read off the architecture: the dominance of the last ViT block and a contrast between \emph{on}/\emph{under} and left--right answers under a content-only flip. Our contributions are:
\begin{itemize}
\item A training-free probe of the position code of VLMs (`probe' here means this intervention, not a probing classifier). Its position-only counterfactuals mirror or reorder positions per axis, in the ViT or on the LLM side, with pixels fixed and a real flip as reference (\cref{sec:method}). With it, mirroring only the ViT width positions reproduces the answer to a real horizontal flip on most CV-Bench left--right questions of Qwen2.5-VL-7B, Qwen3-VL-4B and InternVL3-2B (\nb{cv-agree-vitw-7b} in Qwen2.5-VL-7B). In Qwen2.5-VL-3B part of this rate is a fallback to its \emph{left} prior. Non-mirror permutations also flip many answers, a half-grid shift most of them. Post hoc, we located the two objects in \nb{objorder-cover-range} of the geometry-consistent CV-Bench items. On these items, random permutations flip more answers when they reverse the order of the objects than when they keep it, but order-preserving ones still flip many. On CV-Bench the answer thus tends to follow the objects' order in the ViT code and is also disturbed by the permutation itself (\cref{sec:res:lr}).
\item A decomposition of the LLM side. With the ViT position code intact, the LLM-side ids and token order add no detectable effect on the left--right direction in the three model series, as a serial architecture makes plausible. Mirroring the LLM-side width ids changes at most \nb{llmw-lr-max} of the geometry-consistent left--right answers (correct on both the original and the flipped image) in every model, option order and tiling we tested, including InternVL3 under its default dynamic tiling. Reordering the visual tokens changes them no more than an off-axis control does. The id test is weak by construction in Qwen2.5-VL; the token order was tested in all three series. With the ViT width code removed, left--right accuracy falls to near chance, and a weak LLM-side contribution is not excluded. In one public fine-tuned model, SpatialLadder, a component swap attributes the accuracy gain to the LLM (\cref{sec:res:llm,sec:res:robust}).
\item A localization of the effect in Qwen2.5-VL-7B to the four global-attention ViT blocks, chiefly the last. The small share of the windowed blocks, which see positions only within their windows, is expected, but the dominance of the last block is not (\cref{sec:res:blocks}).
\item Up--down, support and depth answers. Up--down answers change under the ViT height mirror to a degree that varies with benchmark and model. On What'sUp, flipping the content and the token rows vertically, with the position code unchanged, leaves more \emph{on} and \emph{under} answers at the flipped-image answer than the horizontal version leaves left--right answers. This contrasts support with projective relations, and it counts changes to off-axis options; counted only as returns to the original answer (post hoc), it is clear only in InternVL3-2B. On CV-Bench 3D, the use by Qwen2.5-VL-7B of the cue that higher in the image means farther depends on the intact height code. Random row permutations of that code also push answers away from the cue, however, so the direction of the change is not specific to the mirror. In the 3B model the effect is weaker, and on EmbSpatial the dependence does not replicate (\cref{sec:res:ud,sec:res:depth}).
\end{itemize}

\Cref{sec:related} reviews related work, \cref{sec:method} defines the counterfactuals, \cref{sec:results} reports the results, and \cref{sec:discussion} discusses the errors, related mechanistic accounts and the limitations.

\section{Related work}
\label{sec:related}

\paragraph{Spatial relations and priors}
Benchmarks such as VSR~\cite{liu2023vsr}, What'sUp~\cite{kamath2023whatsup}, CV-Bench~\cite{tong2024cambrian} and EmbSpatial-Bench~\cite{du2024embspatial} document that VLMs often fail at basic spatial relations. With a blank-image control, horizontal relations were found to be grounded in the image in every capable model among 14 VLMs, whereas vertical relations relied largely on priors; the depth axis, though decodable, was deployed with the wrong sign~\cite{ax2606.31257}. Many spatial questions can be answered without the image at all~\cite{ax2607.27069}. Depth judgments follow pictorial cues such as `higher in the image means farther'~\cite{ax2605.30161}, and controlling these cues in the stimulus shows which one is used~\cite{ax2607.01503}. Depth order can even remain linearly decodable from the queried objects' visual tokens in the LLM when the answer follows a misleading cue, and writing the decoded direction back into these tokens shifts the answers~\cite{ax2610.17976}. Such cues are often valid, and following them becomes a shortcut~\cite{geirhos2020shortcut} only where the image contradicts them. VLMs also default to the camera's viewpoint when the reference frame is ambiguous~\cite{ax2410.17385,ax2607.17657}. %

\paragraph{Position encodings in VLMs}
Vision encoders mark patch positions with learned absolute embeddings~\cite{dosovitskiy2021vit} or with two-dimensional RoPE~\cite{su2024roformer,bai2025qwen25vl}. In ViT classifiers, per-axis interventions show that RoPE and absolute embeddings store position differently~\cite{ax2610.13992}. Positional embeddings can also help or hurt spatial reasoning~\cite{ax2601.22231}. On the LLM side, multimodal RoPE assigns separate position components to time, height and width~\cite{wang2024qwen2vl}, and its design has been compared systematically~\cite{ax2510.23095}. The closest work masks, shuffles or fixes these encodings per axis, in the ViT and on the LLM side of Qwen2.5-VL-7B and two other VLMs. It finds that spatial understanding relies more on the vision encoder's encodings than on the LLM's~\cite{ax2509.02359}. Its ablations of the LLM-side $x$ ids also change some \emph{left} answers, but an ablation is not a mirror (\cref{sec:res:llm}). On What'sUp, ablating the ViT width encoding of Qwen2.5-VL-7B collapses its \emph{left} and \emph{right} answers, so the ViT's position encoding is necessary for them; one of these ablations is a random permutation of the ViT width ids. The changed answers are not compared item by item with a real image transform, and where they go is not reported. In our post hoc split on CV-Bench, such random permutations flip more answers when they reverse the order of the two objects than when they keep it (\ref{app:protocols}). On procedurally generated scenes, Liu and Chen~\cite{liu2026wherevision} find a double dissociation of the two axes when they ablate the encoder's width or height positional channel. They also trace the answer through image-token states at an early decoder stage. On What'sUp and ARO they extend this mediation with mirror-flip counterfactuals. We mirror the position code instead of ablating it, keep the pixels fixed, and compare the answer item by item with a real flip. When high-resolution inputs degrade the visual position encodings, shuffling them shifts coordinate predictions consistently toward the model's spatial priors~\cite{tao2026vpsg}. Another study~\cite{ax2509.19191} decomposes ViT attention under 2D-RoPE into horizontal and vertical components and associates the horizontal one with left--right judgments. Its embedding-level interventions replace one object's embedding with another's or mix direction vectors in the image embeddings after the modality connector, and measure the probability of the correct answer or the drop in performance. Removing 2D-RoPE serves only to show that the direction vectors lose their geometry, and rescaling the rotary frequencies of the vision encoder mostly raises benchmark accuracy. None of these interventions mirrors one axis of the position code or compares the answer item by item with a real flip. In CLIP-style encoders trained on synthetic one-dimensional scenes, the interaction of position and token embeddings creates a horizontal attention gradient that breaks left--right symmetry, and ablating it weakens left--right discrimination~\cite{ax2601.12809}. In LLaVA-1.5, randomly permuting the visual tokens barely changes CV-Bench accuracy but lowers accuracy on a synthetic benchmark without semantic shortcuts, so standard benchmarks can mask such position sensitivity~\cite{ax2503.17349}. Removing visual tokens whose RoPE positions act as implicit coordinates, or merging tokens and re-assigning their position ids, lowers visual-grounding accuracy~\cite{ax2602.03060}. We test left and right with structured mirrors in pretrained VLMs of three series and separate the LLM-side ids from the token order (token shuffling versus mirrors: \cref{sec:limitations}). %

\paragraph{Mechanistic analyses of spatial information}
Our counterfactuals follow the logic of activation patching and interchange interventions~\cite{geiger2021causal}. Patching a value from a corrupted run into the clean run (noising) tests whether the variable is necessary, whereas patching the clean value into the corrupted run (denoising) tests whether it is sufficient~\cite{ax2404.15255}. A position-only mirror gives each patch the position id it has in the flipped image. With the flipped-image run as the clean run, it is a denoising patch of the assignment of positions to content; because the variable is an input-side id, its counterfactual value is known exactly and needs no second forward pass. Three analyses that intervene on content or activations also use a flipped or swapped image as the counterfactual. Patching content embeddings locates spatial information mainly in the vision encoder, spread over all visual tokens, and swapping the left and right regions of the visual-token embeddings with those of a counterfactual image flips predictions~\cite{ax2603.22278}. In the same work, a secondary, language-side ordering representation can partly act as a backup when the vision-derived ordering is removed. Inside the LLM, linear spatial identifiers in object-word activations can be edited to flip left--right answers, and activations from the run on a mirrored image, patched into the original run, also flip them~\cite{ax2601.12626}. Patching image, query-text or final-token states between an image and its horizontal flip, on items answered correctly in both, shows in Qwen3-VL-4B that spatial relations are read from the query text tokens~\cite{ax2607.03358}. The first two link these signals to position encoding. The first traces band-like patterns in the visual embeddings to the encoder's positional encoding; in the second, a low-rank linear map from positional encodings explains most of the variance of the spatial identifiers. We add a causal test of this link. In the LLM, activation patching traces how objects are bound to their roles~\cite{ax2609.35486}, and direction patching places the causal effect of spatial directions in middle and deep layers~\cite{ax2608.22916}. Spatial judgments have also been described as coarse object grounding followed by reasoning~\cite{ax2608.23074}. For contrastive encoders, a symmetric readout cancels direction that the frozen encoder still contains~\cite{ax2610.06324}. Methods that improve spatial answers, finally, use flips and subject--object swaps as training-free consistency signals~\cite{ax2609.17004,ax2608.03631}, re-weight attention at inference~\cite{ax2503.01773} or scale attention heads~\cite{ax2606.29579}. Unlike these methods and the sufficiency-type interventions above, ours acts on the input-side position code with the content fixed. In video LLMs, reversing the frame order lowers temporal understanding far more than editing the position encodings~\cite{ax2508.11576}, a precedent for separating position ids from token order. \Cref{tab:neighbors} summarizes the closest works.

\begin{table*}[!t]
\centering
\caption{Closest prior work. The sufficiency-type interventions among the listed works patch activations or visual-token content from a counterfactual image; none of the listed works compares the answer after an edit of the position code alone, item by item, with the answer to a real image transform.}
\label{tab:neighbors}
\small
\setlength{\tabcolsep}{3pt}
\begin{tabular}{>{\raggedright\arraybackslash}p{0.15\textwidth}>{\raggedright\arraybackslash}p{0.22\textwidth}>{\raggedright\arraybackslash}p{0.30\textwidth}>{\raggedright\arraybackslash}p{0.26\textwidth}}
\toprule
Work & Evidence & Intervention & Not addressed \\
\midrule
\cite{ax2509.02359} & necessity (ablation); accuracy per relation & mask / shuffle / constant per axis, ViT ($h$, $w$) and LLM ($t$, $x$, $y$), pixels fixed; What'sUp \emph{left}/\emph{right}/\emph{on}/\emph{under} & per-item comparison with a real flip; ids vs.\ token order; block localization; split by object order \\
\cite{ax2509.19191} & geometry; correlation; method & X/Y decomposition of 2D-RoPE attention; embedding-level interventions; frequency scaling & direction-predicting edit of the position code; LLM-side ids \\
\cite{ax2601.12809} & attention decomposition; ablation & position $\times$ token interaction in CLIP-style encoders (synthetic 1D scenes) & pretrained VLMs; LLM side \\
\cite{ax2603.22278} & content patching; interchange (sufficiency-type) & visual-token content; left and right regions swapped with a counterfactual image & position code; LLM-side ids and order \\
\cite{ax2601.12626} & activation patching (sufficiency-type); mean-difference spatial IDs and ID steering & activations from a mirrored-image run patched into the original run; spatial IDs & input-side position code (ViT or LLM ids) \\
\cite{ax2607.03358} & state patching with horizontal flips (sufficiency-type) & image, query-text and final-token states between an image and its flip (Qwen3-VL-4B) & position code; heads and MLPs \\
\cite{ax2606.31257} & behavior; LLM-side probes & blank image; steering at decision tokens & any position intervention \\
\cite{ax2609.35486,ax2608.22916,ax2608.23074} & activation and direction patching; steering; ablation; mediation & LLM residual stream, heads and tokens & vision encoder; position-code edit judged on answers \\
\cite{ax2503.17349} & token permutation; RoPE sensitivity probe & random permutation of visual tokens before the LLM and extra RoPE rotation of visual keys (LLaVA-1.5); visual-token norms & structured, direction-predicting edit; ids vs.\ token order; ViT side \\
This work & sufficiency (counterfactual); direction post hoc & mirror or reorder ViT or LLM position ids, pixels fixed; per axis; per block & where the relation is computed; models above 8B \\
\bottomrule
\end{tabular}
\end{table*}

\section{Position-only counterfactuals}
\label{sec:method}

\subsection{Where position enters a VLM}
\label{sec:method:where}

A vision-language model (VLM) receives spatial position at two places. Inside the vision encoder, every patch token carries a position code. Qwen2.5-VL~\cite{bai2025qwen25vl} applies a two-dimensional rotary position embedding (2D-RoPE~\cite{su2024roformer,ax2403.13298}) in every block of its vision transformer (ViT~\cite{dosovitskiy2021vit}). The rotation angle depends on the patch row $h$ and column $w$. InternVL3~\cite{zhu2025internvl3} instead adds a learned absolute position embedding once, before the first block. After the encoder, the visual tokens are serialized into the language model (LLM), where they receive LLM-side position ids. Qwen2.5-VL uses multimodal RoPE (M-RoPE~\cite{wang2024qwen2vl}) with separate temporal, height and width components. InternVL3 uses one-dimensional RoPE over the token sequence, trained with variable position increments for visual tokens~\cite{ax2412.09616}. At inference the released implementation assigns consecutive ids (increment 1) to all tokens, and we edit these ids (\ref{app:repro}). A third route is the patch content itself, which can carry orientation cues independently of any position code.

A rotary embedding rotates the query and key vectors of each patch by angles proportional to its row $h$ and column $w$, so that the attention between two patches depends on their offset. Replacing $w$ by $W-1-w$ therefore makes every patch behave as if it sat in the mirrored column, while its pixels stay where they are; for a learned absolute embedding the same edit re-indexes the table of position vectors. If the model then answers as it does for the flipped image, the column index alone is enough to produce the flipped-image answer, which is a test of sufficiency.

\begin{figure*}[t]
\centering
\includegraphics[width=\textwidth]{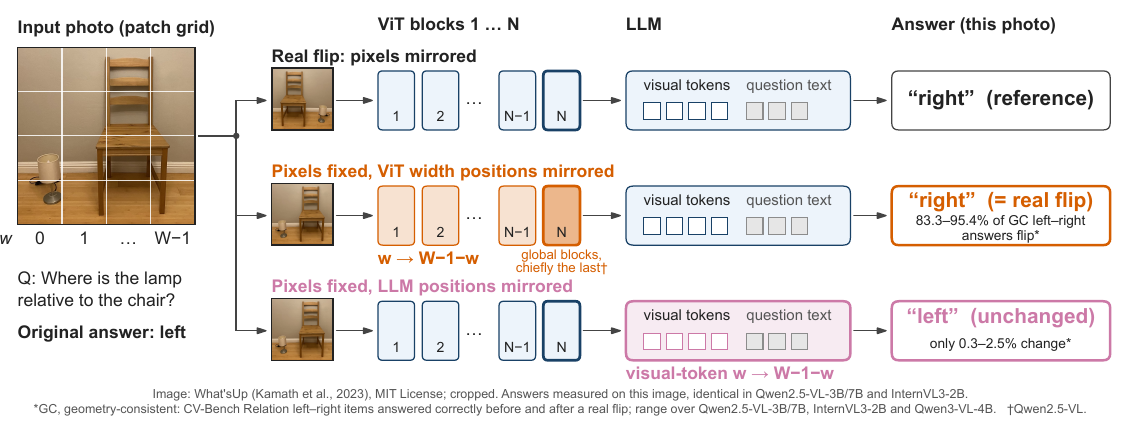}
\caption{Position-only counterfactuals keep every pixel and mirror only a position code. Rows: real horizontal flip (\hflip, reference); \vitw{}: width index $w\to W-1-w$ of each patch in the ViT; \llmw{}: LLM-side width ids of the visual tokens mirrored. Answers measured on this photograph; percentages: changed geometry-consistent left--right answers, four models (\cref{fig:flips}a), not agreement with \hflip{}; last-block note: \cref{fig:blocks}.}
\label{fig:schematic}
\end{figure*}

A real flip changes all three routes at once; our counterfactuals change one route at a time while leaving every pixel unchanged (\cref{fig:schematic}; rates on the geometry-consistent items in \cref{fig:flips}). For a position route, the edited variable is the assignment of positions to content: each patch receives the position id that it has in the flipped image, while its pixels stay in place. We call a route sufficient for the forced-choice answer to the degree that this replacement, with all other routes at their original values, reproduces the answer given for the flipped image. Sufficiency is thus a graded quantity, measured as agreement with the real transform or as a flip rate (\cref{sec:method:metrics}). If the answer stays unchanged, we detect no use of that route for the direction while the other routes are intact; this does not show that the route could not carry direction on its own.

\subsection{Counterfactual operations}
\label{sec:method:ops}

\paragraph{ViT-side mirroring (\vitw, \vith)}
For Qwen2.5-VL we replace the ViT's rotary position ids by their mirror image along one axis, $w' = W-1-w$ (or $h' = H-1-h$) on the $W\times H$ patch grid of the image. The rotary embedding is then recomputed from these ids. The mirrored ids are generated before the encoder's $2{\times}2$ spatial-merge reordering, so the reordering of tokens and the mirroring of id values do not interact. With no mirroring, our re-implementation reproduces the original rotary embedding element-wise; this is checked before every run. For InternVL3 we mirror the learned absolute position embedding on its $32\times32$ patch grid and leave the class token unchanged. To keep this a clean permutation, InternVL3 is run with a single $448\times448$ tile (256 visual tokens) instead of dynamic tiling; a check under the default tiling mirrors the embedding within each tile (\cref{sec:res:llm}). A pixel flip and a mirror of the patch grid then act on the same grid of patches. The image is resized to the tile without preserving its aspect ratio. Qwen2.5-VL keeps the aspect ratio and is limited to $1280\cdot 28\cdot 28$ pixels. The ViT of Qwen3-VL has both a learned absolute embedding and 2D-RoPE, and we mirror both with the same permutation; it keeps the aspect ratio and is limited to $1280\cdot 32\cdot 32$ pixels. %

\paragraph{LLM-side mirroring (\llmw, \llmh)}
For Qwen2.5-VL we mirror the width (or height) component of the M-RoPE ids of the visual tokens within the span of each image, $w' = w_{\min}+w_{\max}-w$. This is a permutation of ids within the image span, so the largest position id and the offsets of the following text tokens are unchanged. M-RoPE allocates its width component to the lowest-frequency channels, so this edit changes rotation angles only slightly (\cref{sec:res:llm}). For InternVL3, whose LLM has only one-dimensional positions, we apply the corresponding permutation to the position ids of the visual tokens (within each row for the width axis, across rows for the height axis). For Qwen3-VL we mirror the width component of its LLM-side ids as in Qwen2.5-VL.

\paragraph{Token reordering}
A causal LLM could also read direction from the serialization order of the visual tokens rather than from their ids, and we separate the two in Qwen2.5-VL, \ivl{} and Qwen3-VL. In \emph{order-only} conditions each visual token keeps its own position id but moves to a new seat in the sequence: the order is mirrored along the width or the height axis, or row-major order is replaced by column-major. Relative rotary positions are unchanged and only the causal order changes. In \emph{id-only} conditions the tokens stay in place and their ids are mirrored (this equals \llmw{} or \llmh); \emph{order-and-id} conditions apply both. As controls for the height axis we add two row permutations of the same strength that do not correspond to a vertical mirror: a cyclic shift of the rows by half the height and a random row shuffle (three seeds).

\paragraph{Pixel references and controls}
Each position-only counterfactual is compared with the matching real image transform: \vitw{} and \llmw{} with a real horizontal flip (\hflip), and \vith{} and \llmh{} with a real vertical flip (\vflip). A position-only $180^\circ$ rotation, which mirrors both axes of the ViT position code and of the LLM-side position ids while leaving the token order unchanged, is compared with a real $180^\circ$ rotation. The off-axis counterfactual, for example \vitw{} on up--down questions, is a control: a directional effect should flip the answers of the matching axis and leave the other axis largely intact. A position-only mirror does not reproduce a real flip in every respect. The pixels inside each patch keep their orientation, and the order in which the encoder merges neighboring patches is unchanged. So is the seat of each visual token in the LLM sequence. `Sufficient' therefore means that, among the differences between an image and its mirror image, the mirrored position code alone is enough to produce the flipped answer. The remaining differences are the competing explanations: the orientation of the pixels within each patch, the window partition of the encoder (which patches attend to each other in windowed blocks) and the order of the $2{\times}2$ merger. As a complementary content-only control, we flip the pixels and flip the ViT and LLM width ids back without reordering the tokens, so that the content, including the within-patch orientation, is mirrored together with the order in which the visual tokens enter the LLM; window partition and merge order are not tested separately.

\paragraph{Depth of the edit}
The ViT of Qwen2.5-VL has 32 blocks; four of them (blocks 7, 15, 23 and 31) use global attention and the rest windowed attention. We also apply the width mirror only in selected blocks: in a single block, in the four global-attention blocks, in all blocks except the last, or in the first or last $L$ blocks ($L=0,4,\dots,32$). Further conditions mirror the 28 windowed blocks alone and other sets of global-attention and late blocks (\cref{sec:res:blocks}). Because RoPE acts anew in every block, these conditions tell which blocks' rotations determine the output, not where the information is first written.

\subsection{Models and data}
\label{sec:method:data}

We study \qsmall-Instruct and \qlarge-Instruct (2D-RoPE in the ViT, M-RoPE in the LLM) and \ivl{} (learned absolute position embedding in the ViT, one-dimensional RoPE in the LLM). Each of the two common position-code designs of current open VLMs is thus represented by one vision encoder; InternVL3-2B and -8B use the same InternViT-300M encoder architecture. We call \qsmall, \qlarge{} and \ivl, on which the main runs and all control analyses were run, the \emph{core models}; the headline results of \cref{sec:intro} use \qlarge, Qwen3-VL-4B and \ivl, one model per series. As a scale check for the cross-axis leakage of the absolute-embedding mirror, we also ran InternVL3-8B on CV-Bench and What'sUp with the same code and settings. Qwen3-VL-4B-Instruct~\cite{ax2511.21631}, the third headline model, was run on both benchmarks with the main ViT- and LLM-side left--right conditions and the CV-Bench up--down mirrors, and, post hoc, with the same-axis permutation controls (\cref{tab:main,tab:whatsup}). Model revisions are fixed and listed in \ref{app:repro}. The main benchmark is the Relation subset of CV-Bench 2D~\cite{tong2024cambrian}: 650 questions, of which 383 ask whether one object is left or right of another and 267 ask whether it is above or below. We replicate the main findings on What'sUp~\cite{kamath2023whatsup}: Controlled Images A has 412 questions over 206 left--right and 206 up--down images, with the four options \emph{left}, \emph{right}, \emph{on} and \emph{under}; a comparison of its two axes is thus also one of projective relations with relations of support. Depth is studied on the Depth subset of CV-Bench 3D (a validation half and a held-out test half of 314 questions). %

\subsection{Scoring and metrics}
\label{sec:method:metrics}

\paragraph{Scoring}
Each question is posed once per condition with the benchmark's multiple-choice prompt (``Answer with the option's letter from the given choices directly''), and the answer is the option whose letter receives the highest next-token logit; no text is generated. The real and position-only $180^\circ$ conditions add the sentence ``Answer in terms of the image exactly as it is displayed (image coordinates), not the real-world orientation of the scene.'' The main CV-Bench and What'sUp runs use bf16, where two options occasionally receive exactly equal logits; ties resolve to the first option. Ties occur in \nb{tie-cv-7b} and \nb{tie-cv-3b} of the CV-Bench rows of \qlarge{} and \qsmall{} (all conditions), in \nb{ties-ivl} of those of \ivl{} and in \nb{tie-rows-wu-ivl}--\nb{tie-rows-wu-3b} of the What'sUp rows. Because the option order of CV-Bench is fixed (\emph{left} and \emph{above} are always option~A), \ref{app:ties} repeats the main analyses with every tied item removed. The option-order re-runs, InternVL3-8B, Qwen3-VL-4B, video input, the block, permutation, content-only, width-removal and tiling controls, and the depth validation half also use bf16 with first-option ties. The depth test half and its permutations, the fine-tuned checkpoints, EmbSpatial, the two-tile runs and the InternVL3-2B blank image score in fp32 (no ties occurred); fine-tuned accuracy and EmbSpatial combine both option orders.

\paragraph{Agreement with a real transform}
For each counterfactual we report the fraction of questions on which its answer equals the answer under the matching real transform (for example \vitw{} versus \hflip). This measures whether the position edit alone reproduces the effect of the image transform, regardless of whether either answer is correct. As a baseline we report the agreement of the original answer with the transformed-image answer.

\paragraph{Flip rate on the geometry-consistent subset}
To isolate items on which the model follows the image geometry, we restrict to questions that the model answers correctly both on the original image and under the matching real transform; we call them \emph{geometry-consistent}. We adopted this subset after the main runs, as the main directional metric, so that the flip rate is not diluted by items the model gets wrong anyway. On these items we report the fraction whose answer changes under the counterfactual. For off-axis rates on CV-Bench the subset is defined by the real transform of the edited axis (for example \vflip{} for \vith{} on left--right items); on What'sUp it is defined by the transform of the question's axis (\hflip{} for left--right items). On What'sUp we also report the fraction that changes to the mirror relation (for example \emph{left}$\to$\emph{right}, or \emph{on}$\to$\emph{under}; the correspondence is defined in the frame of the image), since a change to an off-axis option is not a directional effect; this share, too, was adopted as the main What'sUp metric after the runs.

\paragraph{Statistics}
Proportions are reported with 95\% Wilson intervals~\cite{wilson1927}. Differences between conditions on the same items use a paired bootstrap over questions~\cite{efron1993bootstrap} with 10{,}000 resamples and seed~0. The public checkpoints and the component swap (\ref{app:ft}) and the tile-order margin (\cref{sec:res:llm}) use 2{,}000 resamples. Analyses on subsets or with metrics chosen after the main runs are marked post hoc. What'sUp shows the same object pairs in several arrangements, so its questions are not independent; our intervals treat them as independent and are therefore somewhat too narrow. The two halves of CV-Bench 3D Depth are split by a hash of the item index, not by image, so they may share images. The code and the summary outputs of all analyses are released (\ref{app:repro}). %

\section{Results}
\label{sec:results}

\subsection{ViT width mirror versus a real horizontal flip}
\label{sec:res:lr}

With every pixel unchanged and only the ViT width ids mirrored (\vitw), the model gives the answer it gives for the horizontally flipped image (\hflip) on most CV-Bench left--right questions. \vitw{} agrees with \hflip{} on \nb{cv-agree-vitw-range} of them, against \nb{cv-agree-orig-range} for the original answers (\cref{tab:main}). The high baseline of \qsmall{} (\nb{cv-agree-orig-3b}) comes mostly from answering \emph{left} on both images. The agreement holds for both position-code families, the 2D-RoPE of Qwen2.5-VL and the learned absolute embedding of InternVL3. The converse test points the same way. When the pixels are flipped and the width ids of both ViT and LLM are flipped back, the content and the order of the token columns in the LLM are mirrored; the column order alone changes few answers (\cref{sec:res:llm}). The answer then moves away from the flipped-image answer (`flips back') on \nb{flipback-7b} of the left--right questions for \qlarge{}, \nb{flipback-ivl} for \ivl{} and, post hoc, \nb{qwen3-flipback} for Qwen3-VL-4B, so it follows the positions rather than the mirrored content. For \qsmall{} the rate is \nb{flipback-3b}. This model often gives the same answer to an image and to its mirror image, so its flip-back rate is low by construction.

\begin{table*}[!t]
\centering
\caption{Position-only counterfactuals on CV-Bench Relation (383 left--right, 267 up--down questions). Agreement: share of questions whose counterfactual answer equals the answer under the matching real transform. Flip rate: share of geometry-consistent items (correct on the original and on the real transform) whose answer changes. Brackets: 95\% Wilson intervals (integer percent). In \qsmall{} the original answers already agree with the flipped-image answers on \nb{cv-agree-orig-3b} of the left--right questions, mostly because both are \emph{left}, which inflates its agreement. InternVL3-8B: scale check; equal values in the 7B and 8B columns are coincidences (their answers differ on \nb{cmp7b8b-orig} of \nb{cmp7b8b-n} original and \nb{cmp7b8b-vitw} \vitw{} questions). $^\dagger$Swapped options: \nb{optorder-flip-vith-ud-sw-3b}; mean of both orders \nb{optorder-flip-vith-ud-avg-3b} (\ref{app:ties}). Qwen3-VL-4B: same bf16 scoring; --: the $180^\circ$ conditions and the off-axis \vith{} control were not run, and up--down accuracy and the all-item off-axis rate were not computed; its up--down subsets are defined by the vertical flip; the tie-removed analyses (\ref{app:ties}) do not include it; bf16 ties on \nb{qwen3-cv-tie-k} of its \nb{qwen3-cv-gc-n} geometry-consistent left--right items. $^\ddagger$Original answers without the frame sentence that the $180^\circ$ conditions add (\cref{sec:method:metrics}).}
\label{tab:main}
\footnotesize
\setlength{\tabcolsep}{3pt}
\begin{tabular}{p{0.24\textwidth}rrrrr}
\toprule
 & Qwen2.5-VL & Qwen2.5-VL & InternVL3 & Qwen3-VL & InternVL3 \\
 & 3B & 7B & 2B & 4B & 8B \\
\midrule
ViT position code & 2D-RoPE & 2D-RoPE & learned abs. & abs.\ + 2D-RoPE & learned abs. \\
LLM position ids & $(t,h,w)$ & $(t,h,w)$ & 1D & $(t,h,w)$ & 1D \\
\midrule
Accuracy, left--right (original) & \nb{acc-lr-3b} & \nb{acc-lr-7b} & \nb{acc-lr-ivl} & \nb{qwen3-cv-acc-lr} & \nb{acc-lr-ivl8b} \\
Accuracy, up--down (original) & \nb{acc-ud-3b} & \nb{acc-ud-7b} & \nb{acc-ud-ivl} & -- & \nb{acc-ud-ivl8b} \\
\multicolumn{6}{l}{\emph{Agreement with the real horizontal flip (all 383 left--right items)}}\\
\quad \vitw{} vs.\ \hflip{} & \nb{cv-agree-vitw-3b} {\scriptsize \nb{cv-agree-vitw-3b-ci}} & \nb{cv-agree-vitw-7b} {\scriptsize \nb{cv-agree-vitw-7b-ci}} & \nb{cv-agree-vitw-ivl} {\scriptsize \nb{cv-agree-vitw-ivl-ci}} & \nb{qwen3-cv-agree} {\scriptsize \nb{qwen3-cv-agree-ci}} & \nb{cv-agree-vitw-ivl8b} {\scriptsize \nb{cv-agree-vitw-ivl8b-ci}} \\
\quad \orig{} vs.\ \hflip{} (baseline) & \nb{cv-agree-orig-3b} {\scriptsize \nb{cv-agree-orig-3b-ci}} & \nb{cv-agree-orig-7b} {\scriptsize \nb{cv-agree-orig-7b-ci}} & \nb{cv-agree-orig-ivl} {\scriptsize \nb{cv-agree-orig-ivl-ci}} & \nb{qwen3-cv-agree-orig} {\scriptsize \nb{qwen3-cv-agree-orig-ci}} & \nb{cv-agree-orig-ivl8b} {\scriptsize \nb{cv-agree-orig-ivl8b-ci}} \\
\quad position-only $180^\circ$ vs.\ real $180^\circ$ & \nb{cv-agree-rot-3b} {\scriptsize \nb{cv-agree-rot-3b-ci}} & \nb{cv-agree-rot-7b} {\scriptsize \nb{cv-agree-rot-7b-ci}} & \nb{cv-agree-rot-ivl} {\scriptsize \nb{cv-agree-rot-ivl-ci}} & -- & \nb{cv-agree-rot-ivl8b} {\scriptsize \nb{cv-agree-rot-ivl8b-ci}} \\
\quad \orig{} vs.\ real $180^\circ$ (baseline)$^\ddagger$ & \nb{cv-agree-rotorig-3b} {\scriptsize \nb{cv-agree-rotorig-3b-ci}} & \nb{cv-agree-rotorig-7b} {\scriptsize \nb{cv-agree-rotorig-7b-ci}} & \nb{cv-agree-rotorig-ivl} {\scriptsize \nb{cv-agree-rotorig-ivl-ci}} & -- & \nb{cv-agree-rotorig-ivl8b} {\scriptsize \nb{cv-agree-rotorig-ivl8b-ci}} \\
\multicolumn{6}{l}{\emph{Flip rate on the geometry-consistent subset}}\\
\quad \vitw{}, left--right & \nb{cv-flip-vitw-lr-3b} {\scriptsize \nb{cv-flip-vitw-lr-3b-ci}} & \nb{cv-flip-vitw-lr-7b} {\scriptsize \nb{cv-flip-vitw-lr-7b-ci}} & \nb{cv-flip-vitw-lr-ivl} {\scriptsize \nb{cv-flip-vitw-lr-ivl-ci}} & \nb{qwen3-cv-vitw} {\scriptsize \nb{qwen3-cv-vitw-cii}} & \nb{cv-flip-vitw-lr-ivl8b} {\scriptsize \nb{cv-flip-vitw-lr-ivl8b-ci}} \\
\quad \vitw{}, up--down (off-axis control) & \nb{cv-flip-vitw-ud-3b} {\scriptsize \nb{cv-flip-vitw-ud-3b-ci}} & \nb{cv-flip-vitw-ud-7b} {\scriptsize \nb{cv-flip-vitw-ud-7b-ci}} & \nb{cv-flip-vitw-ud-ivl} {\scriptsize \nb{cv-flip-vitw-ud-ivl-ci}} & \nb{qwen3-cv-vitw-ud} {\scriptsize \nb{qwen3-cv-vitw-ud-ci}} & \nb{cv-flip-vitw-ud-ivl8b} {\scriptsize \nb{cv-flip-vitw-ud-ivl8b-ci}} \\
\quad \llmw{}, left--right & \nb{cv-flip-llmw-lr-3b} {\scriptsize \nb{cv-flip-llmw-lr-3b-ci}} & \nb{cv-flip-llmw-lr-7b} {\scriptsize \nb{cv-flip-llmw-lr-7b-ci}} & \nb{cv-flip-llmw-lr-ivl} {\scriptsize \nb{cv-flip-llmw-lr-ivl-ci}} & \nb{qwen3-cv-llmw} {\scriptsize \nb{qwen3-cv-llmw-cii}} & \nb{cv-flip-llmw-lr-ivl8b} {\scriptsize \nb{cv-flip-llmw-lr-ivl8b-ci}} \\
\quad \vith{}, up--down & \nb{cv-flip-vith-ud-3b}$^\dagger$ {\scriptsize \nb{cv-flip-vith-ud-3b-ci}} & \nb{cv-flip-vith-ud-7b} {\scriptsize \nb{cv-flip-vith-ud-7b-ci}} & \nb{cv-flip-vith-ud-ivl} {\scriptsize \nb{cv-flip-vith-ud-ivl-ci}} & \nb{qwen3-cv-vith-ud} {\scriptsize \nb{qwen3-cv-vith-ud-ci}} & \nb{cv-flip-vith-ud-ivl8b} {\scriptsize \nb{cv-flip-vith-ud-ivl8b-ci}} \\
\quad \vith{}, left--right (off-axis control) & \nb{cv-flip-vith-lr-3b} {\scriptsize \nb{cv-flip-vith-lr-3b-ci}} & \nb{cv-flip-vith-lr-7b} {\scriptsize \nb{cv-flip-vith-lr-7b-ci}} & \nb{cv-flip-vith-lr-ivl} {\scriptsize \nb{cv-flip-vith-lr-ivl-ci}} & -- & \nb{cv-flip-vith-lr-ivl8b} {\scriptsize \nb{cv-flip-vith-lr-ivl8b-ci}} \\
\quad \llmh{}, up--down & \nb{cv-flip-llmh-ud-3b} {\scriptsize \nb{cv-flip-llmh-ud-3b-ci}} & \nb{cv-flip-llmh-ud-7b} {\scriptsize \nb{cv-flip-llmh-ud-7b-ci}} & \nb{cv-flip-llmh-ud-ivl} {\scriptsize \nb{cv-flip-llmh-ud-ivl-ci}} & \nb{qwen3-cv-llmh-ud} {\scriptsize \nb{qwen3-cv-llmh-ud-ci}} & \nb{cv-flip-llmh-ud-ivl8b} {\scriptsize \nb{cv-flip-llmh-ud-ivl8b-ci}} \\
\multicolumn{6}{l}{\emph{Off-axis rate on all 267 up--down items}}\\
\quad \vitw{}, up--down & \nb{leak-all-3b} & \nb{leak-all-7b} & \nb{leak-all-2b} & -- & \nb{leak-all-8b} \\
\quad $n$: subsets of \vitw{} left--right / \vitw{} up--down / \vith{} up--down / \vith{} left--right (\llmw{}, \llmh{} as \vitw{}, \vith{} on the same axis) & \nb{cv-flip-vitw-lr-3b-n} / \nb{cv-flip-vitw-ud-3b-n} / \nb{cv-flip-vith-ud-3b-n} / \nb{cv-flip-vith-lr-3b-n} & \nb{cv-flip-vitw-lr-7b-n} / \nb{cv-flip-vitw-ud-7b-n} / \nb{cv-flip-vith-ud-7b-n} / \nb{cv-flip-vith-lr-7b-n} & \nb{cv-flip-vitw-lr-ivl-n} / \nb{cv-flip-vitw-ud-ivl-n} / \nb{cv-flip-vith-ud-ivl-n} / \nb{cv-flip-vith-lr-ivl-n} & \nb{qwen3-cv-gc-n} / \nb{qwen3-cv-ud-n} / \nb{qwen3-cv-ud-n} / -- & \nb{cv-flip-vitw-lr-ivl8b-n} / \nb{cv-flip-vitw-ud-ivl8b-n} / \nb{cv-flip-vith-ud-ivl8b-n} / \nb{cv-flip-vith-lr-ivl8b-n} \\
\bottomrule
\end{tabular}
\end{table*}

\begin{figure*}[t]
\centering
\includegraphics[width=\textwidth]{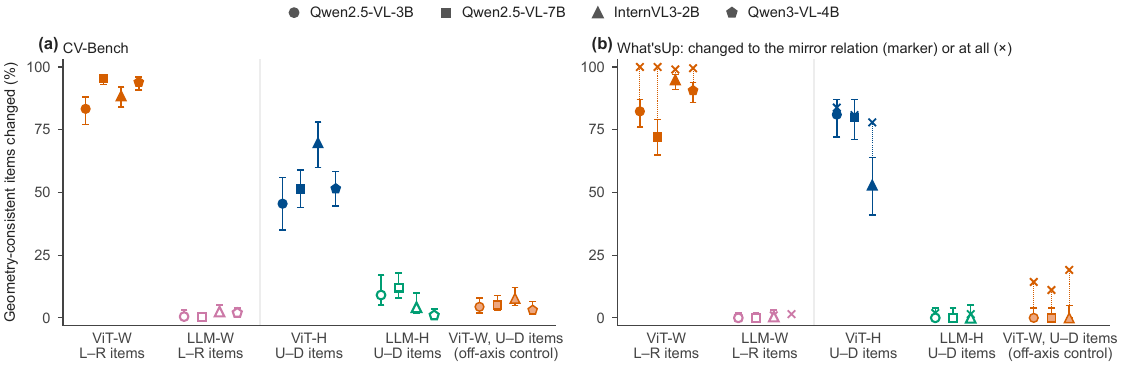}
\caption{Mirroring the ViT width positions flips most left--right answers; mirroring the LLM-side positions changes almost none of them. Share of geometry-consistent items whose answer changes, with 95\% Wilson intervals; color: condition, open: LLM-side mirror, shape: model. (a) CV-Bench Relation, \qsmall{} in the original option order. (b) What'sUp: change to the mirror relation (markers) and any change ($\times$); Qwen3-VL-4B: left--right conditions only, and for its \llmw{} only any change. Subset sizes in \cref{tab:main,tab:whatsup}.}
\label{fig:flips}
\end{figure*}

The effect is directional in the sense of axis selectivity. On the geometry-consistent items of the core models, \vitw{} changes \nb{cv-flip-vitw-lr-range} of the left--right answers but only \nb{cv-flip-vitw-ud-range} of the up--down answers (\cref{fig:flips}a). With two options, part of these flips would follow from any disturbance of the width code. Post hoc, in the core models and Qwen3-VL-4B, we located the two objects in \nb{objorder-cover-range} of the geometry-consistent items. On these items, permutations that keep the column order of the two objects flip \nb{objorder-keep-range} of the answers, permutations that reverse it \nb{objorder-rev-range} and the mirror \nb{objorder-mirror-range}. A half-grid shift flips \nb{perm-shift-cv-lr-range} of all geometry-consistent left--right answers (protocol, including how the objects were located, in \ref{app:protocols}). Among the random permutations alone, those that reverse the order flip \nb{objorder-shuf-diff-range} points more answers than those that keep it in all four models; the mirror exceeds the larger non-mirror permutation by margins whose intervals include zero in \qsmall{} and \ivl{} (\ref{app:protocols}). On CV-Bench, the answers thus tend to follow the objects' order in the code, which we did not separate from the half of the image in which the subject lands, but the flip rate alone does not measure the direction. Because a change under \vitw{} is necessarily a flip, we also checked whether the flips could reflect a fallback to a \emph{left} prior, which would flip only original \emph{right} answers. In \qlarge{} and \ivl{} the flip rates are similar for original \emph{left} and \emph{right} answers (\cref{tab:split}), and post hoc in Qwen3-VL-4B as well (\nb{qwen3-cv-vitw-left} of \nb{qwen3-cv-vitw-left-n} original \emph{left} and \nb{qwen3-cv-vitw-right} of \nb{qwen3-cv-vitw-right-n} \emph{right} answers flip). Flips from \emph{left} to \emph{right}, which a fallback to the \emph{left} prior cannot produce, occur on \nb{leftsplit-flip-left-7b} and \nb{leftsplit-flip-left-ivl} of the original \emph{left} answers in these two models (\nb{leftsplit-flip-left-3b} in \qsmall). With swapped options the gap in \qlarge{} widens, but its overall flip rate stays high at \nb{optorder-flip-vitw-lr-sw-7b} (\ref{app:ties}). \qsmall{} already answers \emph{left} on \nb{cv-left-pred-3b} of left--right questions, against \nb{cv-left-gold} in the gold labels. In a post hoc analysis of this model, all original \emph{right} answers flip but only \nb{leftsplit-flip-left-3b} of the \emph{left} ones, so part of its effect is a fallback to the prior (\cref{tab:split}). We therefore rest the main conclusion on \qlarge, Qwen3-VL-4B, \ivl{} and the four-option benchmark of \cref{sec:res:whatsup}. A position-only $180^\circ$ rotation, which mirrors both axes of the ViT and LLM-side position ids, agrees with the real rotation on \nb{cv-agree-rot-range} of left--right questions in the core models (\cref{tab:main}).

On all up--down items, the width mirror changes \nb{leak-all-3b}, \nb{leak-all-7b} and \nb{leak-all-2b} of the answers (\qsmall, \qlarge, \ivl; \cref{tab:main}), more than on the geometry-consistent ones.
\ivl{} also shows cross-axis leakage in the other direction. On What'sUp its height mirror swaps left and right on \nb{wu-offaxis-swap-ivl-pct} of the geometry-consistent left--right items (\nb{wu-offaxis-swap-ivl} of \nb{wu-offaxis-swap-ivl-n}), which the Qwen2.5-VL models almost never do.

Two explanations remain: the mirror may be a less local edit for learned absolute embeddings, or the small model may be unstable. To separate them we ran the same protocol on InternVL3-8B, which pairs the same vision encoder (InternViT-300M) with a larger LLM (last column of \cref{tab:main}). Its main effect matches the other models. The leakage of the width mirror into up--down answers shrinks from \nb{leak-all-2b} to \nb{leak-all-8b} on all items, the level of \qlarge{} (\nb{leak-all-7b}). On What'sUp, the height mirror also swaps left and right less often (\ref{app:protocols}). On the geometry-consistent CV-Bench left--right items, its height mirror changes \nb{cv-flip-vith-lr-ivl8b} of the answers, against \nb{cv-flip-vith-lr-3b} and \nb{cv-flip-vith-lr-7b} in the Qwen2.5-VL models, a difference we did not test (\cref{tab:main}). On the geometry-consistent items the reduction of the width-mirror leakage is not significant (\nb{leak-gc-diff} points \nb{leak-gc-diff-ci}). The pattern may point to model size rather than to the way the absolute embedding is mirrored, but a single larger model cannot show this (\cref{sec:limitations}).

\subsection{LLM-side position ids and token order}
\label{sec:res:llm}

Mirroring the width component of the LLM-side position ids of the visual tokens (\llmw) changes at most \nb{cv-flip-llmw-lr-ivl} of the geometry-consistent left--right answers in the core models. The largest Wilson upper bound is about \nb{cv-flip-llmw-lr-ivl-hi}. \llmw{} agrees with the original answers, not with the flipped-image answers (\cref{tab:main}). The two model families test this differently. In Qwen2.5-VL, M-RoPE assigns the width component to the lowest-frequency rotary channels (at most \nb{mrope-w-freq-hi} radians per position). For an image 36 merged tokens wide, mirroring the width ids thus changes no rotation angle by more than \nb{mrope-w-angle-36} rad. The Qwen width channel carries little within-image position by construction, and a null effect of \llmw{} in Qwen is weak evidence on its own. In \ivl, whose LLM uses one-dimensional RoPE over all frequencies, \llmw{} permutes the position ids of the visual tokens within each row and changes their relative positions substantially. Ids move by up to \nb{ivl-angle-ivl-llmw-maxshift} positions, which changes the rotation angles, wrapped to $[0,\pi]$, by \nb{ivl-angle-ivl-llmw-meanang-wrap} rad on average over channels and tokens, against at most \nb{mrope-w-angle-36} rad in Qwen2.5-VL. The edit thus changes rotation angles far more than in Qwen2.5-VL, and it too leaves almost all left--right answers unchanged. With the ViT code intact, no additional effect of the LLM-side ids is thus detected in \ivl. InternVL3 was trained with variable position increments (V2PE; \cref{sec:limitations}), but its LLM is not insensitive to these ids: once the ViT width code is removed, the id mirror alone flips \nb{nowidth-llmw-ivl} \nb{nowidth-llmw-ivl-ci} of the few geometry-consistent left--right answers that remain (\ref{app:protocols}). For Qwen2.5-VL, the token-order test below supports the same conclusion.

A causal LLM could also read direction from the order in which the visual tokens arrive rather than from their ids. In Qwen2.5-VL, moving the tokens to mirrored seats while each keeps its own id (order only, width) changes \nb{order-only-w-lr-range} of the left--right answers (\cref{tab:order}). This rate is no higher than the off-axis control that mirrors the row order (\nb{order-only-h-lr-range}) or a non-directional change from row-major to column-major order (\nb{colmajor-lr-range}). In \ivl, the same order-only test changes \nb{ivl-ordonly-w} \nb{ivl-ordonly-w-ci} of the geometry-consistent left--right answers. Within a single image tile, then, and with the ViT position code intact, we detected no use of the ids (with far larger angle changes in \ivl) or of the order of the visual tokens for the left--right direction.

In Qwen3-VL-4B~\cite{ax2511.21631}, \vitw{} flips \nb{qwen3-cv-vitw} \nb{qwen3-cv-vitw-ci} of the \nb{qwen3-cv-gc-n} geometry-consistent left--right answers on CV-Bench and \nb{qwen3-wu-vitw} \nb{qwen3-wu-vitw-ci} of the \nb{qwen3-wu-gc-n} on What'sUp. \llmw{} flips \nb{qwen3-cv-llmw} \nb{qwen3-cv-llmw-ci} and \nb{qwen3-wu-llmw} \nb{qwen3-wu-llmw-ci}, and the order-only mirror \nb{qwen3-cv-ordonly} \nb{qwen3-cv-ordonly-ci} and \nb{qwen3-wu-ordonly} \nb{qwen3-wu-ordonly-ci}. The ViT width code thus suffices to flip most answers, and no use of the LLM-side width ids or of the token order was detected on either benchmark. On CV-Bench, \vitw{} agrees with \hflip{} on \nb{qwen3-cv-agree} of all left--right questions, against \nb{qwen3-cv-agree-orig} for the original answers. The same-axis permutation controls, run post hoc, give the same picture as in the core models (\ref{app:protocols}). On CV-Bench, the half-grid shift flips \nb{perm-shift-cv-lr-qwen3} of the geometry-consistent left--right answers and the random permutations \nb{perm-ratio-cv-qwen3} \nb{perm-ratio-cv-qwen3-ci} as many as the mirror. Among random permutations alone, those that reverse the column order of the two objects flip \nb{objorder-shuf-diff-qwen3} points more of these answers than those that keep it. On What'sUp, a change to the mirror relation comes with the mirror (\nb{perm-mrel-wu-mirror-qwen3} of the changed answers) and the shift (\nb{perm-mrel-wu-shift-qwen3}), not with random permutations (\nb{perm-mrel-wu-shuf-qwen3-range}). In this model, too, the order reading rests on the random permutations on CV-Bench, whereas on What'sUp they rarely move its answers to the mirror relation; the changes also reflect disruption.

The main runs give \ivl{} a single tile. Its default dynamic tiling splits the image into up to 12 tiles and adds a thumbnail, and the LLM-side result holds there too. Mirroring the LLM-side width ids across the whole image, which also reverses the order of the tile columns, flips \nb{tiling-llmwimg} \nb{tiling-llmwimg-ci} of the \nb{tiling-gc-n} geometry-consistent left--right answers, and the corresponding order-only mirror flips \nb{tiling-orderonlywimg} \nb{tiling-orderonlywimg-ci}. Mirroring the absolute embedding within each tile and within the thumbnail flips \nb{tiling-vitwtile} \nb{tiling-vitwtile-ci}. In all three model series, and in \ivl{} also under its default tiling, no LLM-side use of the width ids or of the token order for left and right was detected with the ViT code intact. %

We also removed the ViT width position code, with constant ViT width ids in Qwen2.5-VL and absolute embeddings averaged over the width in \ivl. Left--right accuracy then falls to \nb{nowidth-lr-range} in the core models, while up--down accuracy changes by at most \nb{nowidth-ud-maxdiff} points (\ref{app:protocols}). In Qwen2.5-VL, constant width ids remove the relative width rotation and thus repeat the Mask-VE-w ablation of~\cite{ax2509.02359}, which already collapsed the left--right answers of Qwen2.5-VL-7B on What'sUp; this control adds the LLM-side mirrors on the items that remain. In \qsmall{} the fall comes about because, post hoc, \nb{nowidth-left-3b} of its answers collapse to \emph{left}, so its accuracy alone does not distinguish chance-level answering from such a collapse. On the few geometry-consistent items that remain (\nb{nowidth-gc-n-7b} in \qlarge{} and \nb{nowidth-gc-n-ivl} in \ivl), mirroring the LLM-side order and ids together still flips \nb{nowidth-oid-7b} \nb{nowidth-oid-7b-ci} and \nb{nowidth-oid-ivl} \nb{nowidth-oid-ivl-ci} of the answers, so a weak LLM-side contribution on these items is not excluded. This shows that the ViT route dominates in these trained models. It does not show that an LLM trained differently could not carry the direction (\cref{sec:limitations}). Across separately encoded tiles, however, the LLM uses their order: on the CV-Bench validation questions, splitting the image into two side-by-side tiles without a thumbnail lowers the left--right accuracy of \ivl{} from \nb{tiles-s-lr} to \nb{tiles-ht-lr}, and swapping the two tiles lowers it to \nb{tiles-swap-lr} (mean log-odds margin of the gold answer \nb{tiles-swap-dm} nats \nb{tiles-swap-dm-ci}). The result is consistent with the report that visual-token norms suppress the LLM's position encoding in LLaVA~\cite{ax2503.17349}, although we did not measure token norms. %

\begin{table*}[!t]
\centering
\caption{LLM-side token order in Qwen2.5-VL: fraction of answers that change when the visual tokens are reordered in the LLM input (\emph{order only}: each token keeps its own M-RoPE id; \emph{id only}: tokens stay in place and their ids are mirrored). Pooling the three shuffle seeds overstates the precision of that control; in neither model does the mirrored row order change clearly more up--down answers than both non-mirror permutations, and in 3B only about half of its flips agree with the real vertical flip.}
\label{tab:order}
\small
\setlength{\tabcolsep}{4pt}
\begin{tabular}{p{0.46\textwidth}rr}
\toprule
 & Qwen2.5-VL & Qwen2.5-VL \\
 & 3B & 7B \\
\midrule
\multicolumn{3}{l}{\emph{Left--right items (383), all items}}\\
\quad order only, width (ids follow tokens) & \nb{order-only-w-lr-3b} & \nb{order-only-w-lr-7b} \\
\quad order only, height (off-axis control) & \nb{order-only-h-lr-3b} & \nb{order-only-h-lr-7b} \\
\quad id only, width (the \llmw{} edit; all items) & \nb{id-only-w-lr-3b} & \nb{id-only-w-lr-7b} \\
\quad order and id, width & \nb{order-and-id-w-lr-3b} & \nb{order-and-id-w-lr-7b} \\
\quad row-major $\to$ column-major & \nb{colmajor-lr-3b} & \nb{colmajor-lr-7b} \\
\multicolumn{3}{l}{\emph{Up--down items (267), all items}}\\
\quad order only, height & \nb{order-only-h-ud-3b} & \nb{order-only-h-ud-7b} \\
\quad id only, height (the \llmh{} edit; all items) & \nb{id-only-h-ud-3b} & \nb{id-only-h-ud-7b} \\
\quad row-major $\to$ column-major (non-directional) & \nb{colmajor-ud-3b} & \nb{colmajor-ud-7b} \\
\multicolumn{3}{l}{\emph{Up--down items, geometry-consistent subset}}\\
\quad order only, height & \nb{rowperm-onlyh-3b} & \nb{rowperm-onlyh-7b} \\
\quad cyclic row shift (control) & \nb{rowperm-rowshift-3b} & \nb{rowperm-rowshift-7b} \\
\quad random row shuffle, pooled over 3 seeds (control) & \nb{rowperm-shuffle-3b} & \nb{rowperm-shuffle-7b} \\
\quad share of order-only flips also flipped by \vflip{} (all up--down items) & \nb{rowperm-vflip-share-3b} & \nb{rowperm-vflip-share-7b} \\
\bottomrule
\end{tabular}
\end{table*}

\subsection{Replication on What'sUp}
\label{sec:res:whatsup}

What'sUp~\cite{kamath2023whatsup} photographs the same objects in controlled arrangements. Its four options allow a stricter test: a directional effect should turn \emph{left} into \emph{right}, not into \emph{on} or \emph{under}. On the geometry-consistent left--right items, \vitw{} changes \nb{wu-flip-vitw-lr-range} of the answers and turns \nb{wu-mirror-vitw-lr-range} into the mirror relation, while \llmw{} changes \nb{wu-flip-llmw-lr-range} (Wilson upper bounds \nb{wu-flip-llmw-lr-hi-range}; \cref{tab:whatsup}, \cref{fig:flips}b). The mirror-relation rate, which we adopted as the main metric here after the run, is stricter than the change rate. The half-grid shift turns a similar share of the answers it changes into the mirror relation (\nb{perm-mrel-wu-shift-range}; \ref{app:protocols}), so the share shows that the change goes to the opposite side, not that it is specific to the mirror.
The four options also expose a non-directional component of the width mirror. In Qwen2.5-VL, \nb{wu-nonmirror-vitw-3b} (\qsmall) and \nb{wu-nonmirror-vitw-7b} (\qlarge) of these items move to \emph{on} or \emph{under} instead of the opposite side. The width mirror also changes \nb{wu-flip-vitw-ud-range} of the up--down answers, more than on CV-Bench. Conversely, the off-axis height mirror changes \nb{wu-flip-vith-lr-range} of the left--right answers, but in the Qwen2.5-VL models almost never to the opposite side. Counting all left--right items of \qlarge, \nb{wu-offaxis-onunder-7b} of the \nb{wu-offaxis-changed-7b} answers that change under \vith{} go to \emph{on} or \emph{under}. These off-option changes also explain why the agreement of \vitw{} with \hflip{} over all left--right items of \qlarge{} is lower on What'sUp (\nb{wu-agree-vitw-7b}) than on CV-Bench (\nb{cv-agree-vitw-7b}). Whatever the reason for these non-directional changes, the directional part of the width effect is large, and the LLM-side mirror changes few answers (\nb{wu-flip-llmw-lr-range}).

\subsection{Localization within the ViT}
\label{sec:res:blocks}

\begin{figure*}[t]
\centering
\includegraphics[width=\textwidth]{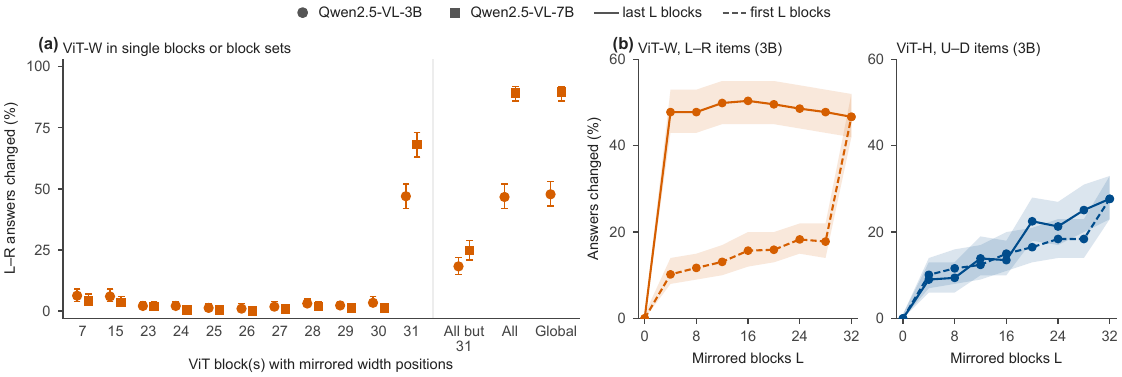}
\caption{The left--right effect of the width mirror is concentrated in the global-attention ViT blocks, chiefly the last, whereas in the 3B model the up--down effect of the height mirror builds up across blocks (Qwen2.5-VL, all CV-Bench Relation items, 95\% Wilson intervals). (a) \vitw{} in single blocks or block sets (shape: model); `Global': the four full-attention blocks 7, 15, 23 and 31. (b) Dose--response in the 3B model, mirroring the first or the last $L$ blocks.}
\label{fig:blocks}
\end{figure*}

We applied the width mirror in single blocks (7, 15 and 23 to 31), in the four full-attention blocks together, in all blocks except the last, and in all blocks. In Qwen2.5-VL, mirroring the width ids in the last block alone changes \nb{blk31-3b} of the left--right answers in \qsmall{} and \nb{blk31-7b} in \qlarge. This is the full effect of mirroring all 32 blocks in the 3B model (\nb{blk-ratio-last-3b}) and \nb{blk-ratio-last-7b} of it in the 7B model. Mirroring every block except the last leaves \nb{blk-ratio-butlast-range} of the full effect, and every other single block we tested changes few answers (\cref{fig:blocks}a): blocks 7, 15 and 23 change \nb{blk-glob-other-3b-range} (\qsmall) and \nb{blk-glob-other-7b-range} (\qlarge), and blocks 24--30 \nb{blk-late-3b-range} and \nb{blk-late-7b-range}; blocks 0--6, 8--14 and 16--22 were not mirrored singly. Post hoc, the direction of these changes differs. In \qsmall{} the last block alone changes answers in both directions (\nb{blkdir-blk31-rl-3b} from \emph{right} to \emph{left} and \nb{blkdir-blk31-lr-3b} the other way). Mirroring every block except the last instead changes them almost only from \emph{right} to \emph{left} (\nb{blkdir-butlast-rl-3b} against \nb{blkdir-butlast-lr-3b}), a fallback to the model's \emph{left} prior. In \qsmall{} the effect therefore sits in the last block, and the earlier blocks mainly disrupt. In \qlarge{} the changes from all but the last block go both ways (\nb{blkdir-butlast-rl-7b} and \nb{blkdir-butlast-lr-7b}). In \qsmall, a dose--response curve agrees: mirroring only the last four blocks changes \nb{dose-suf4-lr} of the answers, as many as mirroring all blocks (\nb{dose-all-lr}), whereas mirroring the first 28 blocks changes \nb{dose-pre28-lr} (\cref{fig:blocks}b). In \qlarge{} the four full-attention blocks together reach the full effect (\nb{blkglobal-7b}, against \nb{blkall-7b}), although the three earlier ones change few answers on their own.

Block sets separate the position of a block in the stack from its attention pattern. On the \nb{blkset-n-7b} geometry-consistent left--right items of \qlarge, mirroring the width ids in the four global-attention blocks flips \nb{blkset-g4-7b} of the answers, as many as mirroring all blocks (\nb{blkset-all-7b}; ratio \nb{blkset-gratio-7b} \nb{blkset-gratio-7b-ci}). The 28 windowed blocks together flip only \nb{blkset-nonglobal-7b}, and the windowed blocks 24--30 \nb{blkset-win2430-7b}. Mirroring the last eight blocks adds \nb{blkset-l8b31-7b} points \nb{blkset-l8b31-7b-ci} to block 31 alone (\nb{blkset-last8-7b} against \nb{blkset-b31-7b}), and the three earlier global-attention blocks together flip \nb{blkset-g3early-7b}. The effect in \qlarge{} thus sits in the global-attention blocks, chiefly the last, not in the last blocks as such. In \qsmall{} the two readings cannot be separated: block 31 alone already gives the full effect (\nb{blkset-b31-3b} against \nb{blkset-all-3b} on \nb{blkset-n-3b} items), and the windowed blocks together flip \nb{blkset-nonglobal-3b}. %

The two models differ in how concentrated the effect is. In \qsmall{} the last block alone carries the full effect, while block 23, another global-attention block, changes only \nb{blk23-3b} of the answers; in this model the effect thus sits in the last block rather than in global attention as such. In both models the last block, itself a global-attention block, contributes most, and in \qsmall{} it is sufficient alone. In \qlarge{} the rest of the effect comes from the other three global-attention blocks, not from the windowed blocks next to the last one. Because RoPE acts in every block, the result says which blocks' rotations matter for the output, not where the direction is first represented.

\subsection{Up--down judgments}
\label{sec:res:ud}

Earlier behavioral work found that, on camera-perspective benchmarks, VLMs ground horizontal relations in the image but answer vertical relations largely from priors, whereas their \emph{on}/\emph{under} answers in simple tabletop scenes are grounded in the image~\cite{ax2606.31257}. Our counterfactuals give a partial, position-level characterization of this asymmetry. On CV-Bench, the ViT height mirror changes only \nb{cv-flip-vith-ud-range-avg} of the geometry-consistent up--down answers in the core models. For \qsmall{} this is the mean of both option orders (\ref{app:ties}), whereas \cref{fig:flips}a shows its rate in the original order, \nb{cv-flip-vith-ud-3b}. The LLM-side height mirror still changes \nb{cv-flip-llmh-ud-range} in these models, so on this axis the LLM side is not silent. On What'sUp, with controlled scenes, the ViT height mirror changes \nb{wu-flip-vith-ud-range} (\cref{tab:whatsup}). How much the vertical judgment depends on the height position thus varies between data sets. With two options, a CV-Bench change rate near one half is also what answering at random would give once the vertical information is disrupted. Non-mirror row permutations of the height code (a half-grid shift and random permutations; \ref{app:protocols}) separate the two readings in part. Post hoc, the height mirror flips \nb{hperm-mmax-cv-7b} \nb{hperm-mmax-cv-7b-ci} and \nb{hperm-mmax-cv-ivl} \nb{hperm-mmax-cv-ivl-ci} points more up--down answers than the larger of these permutations in \qlarge{} and \ivl, but only \nb{hperm-mmax-cv-3b} \nb{hperm-mmax-cv-3b-ci} in \qsmall; on What'sUp the margin is \nb{hperm-mmax-wu-range} points. The dose--response curve shows a second difference: in \qsmall{} the effect of the height mirror accumulates over the blocks instead of arising at the end (\nb{dose-suf4-ud} for the last four blocks, \nb{dose-all-ud} for all; \cref{fig:blocks}b).

A content-only control locates a second source of the up--down answers. As for the horizontal axis (\cref{sec:res:lr}), we flip the pixels, now vertically, and flip the ViT and LLM height ids back, so that the content and the order of the token rows in the LLM are mirrored. On What'sUp, in the core models, the answer then moves away from the flipped-image answer on \nb{conly-wu-fbv-range} of the up--down questions, against \nb{conly-wu-fbh-range} of the left--right questions under the horizontal version. The difference is \nb{conly-wu-d-range} points, and no interval reaches 0 in any of the three core models. This flip-back rate also counts changes to an off-axis option, which are not directional (\cref{sec:method:metrics}); a stricter count follows at the end of this paragraph. Mirroring the row order alone changes \nb{order-only-h-ud-range} of the up--down answers in the Qwen2.5-VL models (\cref{tab:order}), so the asymmetry can be attributed to the orientation of the content only in part. Part of this difference comes from questions whose answer a real flip leaves unchanged: post hoc, the real vertical flip leaves \nb{udcontent-same-ud-range} of the up--down answers unchanged, against \nb{udcontent-same-lr-range} of the left--right answers under the horizontal flip. Neither vertical flip adds the instruction to answer in image coordinates that the $180^\circ$ conditions add (\cref{sec:method:metrics}), so some of these unchanged answers may come from a gravity-based frame or from the support that \emph{on} and \emph{under} describe. On the questions whose answer the real flip changes, the difference is \nb{udcontent-dis-d-range} points (\qlarge: \nb{udcontent-dis-d-7b} \nb{udcontent-dis-d-7b-ci}; \qsmall: \nb{udcontent-dis-d-3b} \nb{udcontent-dis-d-3b-ci}; \ivl: \nb{udcontent-dis-d-ivl} \nb{udcontent-dis-d-ivl-ci}). Post hoc, CV-Bench shows the same over all items for \qlarge{} (\nb{conly-cv-fbv-7b} against \nb{conly-cv-fbh-7b}) and \ivl{} (\nb{conly-cv-fbv-ivl} against \nb{conly-cv-fbh-ivl}), but not for \qsmall{} (\nb{conly-cv-fbv-3b} against \nb{conly-cv-fbh-3b}). All items include questions whose answer a real flip leaves unchanged; on the geometry-consistent items, the content-only flip leaves \nb{conly-cv-gcv-7b} of the up--down and \nb{conly-cv-gch-7b} of the left--right answers at the original answer in \qlarge{} and \nb{conly-cv-gcv-ivl} and \nb{conly-cv-gch-ivl} in \ivl, but \nb{conly-cv-gcv-3b} and \nb{conly-cv-gch-3b} in \qsmall. Local cues in the content may contribute: the \emph{on} and \emph{under} options imply support, which is visible in the content, whereas a horizontal mirror leaves such cues intact, although it can still be detected from text or hands~\cite{lin2020chirality}. The up--down statistics of natural images are also asymmetric, so vertically flipped content is itself an unusual input. The up--down questions of What'sUp ask about support (\emph{on}/\emph{under}), whereas its left--right questions relate two separate objects. Relation type and axis are confounded, and the contrast cannot be attributed to the vertical axis as such. Up--down answers thus depend on the height position code together with the orientation of the content; a contribution of the order of the token rows cannot be excluded. The stricter count, post hoc, uses the geometry-consistent What'sUp items, which the model answers correctly in both orientations, and asks whether the content-only flip returns the answer to the original one. It does so for \nb{conly-wu-gcv-7b} of the up--down and \nb{udcontent-gc-lr-7b} of the left--right answers in \qlarge{} (difference \nb{udcontent-gc-d-7b} points \nb{udcontent-gc-d-7b-ci}) and for \nb{conly-wu-gcv-3b} and \nb{udcontent-gc-lr-3b} in \qsmall{} (\nb{udcontent-gc-d-3b} points \nb{udcontent-gc-d-3b-ci}). In \ivl{} the shares are \nb{conly-wu-gcv-ivl} and \nb{udcontent-gc-lr-ivl} (\nb{udcontent-gc-d-ivl} points \nb{udcontent-gc-d-ivl-ci}). By this count the contrast is clear only in \ivl. On the same items, the content-only flip moves \nb{udcontent-gc-udreal-range} of the up--down answers to the flipped-image answer, against \nb{udcontent-gc-lrreal-range} of the left--right answers; the other changed left--right answers go to an off-axis option (in \qlarge{} all of them, \nb{udcontent-gc-lroff-7b} of the items).

\begin{table*}[!t]
\centering
\caption{What'sUp (Controlled Images A; 412 questions with four options: \emph{left}, \emph{right}, \emph{on}, \emph{under}). The main metric is the fraction that changes to the mirror relation (left$\leftrightarrow$right or on$\leftrightarrow$under); a change to an off-axis option is not a directional effect. On up--down items, random row permutations also turn \nb{hperm-mrel-wu-shuf-q-range} (Qwen2.5-VL) and \nb{hperm-mrel-wu-shuf-ivl-range} (InternVL3-2B) of the answers they change into on$\leftrightarrow$under, so the up--down mirror-relation row does not show direction. Qwen3-VL-4B: left--right conditions only; --: not run.}
\label{tab:whatsup}
\small
\setlength{\tabcolsep}{4pt}
\begin{tabular}{p{0.36\textwidth}rrrr}
\toprule
 & Qwen2.5-VL & Qwen2.5-VL & InternVL3 & Qwen3-VL \\
 & 3B & 7B & 2B & 4B \\
\midrule
Accuracy, left--right / up--down (original) & \nb{wu-acc-lr-3b} / \nb{wu-acc-ud-3b} & \nb{wu-acc-lr-7b} / \nb{wu-acc-ud-7b} & \nb{wu-acc-lr-ivl} / \nb{wu-acc-ud-ivl} & \nb{qwen3-wu-acc-lr} / -- \\
\multicolumn{5}{l}{\emph{Left--right items, geometry-consistent subset}}\\
\quad \vitw{}: answer changes & \nb{wu-flip-vitw-lr-3b} & \nb{wu-flip-vitw-lr-7b} & \nb{wu-flip-vitw-lr-ivl} & \nb{qwen3-wu-vitw} \\
\quad \vitw{}: changes to the mirror relation & \nb{wu-mirror-vitw-lr-3b} {\scriptsize \nb{wu-mirror-vitw-lr-3b-ci}} & \nb{wu-mirror-vitw-lr-7b} {\scriptsize \nb{wu-mirror-vitw-lr-7b-ci}} & \nb{wu-mirror-vitw-lr-ivl} {\scriptsize \nb{wu-mirror-vitw-lr-ivl-ci}} & \nb{qwen3-wu-mirror} {\scriptsize \nb{qwen3-wu-mirror-ci}} \\
\quad \llmw{}: answer changes & \nb{wu-flip-llmw-lr-3b} & \nb{wu-flip-llmw-lr-7b} & \nb{wu-flip-llmw-lr-ivl} & \nb{qwen3-wu-llmw} \\
\quad \vith{} (off-axis): answer changes & \nb{wu-flip-vith-lr-3b} & \nb{wu-flip-vith-lr-7b} & \nb{wu-flip-vith-lr-ivl} & -- \\
\multicolumn{5}{l}{\emph{Up--down items, geometry-consistent subset}}\\
\quad \vith{}: answer changes & \nb{wu-flip-vith-ud-3b} & \nb{wu-flip-vith-ud-7b} & \nb{wu-flip-vith-ud-ivl} & -- \\
\quad \vith{}: changes to the mirror relation & \nb{wu-mirror-vith-ud-3b} & \nb{wu-mirror-vith-ud-7b} & \nb{wu-mirror-vith-ud-ivl} & -- \\
\quad \vitw{} (off-axis): answer changes & \nb{wu-flip-vitw-ud-3b} & \nb{wu-flip-vitw-ud-7b} & \nb{wu-flip-vitw-ud-ivl} & -- \\
$n$ (left--right / up--down subset) & \nb{wu-flip-vitw-lr-3b-n} / \nb{wu-flip-vith-ud-3b-n} & \nb{wu-flip-vitw-lr-7b-n} / \nb{wu-flip-vith-ud-7b-n} & \nb{wu-flip-vitw-lr-ivl-n} / \nb{wu-flip-vith-ud-ivl-n} & \nb{qwen3-wu-gc-n} / -- \\
\vitw{} vs.\ \hflip{} agreement (all left--right) & \nb{wu-agree-vitw-3b} & \nb{wu-agree-vitw-7b} & \nb{wu-agree-vitw-ivl} & \nb{qwen3-wu-agree} \\
\bottomrule
\end{tabular}
\end{table*}

\begin{figure*}[t]
\centering
\includegraphics[width=\textwidth]{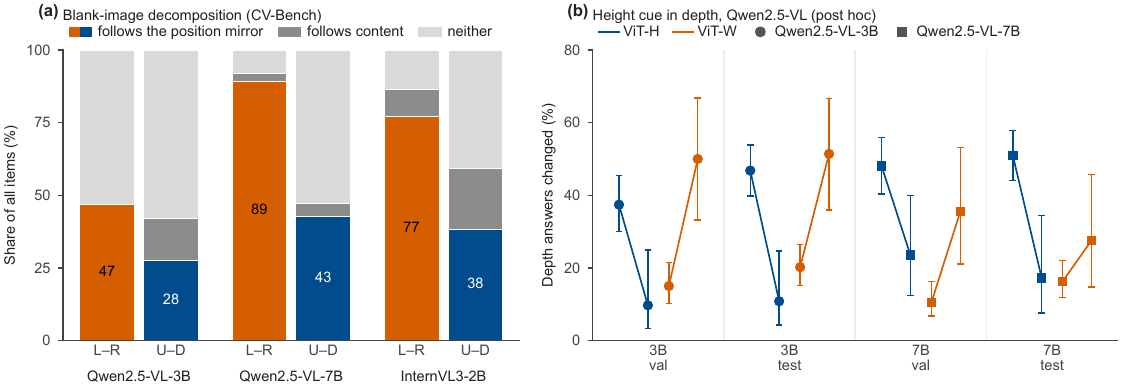}
\caption{Up--down answers follow the position code less than left--right answers, and depth answers that follow the height cue depend on the ViT height positions. (a) Blank-image decomposition of all CV-Bench Relation items into answers that follow position (flip under the matching ViT mirror, disruption included), follow content, or neither (which includes image-dependent answers). (b) CV-Bench 3D depth, Qwen2.5-VL, validation and test halves (post hoc): answers changed by \vith{} and \vitw{} when the original answer follows the `higher means farther' heuristic (left of each pair) or not (right); 95\% Wilson intervals.}
\label{fig:decomp}
\end{figure*}

To decompose the answers, we combine the position mirror with a blank gray image, the control used by~\cite{ax2606.31257}. An answer \emph{follows position} if it flips under the matching ViT mirror (with two options, this includes flips caused by disruption), \emph{follows content} if it does not flip but differs from the blank-image answer, and \emph{follows neither} otherwise. Following position is the largest class for left--right answers in \qlarge{} and \ivl{} (\nb{blank-lr-pos-range} of the questions). It is not in \qsmall, whose left--right answers follow position in \nb{blank-lr-pos-3b} and neither in \nb{blank-lr-prior-3b}, in line with its \emph{left} prior. The third class is larger for up--down than for left--right answers in all three core models: up--down answers follow position in only \nb{blank-ud-pos-range} in \qlarge{} and \ivl{} (\cref{fig:decomp}a).

The third class is not a pure label prior. Post hoc, \nb{blankclass-ud-real-3b}, \nb{blankclass-ud-real-7b} and \nb{blankclass-ud-real-ivl} of the up--down answers in this class (\qsmall, \qlarge, \ivl) change under a real vertical flip, so this class contains image-dependent answers that the height position code does not explain.

On the LLM side, up--down answers are more sensitive to reordering the rows of image tokens than left--right answers are to reordering the columns: \nb{order-only-h-ud-range} of up--down answers change when the row order is mirrored. Whether the row order itself carries up--down information is not established by this rate (\cref{tab:order}).

\subsection{The `higher means farther' depth cue}
\label{sec:res:depth}

VLMs tend to judge an object that is higher in the image to be farther away~\cite{ax2605.30161}, and we asked whether they read this cue from the ViT height positions. The CV-Bench 3D depth questions were split by item index into a validation half and a held-out test half before the experiment. A question counts as heuristic-consistent if the object higher in the image is the farther one (\ref{app:protocols}). On the test half, mirroring only the ViT height ids changes \nb{depth-test-vith-7b} \nb{depth-test-vith-7b-ci} of the answers of \qlarge. This is more than mirroring the width ids (\nb{depth-test-vitw-7b}) or the LLM-side height ids (\nb{depth-test-llmh-7b}; \cref{tab:depth}). In a post hoc paired bootstrap, the difference between the height and the width mirror is \nb{depth-hw-diff-7b} points \nb{depth-hw-diff-7b-ci}. The result replicates the validation half, which is not fully independent because \nb{depth-halves-shared-test} test questions share source images with it (\ref{app:protocols}). Its directional part, the share of changed answers that now name the higher object as closer, does not discriminate: the width mirror also reaches \nb{depth-test-dir-vitw-7b}, against \nb{depth-test-dir-vith-7b} for the height mirror. The evidence therefore rests on the comparison of the two change rates, which can also differ because the two mirrors disrupt the image to different degrees.

Further controls show that the direction of the change is not specific to the mirror either. Row permutations of the ViT height code change fewer answers than the mirror in \qlarge{} (half-grid cyclic shift \nb{depth-perm-shift-7b}; random row permutations \nb{depth-perm-shuf-7b}, mean of three seeds). Pooled over these four permutations, \nb{depth-perm-dir-perm-7b} of the answers they change now name the higher object as closer, against \nb{depth-perm-dir-vith-7b} for the mirror. The share lies close to the \nb{depth-heur-base-7bt} of original answers that follow the heuristic, as expected if the changes fall mainly on these answers. The share is not fixed by that baseline, since the width permutations reach only \nb{depth-perm-w-dir-perm-7b}. The pooled permutations also agree with a real vertical flip at least as often as the mirror (\nb{depth-vflipagree-perm-7b} against \nb{depth-vflipagree-vith-7b}). Neither comparison separates the mirror from the permutations. What the results support is that the use of the cue depends on the intact height code.

In \qsmall{} the effect is weaker on both halves (validation half: height mirror \nb{depth-val-vith-3b} \nb{depth-val-vith-3b-ci}, width mirror \nb{depth-val-vitw-3b} \nb{depth-val-vitw-3b-ci}). Its height mirror changes \nb{depth-test-vith-3b} of the answers and the width mirror \nb{depth-test-vitw-3b}, and the lower Wilson bound of the first falls just below the upper bound of the second (\cref{tab:depth}). The post hoc paired difference, \nb{depth-hw-diff-3b} points \nb{depth-hw-diff-3b-ci}, points the same way as in \qlarge. We therefore read the 3B result as a weaker effect in the same direction, not as the absence of the cue's use. On depth questions the width mirror is also a less selective control than on CV-Bench 2D. Since pixels are unchanged, a correct depth answer should not change under a position mirror. The result identifies the route of the cue, not a source of depth information. %

A post hoc analysis on CV-Bench 3D shows which answers depend on the height code, in both models. If the model reads height position as a depth cue, the height mirror should change mainly the answers that originally follow the heuristic. It should rarely change those that go against it, because for them the mirrored heuristic points to the original answer. This is what we observe. In the \qlarge{} test half the height mirror changes \nb{depth-spec-vith-con-7bt} of the heuristic-following answers and \nb{depth-spec-vith-inc-7bt} of the answers that go against the heuristic, whereas the width mirror shows the reverse (\nb{depth-spec-vitw-con-7bt} versus \nb{depth-spec-vitw-inc-7bt}). The gap for the height mirror is \nb{depth-spec-gap-range} percentage points across both models and both halves (\cref{tab:depth}, \cref{fig:decomp}b). The difference in differences, the height-mirror gap minus the width-mirror gap, is \nb{depth-did-did-7bt} points \nb{depth-did-did-7bt-ci} in the \qlarge{} test half and \nb{depth-did-did-3bt} \nb{depth-did-did-3bt-ci} in \qsmall{} (paired bootstrap; \nb{depth-spec-did-range} points across both models and halves). It adds up two gaps of opposite sign: the height mirror changes the heuristic-following answers more often than the others, whereas the width mirror changes the answers against the heuristic more often, which moves them back to the heuristic. The groups of answers against the heuristic are small (\nb{depth-spec-vith-inc-n-7bt} and \nb{depth-spec-vith-inc-n-3bt} items in the test halves). Removing the cue by permuting the rows of the height code predicts the same pattern, and post hoc the row permutations show it, more weakly. Pooled over the four permutations, they change \nb{depth-permsplit-h-follow-7b} of the heuristic-following and \nb{depth-permsplit-h-against-7b} of the other answers in \qlarge{} (\qsmall: \nb{depth-permsplit-h-follow-3b} and \nb{depth-permsplit-h-against-3b}), whereas the width permutations change \nb{depth-permsplit-w-follow-7b} and \nb{depth-permsplit-w-against-7b}. This analysis, too, shows that the use of the cue depends on the height code, not that the mirror specifically reverses it. At a similar overall change rate, a height permutation (\nb{depth-matched-match-h-7b} of answers changed) remains far more selective for heuristic-following answers than a width permutation (\nb{depth-matched-match-w-7b}): the difference in gaps is \nb{depth-matched-match-7b} points \nb{depth-matched-match-7b-ci} in \qlarge{} and \nb{depth-matched-match-3b} \nb{depth-matched-match-3b-ci} in \qsmall{} (post hoc), so a difference in the strength of disruption does not explain the selectivity. On the conflict items, however, the height mirror corrects \nb{depth-conflict-vith-7b} of the \nb{depth-conflict-n-7b} original errors of \qlarge{} and the width mirror \nb{depth-conflict-vitw-7b} (difference \nb{depth-conflict-diff-7b} points \nb{depth-conflict-diff-7b-ci}; \qsmall: \nb{depth-conflict-vith-3b} and \nb{depth-conflict-vitw-3b} of \nb{depth-conflict-n-3b}; post hoc), so the two mirrors cannot be told apart there. On EmbSpatial the height mirror corrects hardly more conflict errors than the width mirror, and the same split does not replicate (\ref{app:emb}). Height in the visual field is a valid monocular depth cue, and the models do not follow it more often than the gold labels do: on the consistent and conflict items of all CV-Bench 3D depth questions, \qlarge{} follows it on \nb{depth-follow-follow-all-7b} and \qsmall{} on \nb{depth-follow-follow-all-3b}, against \nb{depth-follow-base} in the gold labels (differences \nb{depth-follow-diff-all-7b} \nb{depth-follow-diff-all-7b-ci} and \nb{depth-follow-diff-all-3b} \nb{depth-follow-diff-all-3b-ci} points; post hoc). The data therefore show use of the cue, not over-use: the models follow it about as often as it holds and tell conflict items apart only in part, which lowers accuracy on them.

\begin{table*}[!t]
\centering
\caption{Depth questions of CV-Bench 3D. Top: fraction of answers that change under each counterfactual on the held-out test half (95\% Wilson intervals). Bottom (post hoc): change rates split by whether the original answer follows the `higher in the image means farther' heuristic (following) or not (against); group sizes in parentheses (they differ slightly between conditions because tied items are excluded per condition). The remaining test-half items are ambiguous for the heuristic.}
\label{tab:depth}
\footnotesize
\setlength{\tabcolsep}{4pt}
\begin{tabular}{p{0.30\textwidth}rr}
\toprule
 & Qwen2.5-VL & Qwen2.5-VL \\
 & 3B & 7B \\
\midrule
\multicolumn{3}{l}{\emph{Test half}}\\
\quad \vith{} & \nb{depth-test-vith-3b} \nb{depth-test-vith-3b-ci} & \nb{depth-test-vith-7b} \nb{depth-test-vith-7b-ci} \\
\quad \vitw{} & \nb{depth-test-vitw-3b} \nb{depth-test-vitw-3b-ci} & \nb{depth-test-vitw-7b} \nb{depth-test-vitw-7b-ci} \\
\quad \llmh{} & \nb{depth-test-llmh-3b} & \nb{depth-test-llmh-7b} \\
\quad $n$ & \nb{depth-test-n-3b} & \nb{depth-test-n-7b} \\
\midrule
\multicolumn{3}{l}{\emph{Post hoc: change rate, answers following / against the heuristic}}\\
\quad \vith{}, validation & \nb{depth-spec-vith-con-3bv} (\nb{depth-spec-vith-con-n-3bv}) / \nb{depth-spec-vith-inc-3bv} (\nb{depth-spec-vith-inc-n-3bv}) & \nb{depth-spec-vith-con-7bv} (\nb{depth-spec-vith-con-n-7bv}) / \nb{depth-spec-vith-inc-7bv} (\nb{depth-spec-vith-inc-n-7bv}) \\
\quad \vitw{}, validation & \nb{depth-spec-vitw-con-3bv} (\nb{depth-spec-vitw-con-n-3bv}) / \nb{depth-spec-vitw-inc-3bv} (\nb{depth-spec-vitw-inc-n-3bv}) & \nb{depth-spec-vitw-con-7bv} (\nb{depth-spec-vitw-con-n-7bv}) / \nb{depth-spec-vitw-inc-7bv} (\nb{depth-spec-vitw-inc-n-7bv}) \\
\quad \vith{}, test & \nb{depth-spec-vith-con-3bt} (\nb{depth-spec-vith-con-n-3bt}) / \nb{depth-spec-vith-inc-3bt} (\nb{depth-spec-vith-inc-n-3bt}) & \nb{depth-spec-vith-con-7bt} (\nb{depth-spec-vith-con-n-7bt}) / \nb{depth-spec-vith-inc-7bt} (\nb{depth-spec-vith-inc-n-7bt}) \\
\quad \vitw{}, test & \nb{depth-spec-vitw-con-3bt} (\nb{depth-spec-vitw-con-n-3bt}) / \nb{depth-spec-vitw-inc-3bt} (\nb{depth-spec-vitw-inc-n-3bt}) & \nb{depth-spec-vitw-con-7bt} (\nb{depth-spec-vitw-con-n-7bt}) / \nb{depth-spec-vitw-inc-7bt} (\nb{depth-spec-vitw-inc-n-7bt}) \\
\bottomrule
\end{tabular}
\end{table*}

\subsection{Robustness and side effects}
\label{sec:res:robust}

\paragraph{Robustness}
The left--right result does not depend on the input mode or on the tie rule. When Qwen2.5-VL-7B receives each image of the CV-Bench validation split (\nb{cv-val-n} of the \nb{cv-split-n} questions) as a video of four identical frames, \vitw{} flips \nb{video-vid-vitw-7b} of the \nb{video-n-vid-7b} geometry-consistent left--right answers of \qlarge{} (image mode on the same items: \nb{video-img-vitw-7b}). In the same video mode, \llmw{} flips \nb{video-vid-llmw-7b} of these answers (Wilson upper bound \nb{video-vid-llmw-7b-hi}). Removing every item that involves a bf16 tie changes the main metrics little and reverses no conclusion (\cref{tab:ties}).

\begin{figure*}[t]
\centering
\includegraphics[width=\textwidth]{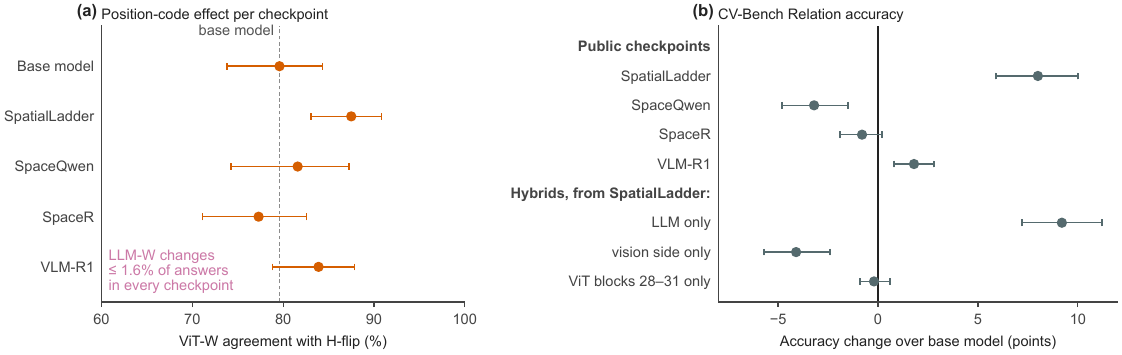}
\caption{In four public fine-tuned \qsmall{} checkpoints the position-code effect is preserved, and the gain of SpatialLadder follows its LLM. (a) Agreement of \vitw{} with \hflip{} on left--right items that the real flip changes, with 95\% Wilson intervals; dashed line: base model. (b) CV-Bench Relation accuracy change over the base model for the public checkpoints and for hybrids that take only the named components from SpatialLadder; 95\% paired-bootstrap intervals.}
\label{fig:ft}
\end{figure*}

The result also holds in four publicly released spatially fine-tuned versions of \qsmall{} (SpatialLadder, SpaceQwen, SpaceR and VLM-R1; \cref{fig:ft}a). On left--right items whose answer changes under the real flip, \vitw{} agrees with \hflip{} on \nb{pubft-agree-ft-range} of the items, against \nb{pubft-agree-base} for the base model, and \llmw{} changes at most \nb{pubft-llmw-max} of the answers. The mechanism is thus preserved in all four (\ref{app:ft}). Swapping components between SpatialLadder, the checkpoint with the largest accuracy gain, and its base model attributes its gain on CV-Bench Relation to the LLM (\cref{fig:ft}b). The base vision encoder with the fine-tuned LLM recovers the full gain (share \nb{swap-share-llms} \nb{swap-share-llms-ci}). The fine-tuned vision encoder with the base LLM is less accurate than the base model (\nb{swap-diff-vits} points \nb{swap-diff-vits-ci}). The gain thus follows the LLM. This result is consistent with fine-tuning changing how the LLM uses the output of an unchanged encoder. With one checkpoint, it does not establish this for spatial fine-tuning in general. The swap assumes that the two components can be separated: the drop with the fine-tuned vision encoder and the base LLM may also reflect a mismatch between components that were adapted to each other.

\begin{figure*}[t]
\centering
\includegraphics[width=\textwidth]{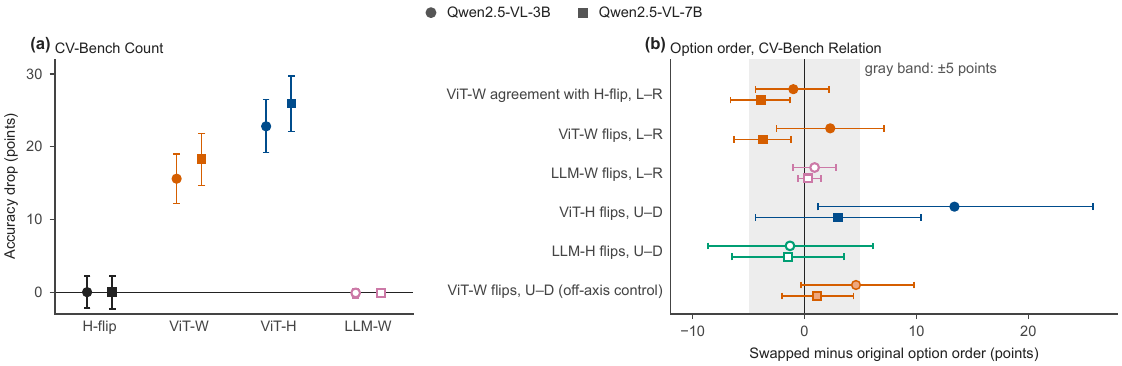}
\caption{The ViT mirrors impair counting, while swapping the option order moves the left--right results by less than 5 points (Qwen2.5-VL, re-run; 95\% paired-bootstrap intervals). (a) Accuracy drop on CV-Bench Count. (b) Swapped minus original option order for six main CV-Bench Relation quantities; gray band: $\pm$5 points. Only the point estimate of the 3B \vith{} up--down flip rate lies outside the band; its mean over both orders is reported.}
\label{fig:robust}
\end{figure*}

\paragraph{Side effects of the mirror}
The width mirror is not a clean intervention on fine content. When the model reads a single printed word, the mirror tends to reverse the order of the letters but also impairs glyph recognition. Accuracy under \vitw{} drops to zero, yet \nb{words-rev2-range} of the readings begin with the first letters of the word spelled backwards, against \nb{words-ctl2-range} for a control word (post hoc, \ref{app:words}). In referring-expression localization on RefCOCO~\cite{yu2016refcoco}, the \vitw{} box coincides post hoc with the box that the model outputs for the really flipped image (IoU $\geq 0.5$) on \nb{bx-hflip-vitw-7b} \nb{bx-hflip-vitw-7b-ci} of the \nb{bx-n-7b} usable items of \qlarge{} and on \nb{bx-hflip-vitw-3b} \nb{bx-hflip-vitw-3b-ci} of the \nb{bx-n-3b} items of \qsmall, against \nb{bx-hflip-orig-7b} and \nb{bx-hflip-orig-3b} for the original input (\llmw: \nb{bx-hflip-llmw-7b} and \nb{bx-hflip-llmw-3b}). Under the mirror the models thus output the box they give for the flipped image more often than not: they pick the object that the flip moves to the queried side and report it in the flipped frame. The boxes of \nb{bx-other-7b} of the items (\qlarge) fit none of the classes. The mirrors also impair a non-spatial judgment. On the \nb{optorder-cnt-n} counting questions of CV-Bench, the accuracy drop (original minus edited) of the two Qwen2.5-VL models is \nb{optorder-cnt-dropabs-vitw-3b} and \nb{optorder-cnt-dropabs-vitw-7b} points under \vitw{} and \nb{optorder-cnt-dropabs-vith-3b} and \nb{optorder-cnt-dropabs-vith-7b} points under \vith{}. The real flip and \llmw{} leave accuracy unchanged (\cref{fig:robust}a; \ref{app:ties}). Counting is thus clearly impaired. The left--right conclusion therefore does not rest on the change rate. With two options any change lands on the opposite relation; the direction is supported by the four options of What'sUp, where \nb{wu-mirror-vitw-lr-range} of the geometry-consistent left--right answers turn into the mirror relation (\cref{sec:res:whatsup}), and by the post hoc order split (\cref{sec:res:lr}). The half-grid shift, however, turns a similar share of the answers it changes into the mirror relation. Across the two option orders, \vitw{} agrees with the real flip on \nb{optorder-agree-vitw-lr-sw-3b}--\nb{cv-agree-vitw-3b} (3B) and \nb{optorder-agree-vitw-lr-sw-7b}--\nb{cv-agree-vitw-7b} (7B) of the CV-Bench left--right items (\cref{fig:robust}b), while the other axis changes on only \nb{cv-flip-vitw-ud-3b}--\nb{optorder-flip-vitw-ud-sw-3b} and \nb{cv-flip-vitw-ud-7b}--\nb{optorder-flip-vitw-ud-sw-7b} of the geometry-consistent up--down items. Our conclusions therefore concern relational answers and, post hoc, localized boxes, not fine content (\cref{sec:limitations}).

\section{Discussion}
\label{sec:discussion}

\subsection{Where the errors come from}
\label{sec:disc:errors}

A post hoc analysis of the errors offers a partial answer to why the models still get left and right wrong when an edit of the ViT's position code alone already changes most answers in the direction of the flip. When the subject and the reference object of a question are swapped, a correct answer should flip. Across the core models, \nb{swaperr-prior-range} of the left--right and up--down errors do not flip, that is, the model gives the same relation regardless of which object it is asked about. Part of these errors is a pure label prior that a blank image also elicits (\cref{sec:res:ud}); the rest depends on the particular pair of objects and is invisible to a blank-image control. Post hoc, \nb{blankerr-all} \nb{blankerr-all-ci} of the CV-Bench errors of the core models equal the model's answer to a blank image. This share exceeds that among correct answers only for the left--right answers of the Qwen2.5-VL models: \nb{blankerr-lr-3b} against \nb{blankcorr-lr-3b} in \qsmall, in line with its \emph{left} prior, and \nb{blankerr-lr-7b} against \nb{blankcorr-lr-7b} in \qlarge. For up--down answers and in \ivl{} it does not. This pattern is consistent with a failure to bind the relation to the subject and the reference object, the role computation that activation patching traces in the LLM residual stream~\cite{ax2609.35486}. It also fits the report that a lexical bias toward spatial words among the answer options arises on the language side rather than the visual side~\cite{ax2606.01914} and that video LLMs encode motion direction in the vision encoder but fail to bind it to the answer option~\cite{ax2605.22823}. This pattern may also explain why inference-time methods that aggregate likelihoods over flips and subject--object swaps, or check answers against a swap, help on spatial-relation benchmarks~\cite{ax2609.17004,ax2608.03631}: such checks discount answer preferences that do not depend on the roles. The swap test alone cannot distinguish such a binding failure from a failure to locate the objects, or from ignoring the reference object, that is, answering from the subject's absolute position. Such an answer gives the same label after a swap when both objects lie on the same side of the midline.

\subsection{Relation to other mechanistic accounts}
\label{sec:disc:relation}

Our results complement interventions on the LLM side. Direction patching places the causal effect of spatial directions in the middle and deep LLM layers, including in \qlarge~\cite{ax2608.22916}. Spatial relation judgments have also been described as coarse object grounding followed by reasoning, without precise localization, mediated by a few LLM heads~\cite{ax2608.23074}. Together with our localization, this suggests a possible chain: the position code of the global-attention ViT blocks would shape the direction carried by the visual tokens, and the LLM would carry it to the answer in its middle layers. We have not tested this chain as a whole. In Qwen3-VL-4B, causal patching between images and their horizontal flips, on items answered correctly in both orientations, indicates that spatial relations are read out through the text tokens of the query rather than directly from the image tokens~\cite{ax2607.03358}. Our edits act earlier, on the position code with the content fixed, and are compatible with such a route.

\subsection{Limitations}
\label{sec:limitations}

\paragraph{Side effects on fine content}
The width mirror tends to reverse the order of letters but impairs glyph recognition, and the ViT mirrors impair counting (\cref{sec:res:robust}). Our conclusions therefore concern relational answers and, post hoc, localized boxes, not fine content. On four-option questions part of the width-mirror effect is non-directional (\cref{sec:res:whatsup}), and in \qsmall{} part of the flips reflects a fallback to its \emph{left} prior (\cref{sec:res:lr}). On four-option natural images (EmbSpatial) the height mirror changes about half of the originally correct answers on conflict items (\ref{app:emb}). Same-axis permutations that are not mirrors also flip many left--right answers, the half-grid shift most of them, so part of the mirror effect is shared by random permutations that reverse the order of the two objects (post hoc), and part is caused by disruption alone (\ref{app:protocols}).

\paragraph{Cross-axis leakage and model scale}
In \ivl, mirroring the absolute embedding along one axis also changes some answers on the other axis. On all items the leakage appears smaller in InternVL3-8B, but not significantly so on the geometry-consistent items, and we tested only one larger InternVL model (\cref{sec:res:lr}). All models show a residual off-axis effect of the height mirror on left--right answers (\cref{tab:main}). Our models have at most 8B parameters. The LLM-side result covers three model series: Qwen2.5-VL, where the id test is weak by construction, InternVL3, trained with variable position increments (see below), and Qwen3-VL (\cref{sec:res:llm}).

\paragraph{Scope of the claims}
Our measurements concern answers to relational questions with few options. They show which edit of the position code changes the answer and in which direction; whether the internal representation is mirrored, and how open-ended descriptions behave, lies outside them. The post hoc split by object order on CV-Bench does not separate the order of the two objects from the half of the image in which the subject lands (\ref{app:protocols}).

\paragraph{Scoring format}
CV-Bench questions have two options in a fixed order, and the main runs use bf16 with first-option tie breaking. Removing tied items does not change the conclusions. A re-run of the main CV-Bench experiment with swapped options leaves them unchanged in both Qwen2.5-VL models. Only the height-mirror flip rate of \qsmall{} moves by more than five points, and we report the mean of both orders (\cref{fig:robust}b; \ref{app:ties}). A re-run of \ivl{} with swapped options leaves its conclusions unchanged as well; only its LLM-side height-mirror flip rate moves by more than five points (\nb{ivl-optorder-flipllmhud-orig} to \nb{ivl-optorder-flipllmhud-sw}; \cref{tab:optorder}). Its main runs use a single tile whose aspect ratio is not preserved (\cref{sec:method:ops}); under the default dynamic tiling the LLM-side tests give the same result (\cref{sec:res:llm}).

\paragraph{Support relations and depth}
On What'sUp, a flip of the content and token rows leaves \emph{on} and \emph{under} answers at the flipped-image answer more often than left--right answers. Counted only as returns to the original answer on geometry-consistent items (post hoc), the contrast is clear only in \ivl. It compares support relations, which imply contact, with relations between two separate objects; relation type and axis are confounded, so the contrast cannot be attributed to the vertical axis as such. The row order alone also changes up--down answers, so the contrast is not due to the content alone, and the sensitivity of up--down answers to the LLM-side row order could not be attributed to direction (\cref{sec:res:ud}). The depth result is clear only in \qlarge; in \qsmall{} the effect points the same way but is weaker (\cref{sec:res:depth}). On the depth test half, row permutations of the ViT height code change fewer answers than the height mirror but move the changed answers toward `higher means closer' as often (pooled \nb{depth-perm-dir-perm-7b} against \nb{depth-perm-dir-vith-7b} in \qlarge). Both shares are close to the \nb{depth-heur-base-7bt} of original answers that follow the heuristic, whereas the width permutations reach \nb{depth-perm-w-dir-perm-7b}. The directional part of the depth result is therefore not specific to the mirror.

\paragraph{LLM-side contribution}
With the ViT width code removed, left--right accuracy falls to near chance, in \qsmall{} through a collapse to \emph{left} (\cref{sec:res:llm}). On the few geometry-consistent items that remain, however, the LLM-side order-and-id mirror flips close to 30\% of the answers. Cui et al.~\cite{ax2603.22278} describe a secondary, language-side representation of spatial relations in intermediate LLM layers that can partly act as a backup when the vision-derived ordering is removed. Our results leave room for such a weak LLM-side contribution and, in our models and benchmarks, left--right accuracy stays near chance once the ViT width code is removed. IVC-Prune~\cite{ax2602.03060} describes the RoPE positions in the LLM as implicit visual coordinates that spatial reasoning depends on, and keeps the tokens that anchor them when pruning. Our LLM-side mirrors leave the left--right answer unchanged when the ViT code is intact. The two results concern different measures (benchmark accuracy after token pruning versus the direction of a forced-choice answer), and both can hold if the LLM-side positions help to select or locate tokens without fixing the direction. InternVL3 was trained with variable position increments for visual tokens (V2PE~\cite{ax2412.09616}), which may have made its LLM insensitive to visual-token ids; yet with the ViT width code removed its id mirror alone still flips some answers, and with the ViT code intact Qwen3-VL gives the same null result (\cref{sec:res:llm}). Three studies report LLM-side order or position effects that our left--right tests did not find. In LLaVA-1.5, shuffling the visual tokens lowers accuracy on a synthetic benchmark without semantic shortcuts~\cite{ax2503.17349}; models trained with the visual tokens in reversed order, or with LLM-side positions assigned concentrically around the image center, differ in object hallucination~\cite{ax2410.15926}; and in video LLMs reversing the frame order lowers temporal understanding~\cite{ax2508.11576}. Alam et al.~\cite{ax2601.09954,ax2603.12545} hold that flattening the image into a one-dimensional token sequence limits spatial reasoning. Their evidence comes from retraining LLaVA-1.5-7B variants with other encoders, with and without 2D-RoPE, not from edits of a pretrained model, and 2D-RoPE helps on some benchmarks and hurts on others. The difference may lie in the models, in the task, in the edit (a random shuffle of all tokens disrupts more than a structured mirror of one axis), in the axis (the order of frames is a sequence that no single-image position code fixes) or in benchmark shortcuts. Shortcuts alone do not explain it: on the controlled images of What'sUp, \llmw{} changes \nb{wu-flip-llmw-lr-range} of the geometry-consistent left--right answers, and in Qwen3-VL-4B the order-only mirror \nb{qwen3-wu-ordonly}. Our LLM-side result is therefore limited to left--right relations within a single image in the three model series. %

\paragraph{Image tiles, video and frame of reference}
Without a thumbnail, the LLM of \ivl{} uses the order of two separately encoded tiles; with the thumbnail of its default tiling, mirroring the order of the tiles and of the tokens within them changed few left--right answers (\cref{sec:res:llm}). Our results concern static images and the viewer's (camera) frame of reference; motion direction in video and questions posed in another reference frame were not tested. Images that were mirrored upstream, such as front-camera selfies or radiographs shown in a different display convention, change the correct answer in the viewer's frame; applications that act on left--right answers need to know whether the image was mirrored.

\section{Conclusion}
\label{sec:conclusion}

In Qwen2.5-VL-7B, Qwen3-VL-4B and InternVL3-2B, the position code of the vision encoder alone suffices to change most answers to left--right questions: editing only that code, with the image untouched, reproduces most of the answers these models give for the flipped image on two benchmarks, a rate that also contains changes from disturbing the code. A half-grid shift of the code, a non-mirror permutation, flips most of them as well. Post hoc and per item on CV-Bench, in these models and in Qwen2.5-VL-3B, random permutations that reverse the order of the two objects flip more answers than those that keep it, which still flip \nb{objorder-shuf-keep-range}. On CV-Bench the answer thus tends to follow the objects' order, which we did not separate from the half of the image in which the subject lands; the changes also reflect disruption. With the four options of What'sUp, a change to the opposite side comes with coherent remappings such as the mirror and the shift, not with random permutations, in Qwen2.5-VL and Qwen3-VL. In Qwen2.5-VL-3B part of the effect is a fallback to the \emph{left} prior. With that code intact, we detected no use by the language model of the position ids or of the order of the visual tokens for left--right relations within a single image, also under the default tiling of InternVL3; the id test is weak in Qwen2.5-VL. With the ViT width code removed, a weak LLM-side contribution is not excluded. In Qwen2.5-VL-7B the effect sits in the global-attention ViT blocks, chiefly the last. Up--down answers change under the ViT height mirror to a degree that varies with benchmark and model; on What'sUp the height mirror flips \nb{hperm-mmax-wu-range} points more up--down answers than the larger of the non-mirror row permutations (post hoc). There, a flip of the content and token rows also leaves \emph{on} and \emph{under} answers at the flipped-image answer more often than left--right answers, a contrast between support and projective relations. Counted only as returns to the original answer (post hoc), it is clear only in InternVL3-2B, and a contribution of the row order cannot be excluded. On CV-Bench 3D, the use by Qwen2.5-VL-7B of the cue that higher in the image means farther depends on the intact height code; in 3B the effect is weaker, and on conflict items the height and width mirrors cannot be told apart.

These results have two implications. For diagnosis, the left--right answer changes with the encoder's position code in the four models with permutation controls (the three headline models and Qwen2.5-VL-3B). Post hoc, on CV-Bench, its direction tends to follow the objects' order, which we did not separate from the half of the image in which the subject lands. For up--down answers this varies with benchmark and model. Where the relation itself is computed was not localized, however, and a post hoc analysis suggests that most remaining errors do not depend on which object a question is about. Repairs may therefore target the binding of a relation to its objects, the localization of the objects, the use of the reference object or, for left--right answers in Qwen2.5-VL, the label prior, rather than only the position code. For spatial fine-tuning, the component swap of SpatialLadder suggests that gains can come from how the LLM uses the output of an unchanged encoder, which the probe can check for other fine-tuned models. Applying the probe requires that the position ids or embeddings can be rewritten, that the visual tokens map to a known patch grid (per tile, if the image is tiled), and that the rewritten code reproduces the original outputs when nothing is mirrored. The cross-axis leakage of the small InternVL model appears smaller at 8B on all items, though not significantly on the geometry-consistent ones. Whether other models with learned absolute embeddings behave alike can be tested with the same protocol. A further step is to test interventions on the binding step rather than on the position code.

\section*{CRediT authorship contribution statement}
\textbf{Enhao Cui:} Conceptualization, Methodology, Software, Formal analysis, Investigation, Visualization, Writing -- original draft, Writing -- review \& editing. \textbf{Junzhi Yang:} Software, Validation, Data curation, Writing -- review \& editing. \textbf{Yaoshun Lin:} Investigation, Resources, Validation, Writing -- review \& editing. \textbf{Yuhan Feng:} Validation, Data curation, Writing -- review \& editing. \textbf{Zijian Chen:} Methodology, Formal analysis, Visualization, Writing -- review \& editing.

\section*{Declaration of competing interest}
The authors declare that they have no known competing financial interests or personal relationships that could have appeared to influence the work reported in this paper.

\section*{Funding}
This research did not receive any specific grant from funding agencies in the public, commercial, or not-for-profit sectors.

\section*{Data availability}
All benchmarks and model checkpoints used are public; their sources and revisions are listed in \ref{app:repro}. The code and the summary output of every analysis are available at \url{https://github.com/chen4study-hub/where-vlms-get-left-right} and archived at Zenodo, \url{https://doi.org/10.5281/zenodo.23691329}~\cite{wherevlms2026data}. %

\section*{Declaration of generative AI and AI-assisted technologies in the manuscript preparation process}
During the preparation of this work the authors used Claude in order to improve language and correct grammar. After using this tool, the authors reviewed and edited the content as needed and take full responsibility for the content of the publication.

\bibliographystyle{elsarticle-num}
\bibliography{refs}

\appendix
\counterwithin{table}{section}%
\counterwithin{figure}{section}
\renewcommand{\thetable}{\Alph{section}.\arabic{table}}%
\renewcommand{\thefigure}{\Alph{section}.\arabic{figure}}
\section{Reproducibility}
\label{app:repro}

\paragraph{Models and data}
\begin{sloppypar}
Qwen/Qwen2.5-VL-3B-Instruct (revision \texttt{66285546d2b8}), Qwen/Qwen2.5-VL-7B-Instruct (\texttt{cc594898137f}), OpenGVLab/InternVL3-2B-hf (\texttt{cb57a075cb75}), Qwen/Qwen3-VL-4B-Instruct (\texttt{ebb281ec}) and, as a scale check, OpenGVLab/InternVL3-8B-hf (\texttt{259a3b64a146}); the public fine-tuned checkpoints hongxingli/SpatialLadder-3B (\texttt{0819c3adf882}), remyxai/SpaceQwen2.5-VL-3B-Instruct (\texttt{332ff3fbc40d}), RUBBISHLIKE/SpaceR-3B (\texttt{079d96362571}) and omlab/Qwen2.5VL-3B-VLM-R1-REC-500steps (\texttt{b9a4c5aa915f}), the last two without a stated license and used for research evaluation only; CV-Bench (nyu-visionx/CV-Bench, \texttt{bc284db50d03}), What'sUp (amitakamath2/whatsup\_vlms, \texttt{373b38706ec0}, MIT license), RefCOCO (lmms-lab-encoder/RefCOCO, \texttt{a5dff0b31947}) and EmbSpatial-Bench~\cite{du2024embspatial} (Phineas476/EmbSpatial-Bench, \texttt{e2733cac}). The images of RefCOCO come from MS-COCO~\cite{lin2014coco}. Full hashes are listed in \texttt{DATA.md} of the repository.
\end{sloppypar}

\paragraph{Environment}
Inference used transformers 4.57.6 and PyTorch 2.5.1 with CUDA 12.4 on a single 32\,GB GPU. %
\begin{sloppypar}
The hooks that rewrite position ids are written against this transformers version; later versions moved some of this logic. For Qwen2.5-VL the hooks replace \nolinkurl{visual.rot_pos_emb} (ViT) and \nolinkurl{model.get_rope_index} (LLM), and for InternVL3 they wrap \nolinkurl{InternVLVisionEmbeddings.interpolate_pos_encoding} (ViT) and pass explicit position ids (LLM). With no mirroring, they reproduce the original rotary embedding element-wise, and the dose--response runs agree item by item with the corresponding whole-model runs (\nb{dose-selfcheck} of the items).
\end{sloppypar}
For InternVL3, the hooks pass explicit consecutive position ids (increment 1), which match the default outputs in a self-check before each run. For Qwen3-VL they replace \nolinkurl{fast_pos_embed_interpolate} and \nolinkurl{rot_pos_emb} (ViT) and \nolinkurl{get_rope_index} (LLM); in the order-only conditions the DeepStack features, which pass intermediate ViT features to early LLM layers, move with their visual tokens. %

\paragraph{Code and outputs}
The code is held in the repository \url{https://github.com/chen4study-hub/where-vlms-get-left-right}, together with the summary output of every analysis (\texttt{outputs/}).

\section{Protocols of the additional analyses}
\label{app:protocols}

\paragraph{Depth (CV-Bench 3D)} The Depth questions were split by item index into a validation half and a held-out test half before the experiments (\nb{depth-test-n-7b} test questions). `Higher' refers to the vertical center of the object's bounding box. A question is heuristic-consistent if the object that is higher in the image is the farther one. The two halves share \nb{depth-halves-shared-img} source images of the underlying data set: \nb{depth-halves-shared-test} of the \nb{depth-halves-test-n} test questions, only \nb{depth-halves-shared-test-counter} of them a conflict item, so the held-out replication is not fully independent. The validation half was scored in bf16 with ties resolved to the first option (the validation rows of \cref{tab:depth} exclude tied items), the test half in fp32 in the original option order with ties counted separately.
\paragraph{Vertical content-only control} What'sUp, all 206 up--down and 206 left--right questions of the core models (\ivl{} with a single tile). The tokens are not reordered, so their rows enter the LLM in mirrored order. The flip-back rate is the share of questions whose answer under the content-only flip differs from the answer for the flipped image; the difference between the two axes uses an unpaired bootstrap (10{,}000 resamples, seed~0), since the two sets of questions differ. The CV-Bench version uses the archived predictions of the main runs (post hoc). %
\paragraph{Error analyses} Subject--object swap: each of the 650 CV-Bench questions is re-asked with subject and reference object exchanged; an error is prior-type if the model gives the same relation label. %
 Referring localization: RefCOCO~\cite{yu2016refcoco} validation expressions containing \emph{left} or \emph{right} with a same-category object on the other side (\nb{refbox-n-7b} usable items for \qlarge).
\paragraph{Same-axis permutations}
The controls permute the merged-token columns (left--right) or rows (up--down) of the ViT position code with pixels unchanged: a half-grid cyclic shift and random permutations with three seeds. They were run on the core models and, post hoc, on Qwen3-VL-4B, whose absolute embedding and 2D-RoPE are permuted together. Each control first re-ran the original condition, which reproduced the original predictions on every item. On CV-Bench the half-grid shift flips \nb{perm-shift-cv-lr-range} of the geometry-consistent left--right answers (mirror: \nb{cv-flip-vitw-lr-range4}), and on What'sUp \nb{perm-mrel-wu-shift-range} of the answers it changes turn into the mirror relation (mirror: \nb{perm-mrel-wu-mirror-range}). On CV-Bench, the mirror flips \nb{perm-mmax-cv-7b} \nb{perm-mmax-cv-7b-ci}, \nb{perm-mmax-cv-3b} \nb{perm-mmax-cv-3b-ci}, \nb{perm-mmax-cv-ivl} \nb{perm-mmax-cv-ivl-ci} and \nb{perm-mmax-cv-qwen3} \nb{perm-mmax-cv-qwen3-ci} points more left--right answers than the larger of the shift and the mean random permutation (\qlarge, \qsmall, \ivl, Qwen3-VL-4B; paired bootstrap over items, 10{,}000 resamples, seed~0), and on What'sUp \nb{perm-mmax-wu-7b} \nb{perm-mmax-wu-7b-ci}, \nb{perm-mmax-wu-3b} \nb{perm-mmax-wu-3b-ci}, \nb{perm-mmax-wu-ivl} \nb{perm-mmax-wu-ivl-ci} and \nb{perm-mmax-wu-qwen3} \nb{perm-mmax-wu-qwen3-ci} points (post hoc). On up--down items, the same row permutations of the height code serve as non-mirror controls for the height mirror (\cref{sec:res:ud}). The shift swaps the left and right halves of the position code and therefore reverses the order of two objects on opposite sides of the midline, so it is not a direction-free control. Random permutations of merged-token columns, which reverse the order of any two objects in different columns with probability one half, flip \nb{perm-ratio-cv-7b} \nb{perm-ratio-cv-7b-ci} as many left--right answers as the mirror in \qlarge{} on CV-Bench, and \nb{perm-ratio-other-range} in the other models and on What'sUp (post hoc). A ratio near one half is expected if the answer follows the order of the two objects, but also if a disrupted code leads to a random choice between the two options. With the four options of What'sUp, a changed answer that falls at random on one of the three other options lands on the mirror relation one time in three. Of the left--right answers that random permutations change, \nb{perm-mrel-wu-shuf-q-range} turn into the mirror relation in Qwen2.5-VL and Qwen3-VL, mostly fewer than one in three, against \nb{perm-mrel-wu-mirror-q-range} under the mirror; in \ivl{} the shares are \nb{perm-mrel-wu-shuf-ivl-range} against \nb{perm-mrel-wu-mirror-ivl-r} (post hoc). In Qwen2.5-VL and Qwen3-VL, a change to the opposite side thus comes with the coherent remappings of the columns (mirror and shift), not with random permutations. For the post hoc split by object order, the two objects of each CV-Bench Relation question were located with the MS-COCO annotations of its source image~\cite{lin2014coco}. Of the \nb{objorder-coco-n} questions with a COCO source image, \nb{objorder-coco-ok} could be mapped: the COCO instance boxes were rescaled to the CV-Bench image, the subject is the instance whose box matches the red box drawn in the image (or the only instance of its category), and every instance of the reference category is used. Each box is assigned to the column of the model's merged-token grid that contains its center (the resized grid of the processor in Qwen2.5-VL and Qwen3-VL, $16\times16$ in \ivl). A permutation reverses the order if it reverses the column order of the subject and every reference instance, and keeps it if it keeps all; item--permutation pairs with an object in the subject's column or with mixed outcomes are excluded (none in the core models, \nb{objorder-ambig-qwen3} in Qwen3-VL-4B). The split pools the half-grid shift and the three random permutations (four per item) over the mapped geometry-consistent left--right items, \nb{objorder-n-7b}, \nb{objorder-n-3b}, \nb{objorder-n-ivl} and \nb{objorder-n-qwen3} of the \nb{cv-flip-vitw-lr-7b-n}, \nb{cv-flip-vitw-lr-3b-n}, \nb{cv-flip-vitw-lr-ivl-n} and \nb{qwen3-cv-gc-n} geometry-consistent items of \qlarge, \qsmall, \ivl{} and Qwen3-VL-4B. Intervals of the differences resample items (10{,}000 resamples, seed~0), keeping the four permutations of an item together. Reversing permutations flip \nb{objorder-rev-7b}, \nb{objorder-rev-3b} and \nb{objorder-rev-ivl} of the geometry-consistent left--right answers in \qlarge, \qsmall{} and \ivl, order-preserving ones \nb{objorder-keep-7b}, \nb{objorder-keep-3b} and \nb{objorder-keep-ivl} (differences \nb{objorder-diff-7b} \nb{objorder-diff-7b-ci}, \nb{objorder-diff-3b} \nb{objorder-diff-3b-ci} and \nb{objorder-diff-ivl} \nb{objorder-diff-ivl-ci} points), the mirror \nb{objorder-mirror-7b}, \nb{objorder-mirror-3b} and \nb{objorder-mirror-ivl}, and off-axis permutations \nb{objorder-offaxis-core-range}; in Qwen3-VL-4B reversing permutations flip \nb{objorder-rev-qwen3}, order-preserving ones \nb{objorder-keep-qwen3} (difference \nb{objorder-diff-qwen3} points \nb{objorder-diff-qwen3-ci}), the mirror \nb{objorder-mirror-qwen3} and off-axis permutations \nb{objorder-offaxis-qwen3}. Split by permutation type, the random permutations alone give differences (reversing minus keeping) of \nb{objorder-shuf-diff-7b} \nb{objorder-shuf-diff-7b-ci}, \nb{objorder-shuf-diff-3b} \nb{objorder-shuf-diff-3b-ci}, \nb{objorder-shuf-diff-ivl} \nb{objorder-shuf-diff-ivl-ci} and \nb{objorder-shuf-diff-qwen3} \nb{objorder-shuf-diff-qwen3-ci} points (\qlarge, \qsmall, \ivl, Qwen3-VL-4B), all above zero. Order-preserving random permutations alone flip \nb{objorder-shuf-keep-7b}, \nb{objorder-shuf-keep-3b}, \nb{objorder-shuf-keep-ivl} and \nb{objorder-shuf-keep-qwen3} of these answers. The half-grid shift alone gives \nb{objorder-shift-diff-7b} \nb{objorder-shift-diff-7b-ci}, \nb{objorder-shift-diff-3b} \nb{objorder-shift-diff-3b-ci} and \nb{objorder-shift-diff-qwen3} \nb{objorder-shift-diff-qwen3-ci} points in \qlarge, \qsmall{} and Qwen3-VL-4B, but \nb{objorder-shift-diff-ivl} \nb{objorder-shift-diff-ivl-ci} in \ivl, where it flips \nb{objorder-shift-rev-ivl} of the items whose order it reverses and \nb{objorder-shift-keep-ivl} of those whose order it keeps. The order reading thus rests on the random permutations; that the shift flips most answers is not explained by the order it reverses. The split does not separate the order of the two objects from the half of the image in which the subject lands. Of the items that flip under all three random permutations, only \nb{objorder-c33-7b} of \nb{objorder-c33n-7b} (\qlarge), \nb{objorder-c33-3b} of \nb{objorder-c33n-3b} (\qsmall), \nb{objorder-c33-ivl} of \nb{objorder-c33n-ivl} (\ivl) and \nb{objorder-c33-qwen3} of \nb{objorder-c33n-qwen3} (Qwen3-VL-4B) have the order reversed by all three. %
\paragraph{Removing the ViT width code} Constant ViT width ids in Qwen2.5-VL and absolute embeddings averaged over the width in \ivl. On the few geometry-consistent items that remain (\nb{nowidth-gc-n-7b} in \qlarge{} and \nb{nowidth-gc-n-ivl} in \ivl), the LLM-side order-and-id mirror flips \nb{nowidth-oid-7b} \nb{nowidth-oid-7b-ci} and \nb{nowidth-oid-ivl} \nb{nowidth-oid-ivl-ci}, the id mirror alone \nb{nowidth-llmw-7b} \nb{nowidth-llmw-7b-ci} and \nb{nowidth-llmw-ivl} \nb{nowidth-llmw-ivl-ci}, and the order-only mirror \nb{nowidth-ordonly-7b} \nb{nowidth-ordonly-7b-ci} and \nb{nowidth-ordonly-ivl} \nb{nowidth-ordonly-ivl-ci}. In \qsmall{} only three geometry-consistent items remain, because post hoc \nb{nowidth-left-3b} of its left--right answers are \emph{left} after the removal. %
\paragraph{Reading single printed words}\label{app:words} We showed the models synthetic images of 100 single printed words and asked them to read the word. Under \vitw{} reading accuracy drops to zero. Yet a post hoc analysis shows that \nb{words-rev2-range} of the readings begin with the first letters of the word spelled backwards, against \nb{words-ctl2-range} for a control word. The letters thus tend to be read in reverse order (post hoc prefix match), while glyph recognition is also impaired. %
\paragraph{InternVL3-8B} Run with the same code and settings as \ivl{} (single $448\times448$ tile; revision \texttt{259a3b64a146}) as a scale check for the cross-axis leakage on CV-Bench and What'sUp; it enters \cref{tab:main} and \cref{sec:res:lr}. %
 On all CV-Bench items, the width-mirror leakage into up--down answers is lower in InternVL3-8B than in \ivl{} (8B minus 2B: \nb{leak-all-diff} points \nb{leak-all-diff-ci}, joint bootstrap). On What'sUp, the height mirror swaps left and right on \nb{wu-offaxis-swap-ivl8b} of \nb{wu-offaxis-swap-ivl8b-n} geometry-consistent left--right items of InternVL3-8B, against \nb{wu-offaxis-swap-ivl} of \nb{wu-offaxis-swap-ivl-n} for \ivl{} (difference \nb{wu-swap-8b-diff} points \nb{wu-swap-8b-diff-ci}). On the geometry-consistent CV-Bench left--right items, the height mirror of InternVL3-8B changes \nb{cv-flip-vith-lr-ivl8b} of the answers, against \nb{cv-flip-vith-lr-3b} for \qsmall{} and \nb{cv-flip-vith-lr-7b} for \qlarge; this difference was not tested.
 On the geometry-consistent subset the decrease of the width-mirror leakage from InternVL3-2B to InternVL3-8B (\nb{leak-gc-diff} points \nb{leak-gc-diff-ci}) is not significant. On all items the leakage of \ivl{} is distinguishable from that of \qlarge{} (\qlarge{} minus \ivl: \nb{leak-all-q7diff} points \nb{leak-all-q7diff-ci}) but not from that of \qsmall{} (\nb{leak-all-q3diff} points \nb{leak-all-q3diff-ci}).

\section{Robustness to bf16 ties, option order and the left prior}
\label{app:ties}

In the main bf16 runs, two option logits are sometimes exactly equal; the prediction then falls on the first option, which is always \emph{left} or \emph{above} on CV-Bench. \Cref{tab:ties} recomputes the main metrics after removing every item that involves a tie in any of the conditions being compared. All values change by a few points at most and none of the conclusions changes. On What'sUp, after the same removal, \vitw{} changes \nb{tie-wu-vitw-range} and \llmw{} \nb{tie-wu-llmw-range} of the geometry-consistent left--right answers in the core models. \Cref{tab:split} splits the left--right flip rates under \vitw{} by the original answer (post hoc).

\begin{table*}[!tp]
\centering
\caption{Main CV-Bench metrics before and after removing every item that involves a bf16 tie between the two option logits.}
\label{tab:ties}
\small
\setlength{\tabcolsep}{4pt}
\begin{tabular}{p{0.28\textwidth}rrr}
\toprule
 & Qwen2.5-VL & Qwen2.5-VL & InternVL3 \\
 & 3B & 7B & 2B \\
\midrule
\vitw{} vs.\ \hflip{} agreement (left--right) & \nb{cv-agree-vitw-3b} $\to$ \nb{tie-agree-vitw-3b} & \nb{cv-agree-vitw-7b} $\to$ \nb{tie-agree-vitw-7b} & \nb{cv-agree-vitw-ivl} $\to$ \nb{tie-agree-vitw-ivl} \\
\vitw{} flip rate (geometry-consistent) & \nb{cv-flip-vitw-lr-3b} $\to$ \nb{tie-flip-vitw-3b} & \nb{cv-flip-vitw-lr-7b} $\to$ \nb{tie-flip-vitw-7b} & \nb{cv-flip-vitw-lr-ivl} $\to$ \nb{tie-flip-vitw-ivl} \\
\llmw{} flip rate (geometry-consistent) & \nb{cv-flip-llmw-lr-3b} $\to$ \nb{tie-flip-llmw-3b} & \nb{cv-flip-llmw-lr-7b} $\to$ \nb{tie-flip-llmw-7b} & \nb{cv-flip-llmw-lr-ivl} $\to$ \nb{tie-flip-llmw-ivl} \\
\bottomrule
\end{tabular}
\end{table*}

\begin{table*}[!tp]
\centering
\caption{Post hoc: CV-Bench left--right flip rate under \vitw{} on the geometry-consistent subset, split by the original answer, and the share of \emph{left} answers under \vitw{} (all left--right items). A model that fell back to a \emph{left} prior would flip original \emph{right} answers but not original \emph{left} answers.}
\label{tab:split}
\small
\setlength{\tabcolsep}{4pt}
\begin{tabular}{p{0.36\textwidth}rrr}
\toprule
 & Qwen2.5-VL & Qwen2.5-VL & InternVL3 \\
 & 3B & 7B & 2B \\
\midrule
Flip rate, original answer \emph{left} & \nb{leftsplit-flip-left-3b} & \nb{leftsplit-flip-left-7b} & \nb{leftsplit-flip-left-ivl} \\
Flip rate, original answer \emph{right} & \nb{leftsplit-flip-right-3b} & \nb{leftsplit-flip-right-7b} & \nb{leftsplit-flip-right-ivl} \\
\emph{left} answers under \vitw{} & \nb{leftsplit-vitw-left-3b} & \nb{leftsplit-vitw-left-7b} & \nb{leftsplit-vitw-left-ivl} \\
\emph{left} answers, original & \nb{cv-left-orig-3b} & \nb{cv-left-orig-7b} & \nb{cv-left-orig-ivl} \\
\bottomrule
\end{tabular}
\end{table*}

On CV-Bench \emph{left} and \emph{above} are always option~A, so we re-ran the main experiment on Qwen2.5-VL with the options in reverse order (\cref{tab:optorder}). The conclusions do not depend on the option order: with swapped options \vitw{} flips \nb{optorder-flip-vitw-lr-sw-3b} and \nb{optorder-flip-vitw-lr-sw-7b} of the geometry-consistent left--right answers and \llmw{} \nb{optorder-flip-llmw-lr-sw-3b} and \nb{optorder-flip-llmw-lr-sw-7b}. Eleven of the twelve quantities move by at most five points; the exception is the height-mirror flip rate of \qsmall{}, for which we report both orders and use their mean in the main text. The lower flip rate of \qsmall{} for original \emph{left} answers (\cref{tab:split}) persists when \emph{left} is option~B (\nb{optorder-flipby-left-sw-3b} against \nb{optorder-flipby-right-sw-3b} for \emph{right}; \qlarge{}: \nb{optorder-flipby-left-sw-7b} and \nb{optorder-flipby-right-sw-7b}). It therefore follows the label, not the option position. A re-run of \ivl{} with swapped options gives the same result (\cref{tab:optorder}); of its six quantities only the \llmh{} up--down flip rate moves by more than five points (\nb{ivl-optorder-flipllmhud-orig} to \nb{ivl-optorder-flipllmhud-sw}). %

\begin{table*}[!tp]
\centering
\caption{Main CV-Bench metrics with the original and the swapped option order (re-runs; flip rates on the geometry-consistent subset of each order). Difference: swapped minus original; for Qwen2.5-VL with a joint bootstrap interval over the 650 items (10{,}000 resamples, seed~0), for \ivl{} point estimates only. bf16 ties: \nb{optorder-ties-orig-3b} / \nb{optorder-ties-sw-3b} (3B) and \nb{optorder-ties-orig-7b} / \nb{optorder-ties-sw-7b} (7B) of 12{,}350 rows per order. For \ivl{}, the off-axis up--down row uses the up--down items answered correctly on the original and the vertically flipped image, so it is not comparable with \cref{tab:main}, whose subset is defined by the horizontal flip.}
\label{tab:optorder}
\footnotesize
\setlength{\tabcolsep}{2pt}
\begin{tabular}{>{\raggedright\arraybackslash}p{0.19\textwidth}rrrrrrrrr}
\toprule
 & \multicolumn{3}{c}{Qwen2.5-VL-3B} & \multicolumn{3}{c}{Qwen2.5-VL-7B} & \multicolumn{3}{c}{InternVL3-2B} \\
\cmidrule(lr){2-4}\cmidrule(lr){5-7}\cmidrule(lr){8-10}
 & orig. & swapped & difference & orig. & swapped & difference & orig. & swapped & diff. \\
\midrule
\vitw{} vs.\ \hflip{}, agreement, left--right & \nb{optorder-agree-vitw-lr-orig-3b} & \nb{optorder-agree-vitw-lr-sw-3b} & \nb{optorder-agree-vitw-lr-diff-3b} {\scriptsize \nb{optorder-agree-vitw-lr-diff-3b-ci}} & \nb{optorder-agree-vitw-lr-orig-7b} & \nb{optorder-agree-vitw-lr-sw-7b} & \nb{optorder-agree-vitw-lr-diff-7b} {\scriptsize \nb{optorder-agree-vitw-lr-diff-7b-ci}} & \nb{ivl-optorder-agreevitwhfliplr-orig} & \nb{ivl-optorder-agreevitwhfliplr-sw} & \nb{ivl-optorder-agreevitwhfliplr-diff} \\
\vitw{} flip, left--right & \nb{optorder-flip-vitw-lr-orig-3b} & \nb{optorder-flip-vitw-lr-sw-3b} & \nb{optorder-flip-vitw-lr-diff-3b} {\scriptsize \nb{optorder-flip-vitw-lr-diff-3b-ci}} & \nb{optorder-flip-vitw-lr-orig-7b} & \nb{optorder-flip-vitw-lr-sw-7b} & \nb{optorder-flip-vitw-lr-diff-7b} {\scriptsize \nb{optorder-flip-vitw-lr-diff-7b-ci}} & \nb{ivl-optorder-flipvitwlr-orig} & \nb{ivl-optorder-flipvitwlr-sw} & \nb{ivl-optorder-flipvitwlr-diff} \\
\llmw{} flip, left--right & \nb{optorder-flip-llmw-lr-orig-3b} & \nb{optorder-flip-llmw-lr-sw-3b} & \nb{optorder-flip-llmw-lr-diff-3b} {\scriptsize \nb{optorder-flip-llmw-lr-diff-3b-ci}} & \nb{optorder-flip-llmw-lr-orig-7b} & \nb{optorder-flip-llmw-lr-sw-7b} & \nb{optorder-flip-llmw-lr-diff-7b} {\scriptsize \nb{optorder-flip-llmw-lr-diff-7b-ci}} & \nb{ivl-optorder-flipllmwlr-orig} & \nb{ivl-optorder-flipllmwlr-sw} & \nb{ivl-optorder-flipllmwlr-diff} \\
\vith{} flip, up--down & \nb{optorder-flip-vith-ud-orig-3b} & \nb{optorder-flip-vith-ud-sw-3b} & \nb{optorder-flip-vith-ud-diff-3b} {\scriptsize \nb{optorder-flip-vith-ud-diff-3b-ci}} & \nb{optorder-flip-vith-ud-orig-7b} & \nb{optorder-flip-vith-ud-sw-7b} & \nb{optorder-flip-vith-ud-diff-7b} {\scriptsize \nb{optorder-flip-vith-ud-diff-7b-ci}} & \nb{ivl-optorder-flipvithud-orig} & \nb{ivl-optorder-flipvithud-sw} & \nb{ivl-optorder-flipvithud-diff} \\
\llmh{} flip, up--down & \nb{optorder-flip-llmh-ud-orig-3b} & \nb{optorder-flip-llmh-ud-sw-3b} & \nb{optorder-flip-llmh-ud-diff-3b} {\scriptsize \nb{optorder-flip-llmh-ud-diff-3b-ci}} & \nb{optorder-flip-llmh-ud-orig-7b} & \nb{optorder-flip-llmh-ud-sw-7b} & \nb{optorder-flip-llmh-ud-diff-7b} {\scriptsize \nb{optorder-flip-llmh-ud-diff-7b-ci}} & \nb{ivl-optorder-flipllmhud-orig} & \nb{ivl-optorder-flipllmhud-sw} & \nb{ivl-optorder-flipllmhud-diff} \\
\vitw{} flip, up--down (off-axis) & \nb{optorder-flip-vitw-ud-orig-3b} & \nb{optorder-flip-vitw-ud-sw-3b} & \nb{optorder-flip-vitw-ud-diff-3b} {\scriptsize \nb{optorder-flip-vitw-ud-diff-3b-ci}} & \nb{optorder-flip-vitw-ud-orig-7b} & \nb{optorder-flip-vitw-ud-sw-7b} & \nb{optorder-flip-vitw-ud-diff-7b} {\scriptsize \nb{optorder-flip-vitw-ud-diff-7b-ci}} & \nb{ivl-optorder-flipvitwud-orig} & \nb{ivl-optorder-flipvitwud-sw} & \nb{ivl-optorder-flipvitwud-diff} \\
\bottomrule
\end{tabular}
\end{table*}

The same re-run measured counting on the \nb{optorder-cnt-n} CV-Bench Count questions. The accuracy drop (original minus edited) under \vitw{} is \nb{optorder-cnt-drop-vitw-3b} \nb{optorder-cnt-drop-vitw-3b-ci} points for \qsmall{} and \nb{optorder-cnt-drop-vitw-7b} \nb{optorder-cnt-drop-vitw-7b-ci} for \qlarge. Under \vith{} it is \nb{optorder-cnt-drop-vith-3b} \nb{optorder-cnt-drop-vith-3b-ci} and \nb{optorder-cnt-drop-vith-7b} \nb{optorder-cnt-drop-vith-7b-ci} points. The real flip gives \nb{optorder-cnt-drop-hflip-3b} and \nb{optorder-cnt-drop-hflip-7b} points, and \llmw{} \nb{optorder-cnt-drop-llmw-3b} and \nb{optorder-cnt-drop-llmw-7b}. Both ViT mirrors thus clearly impair counting.

\section{Depth on EmbSpatial}
\label{app:emb}

On the near--far questions of EmbSpatial-Bench~\cite{du2024embspatial} (four options, natural indoor images; here `higher' refers to the bottom edge of the box), the height mirror corrects hardly more of the original errors on conflict items than the width mirror: \nb{emb-fix-vith-3b} \nb{emb-fix-vith-3b-ci} against \nb{emb-fix-vitw-3b} \nb{emb-fix-vitw-3b-ci} in \qsmall{} and \nb{emb-fix-vith-7b} \nb{emb-fix-vith-7b-ci} against \nb{emb-fix-vitw-7b} \nb{emb-fix-vitw-7b-ci} in \qlarge, and it changes about half of the originally correct answers on these items (\nb{emb-harm-vith-3b} and \nb{emb-harm-vith-7b}). In a post hoc directional analysis, the height mirror moves answers that follow the `higher means farther' heuristic to the opposite end more often than the width mirror does, by \nb{emb-anti-3b} points \nb{emb-anti-3b-ci} in \qsmall{} and \nb{emb-anti-7b} points \nb{emb-anti-7b-ci} in \qlarge. The post hoc split of \cref{sec:res:depth} by whether the original answer follows the heuristic, however, does not replicate on this benchmark. On all items of the near--far questions (post hoc), the height mirror changes \nb{emb-split-vith-con-7b} \nb{emb-split-vith-con-7b-ci} of the answers that follow the heuristic and \nb{emb-split-vith-inc-7b} \nb{emb-split-vith-inc-7b-ci} of those that do not in \qlarge, and \nb{emb-split-vith-con-3b} \nb{emb-split-vith-con-3b-ci} and \nb{emb-split-vith-inc-3b} \nb{emb-split-vith-inc-3b-ci} in \qsmall{}. EmbSpatial therefore gives at most weak and mixed support for \cref{sec:res:depth}. The two benchmarks also define `higher' differently (the vertical center of the box on CV-Bench, its bottom edge here), which may partly explain why the result does not replicate. The height mirror also changes many correct answers on this benchmark (\nb{emb-harm-vith-3b} and \nb{emb-harm-vith-7b} of the originally correct answers in the conflict group), so it is not a selective intervention on four-option natural images.

\section{Public spatially fine-tuned checkpoints}
\label{app:ft}

\begin{table*}[!tp]
\centering
\caption{Public spatially fine-tuned versions of Qwen2.5-VL-3B. Agreement: share of left--right items, among those whose answer changes under \hflip, on which \vitw{} gives the \hflip{} answer ($n$ in parentheses). Gain: CV-Bench accuracy relative to the base model (paired bootstrap, 2{,}000 resamples, both option orders). 3D conflict: accuracy on the 64 CV-Bench 3D depth items where the heuristic is wrong.}
\label{tab:ft}
\footnotesize
\setlength{\tabcolsep}{3pt}
\begin{tabular}{>{\raggedright\arraybackslash}p{0.24\textwidth}rrrr}
\toprule
Model & Agreement & \llmw{} change & Gain (points) & 3D conflict \\
\midrule
Base (Qwen2.5-VL-3B) & \nb{pubft-agree-base} (\nb{pubft-agree-base-n}) & \nb{pubft-llmw-base} & -- & \nb{pubft-d3conf-base} \\
SpatialLadder-3B~\cite{spatialladder} & \nb{pubft-agree-sl} (\nb{pubft-agree-sl-n}) & \nb{pubft-llmw-sl} & \nb{pubft-gain-sl} \nb{pubft-gain-sl-ci} & \nb{pubft-d3conf-sl} \\
SpaceQwen2.5-VL-3B~\cite{spaceqwen} & \nb{pubft-agree-sq} (\nb{pubft-agree-sq-n}) & \nb{pubft-llmw-sq} & \nb{pubft-gain-sq} \nb{pubft-gain-sq-ci} & \nb{pubft-d3conf-sq} \\
SpaceR-3B~\cite{spacer} & \nb{pubft-agree-sr} (\nb{pubft-agree-sr-n}) & \nb{pubft-llmw-sr} & \nb{pubft-gain-sr} \nb{pubft-gain-sr-ci} & \nb{pubft-d3conf-sr} \\
VLM-R1-REC-500steps~\cite{vlmr1} & \nb{pubft-agree-r1} (\nb{pubft-agree-r1-n}) & \nb{pubft-llmw-r1} & \nb{pubft-gain-r1} \nb{pubft-gain-r1-ci} & \nb{pubft-d3conf-r1} \\
\bottomrule
\end{tabular}
\end{table*}

We ran the CV-Bench protocol on four publicly released checkpoints fine-tuned from \qsmall{} for spatial reasoning and on the base model (revisions in \ref{app:repro}), in fp32 and with both option orders (no ties). Because these checkpoints were trained to generate reasoning, each was first checked by greedy generation on 30 items, and a model was included only if its generated answer agrees with the logit-based answer on at least 80\% of them. All four were included. The VLM-R1 checkpoint reached exactly this level (\nb{pubft-gen-r1}) and some of its generations are degenerate; without it, the result below holds for the remaining three. %
\Cref{tab:ft} lists the results, which \cref{fig:ft}a plots. Agreement is computed here on the left--right items whose answer changes under the real flip, so the original answer never matches the flipped-image answer. It is computed in fp32 and in the original option order, and the values are therefore not comparable with the agreement in \cref{tab:main}. All four checkpoints are fine-tuned from Qwen2.5-VL-3B, and the result does not extend to other architectures. In all four, \vitw{} agrees with \hflip{} on \nb{pubft-agree-ft-range} of these items, against \nb{pubft-agree-base} for the base model, and \llmw{} changes at most \nb{pubft-llmw-max} of the answers. The CV-Bench 3D column is descriptive.

In a second experiment we swapped components between Spatial\-Ladder-3B, the checkpoint with the largest gain, and its base model. The two models differ in all 32 ViT blocks, in the merger and in almost every LLM tensor. Each hybrid was assembled from the two checkpoints, reloaded from disk and checked tensor by tensor against its intended source, and evaluated with the code of the checkpoint screening. The re-run base and SpatialLadder models reproduce the screening predictions on all \nb{swap-repro-n} rows of each model (\nb{swap-repro-same} identical). Let $G$ be the gain of SpatialLadder over the base model, and $V$ and $L$ the gains of the hybrids with only the fine-tuned vision side or only the fine-tuned LLM. The hybrid with the fine-tuned LLM recovers \nb{swap-share-llms} \nb{swap-share-llms-ci} of the gain and the hybrid with the fine-tuned vision side \nb{swap-share-vits} \nb{swap-share-vits-ci} (\cref{tab:ftswap}). Assigning the merger to the LLM side instead gives \nb{swap-diff-vitsmb} and \nb{swap-diff-llmsms} points for the two hybrids. %

\begin{table*}[!tp]
\centering
\caption{Component swap between SpatialLadder-3B (S) and its base model Qwen2.5-VL-3B (B) on CV-Bench 2D (650 questions, both option orders, $n=1300$; paired bootstrap 95\% intervals, 2{,}000 resamples). Vision side: vision encoder and merger. Share: difference to the base model divided by the gain of SpatialLadder. Agreement: left--right agreement of \vitw{} with \hflip{} (original option order only, as is the \llmw{} change). Accuracy on the 64 CV-Bench 3D items where the `higher means farther' heuristic is wrong: \nb{swap-d3conf-base} (B/B), \nb{swap-d3conf-ladder} (S/S), \nb{swap-d3conf-llms} (B/S), \nb{swap-d3conf-vits} (S/B).}
\label{tab:ftswap}
\footnotesize
\setlength{\tabcolsep}{3pt}
\begin{tabular}{>{\raggedright\arraybackslash}p{0.24\textwidth}rrrr}
\toprule
Vision side / LLM & Accuracy & $-$ base (points) & Share of gain & Agreement \\
\midrule
B / B (base) & \nb{swap-acc-base} & -- & -- & \nb{swap-agree-base} \\
S / S (SpatialLadder) & \nb{swap-acc-ladder} & \nb{swap-diff-ladder} \nb{swap-diff-ladder-ci} & 1 & \nb{swap-agree-ladder} \\
B / S & \nb{swap-acc-llms} & \nb{swap-diff-llms} \nb{swap-diff-llms-ci} & \nb{swap-share-llms} \nb{swap-share-llms-ci} & \nb{swap-agree-llms} \\
S / B & \nb{swap-acc-vits} & \nb{swap-diff-vits} \nb{swap-diff-vits-ci} & \nb{swap-share-vits} \nb{swap-share-vits-ci} & \nb{swap-agree-vits} \\
ViT blocks 28--31 from S, rest B & \nb{swap-acc-last4} & \nb{swap-diff-last4} \nb{swap-diff-last4-ci} & \nb{swap-share-last4} \nb{swap-share-last4-ci} & \nb{swap-agree-last4} \\
\bottomrule
\end{tabular}
\end{table*}

\end{document}